\documentclass{article}
\usepackage{etoc}
\usepackage{microtype}
\usepackage{graphicx}
\usepackage{subcaption}
\usepackage{booktabs} 
\usepackage{multirow}
\usepackage[table]{xcolor}
\usepackage{float}
\usepackage{pifont}
\usepackage{titletoc}
\definecolor{medgreen}{RGB}{235, 250, 235}
\definecolor{headergray}{gray}{0.93}

\definecolor{color1}{RGB}{221, 236, 248}  
\definecolor{color2}{RGB}{219, 239, 237}  
\definecolor{color3}{RGB}{221, 239, 218}  
\definecolor{color4}{RGB}{231, 224, 241}  

\usepackage{hyperref}

\usepackage[accepted]{icml2026}

\usepackage{amsmath}
\usepackage{amssymb}
\usepackage{mathtools}
\usepackage{amsthm}
\usepackage{enumitem}

\usepackage[section]{placeins} 

\usepackage[capitalize,noabbrev]{cleveref}

\theoremstyle{plain}

\theoremstyle{definition}

\theoremstyle{remark}

\usepackage[textsize=tiny]{todonotes}

\icmltitlerunning{Med-Scout: Curing MLLMs' Geometric Blindness in Medical Perception via Geometry-Aware RL Post-Training}

\begin{document}

\twocolumn[
  \icmltitle{Med-Scout: Curing MLLMs' Geometric Blindness in Medical Perception\\
via Geometry-Aware RL Post-Training}



  \icmlsetsymbol{equal}{*}

  \begin{icmlauthorlist}
    \icmlauthor{Anglin Liu}{hkust(gz)}
    \icmlauthor{Ruichao Chen}{zju}
    \icmlauthor{Yi Lu}{hkust(gz)}
    \icmlauthor{Hongxia Xu}{zju}
    \icmlauthor{Jintai Chen$^\dag$}{hkust(gz),hkust}
  \end{icmlauthorlist}

  \icmlaffiliation{hkust(gz)}{The Hong Kong University of Science and Technology (Guangzhou), Guangzhou, China}
  \icmlaffiliation{zju}{Zhejiang University, Hangzhou, China}
  \icmlaffiliation{hkust}{The Hong Kong University of Science and Technology, Hong Kong, China}
 \icmlcorrespondingauthor{Jintai Chen}{jintaiCHEN@hkust-gz.edu.cn}

  \icmlkeywords{Machine Learning, ICML}

  \vskip 0.3in
]



\printAffiliationsAndNotice{}  

\begin{abstract}
Despite recent Multimodal Large Language Models (MLLMs)' linguistic prowess in medical diagnosis, we find even state-of-the-art MLLMs suffer from a critical perceptual deficit: \textbf{geometric blindness}. This failure to ground outputs in objective geometric constraints leads to plausible yet factually incorrect hallucinations, rooted in training paradigms that prioritize linguistic fluency over geometric fidelity. This paper introduces Med-Scout, a novel framework that ``cures'' this blindness via Reinforcement Learning (RL) that leverages the intrinsic geometric logic latent within unlabeled medical images. Instead of relying on costly expert annotations, Med-Scout derives verifiable supervision signals through three strategic proxy tasks inspired by the systematic reading and reasoning patterns of clinicians: Hierarchical Scale Localization, Topological Jigsaw Reconstruction, and Anomaly Consistency Detection. To rigorously quantify this deficit, we present Med-Scout-Bench, a new benchmark specifically designed to evaluate geometric perception. Extensive evaluations show that Med-Scout significantly mitigates geometric blindness, outperforming leading proprietary and open-source MLLMs by over \textbf{40\%} on our benchmark. Furthermore, this enhanced geometric perception generalizes to broader medical understanding, achieving superior results on radiological and comprehensive medical VQA tasks. The project page is: \url{https://github.com/HKUSTGZ-ML4Health-Lab/Med-Scout}.
\end{abstract}

\section{Introduction}
The deployment of Multimodal Large Language Models (MLLMs) in medical imaging necessitates a fundamental shift in priorities compared to general-domain vision-language tasks~\cite{DBLP:journals/corr/abs-2504-21051}. While general-purpose models often prioritize open-ended creativity and linguistic fluency, medical visual perception demands an uncompromising adherence to objective geometric constraints. A clinical AI system must not only interpret high-level semantic features but also rigorously respect the intrinsic geometric constraints of the image, such as relative anatomical scales, precise topology, and pixel-level structural consistency. However, despite the rapid evolution of both general~\cite{DBLP:journals/corr/abs-2304-08485, DBLP:journals/corr/abs-2502-13923, DBLP:journals/corr/abs-2511-21631} and clinically adapted MLLMs~\cite{DBLP:journals/corr/abs-2406-19280, DBLP:journals/corr/abs-2506-07044, DBLP:journals/corr/abs-2507-05201}, a critical capability gap remains:
while these models excel at generating semantically rich descriptions, they often suffer from \textbf{\textit{geometric blindness}}, failing to ground their outputs in the strict geometric facts.

This misalignment stems largely from the limitations of prevailing training paradigms. Current approaches, including Supervised Fine-Tuning (SFT) and Reinforcement Learning (RL)~\cite{DBLP:journals/corr/abs-2412-05265}, predominantly optimize for semantic alignment, maximizing the likelihood of generating clinically plausible text conditioned on visual encodings. While effective for linguistic fluency, these objectives lack explicit mechanisms to penalize violations of geometric constraints. Consequently, models often generate professionally phrased reports that contradict the actual visual evidence. They essentially master complex medical term while failing to perceive the geometric constraints of the scan accurately. Given the need for such strict geometric verification, RL methods~\cite{DBLP:journals/corr/SchulmanWDRK17}, specifically Group Relative Policy Optimization (GRPO)~\cite{DBLP:journals/corr/abs-2402-03300} presents an ideal optimization framework. Unlike standard supervision that relies on likelihood maximization, GRPO enables the model to learn directly from objective feedback signals. This mechanism is particularly effective for instilling geometric awareness, as it allows us to define and enforce explicit constraints that the model must satisfy.

Building upon this perspective, we introduce \textbf{Med-Scout}, a geometry-aware post-training framework designed to actively cure MLLMs' geometric blindness in medical perception. As illustrated in Figure~\ref{fig:Med-Scout}, our data-centric approach decomposes medical perception into three geometric proxy tasks: 
(1) \textit{Hierarchical Scale Localization} formulates the clinical ``zoom-in'' diagnostic process as a spatial anchoring task, enforcing absolute spatial grounding across varying scales; 
(2) \textit{Topological Jigsaw Reconstruction} translates anatomical topological deduction into a jigsaw puzzle, demanding a robust understanding of global layouts; and 
(3) \textit{Anomaly Consistency Detection} frames fine-grained comparative scrutiny as a ``spot-the-difference'' game, necessitating the precise identification of pixel-level structural discontinuities.
To drive the optimization of these tasks, we present a specialized Dense Geometric Reward (DGR) integrated within the GRPO framework, which provides dense guidance, effectively steering the geometric-semantic alignment to ensure stable and balanced convergence across the constructed tasks.
To support rigorous training and evaluation, we construct a comprehensive dataset comprising over 100K geometrically perturbed samples and curate a balanced 10\% subset to establish Med-Scout-Bench, a novel benchmark that quantifies geometric blindness.

We apply Med-Scout to promote representative general-domain and medical-domain MLLMs. Experiments demonstrate our approach not only significantly reduces geometric blindness but also enables strong generalization to standard medical VQA~\cite{lau2018dataset, DBLP:conf/isbi/LiuZXMYW21, DBLP:journals/corr/abs-2305-10415, DBLP:conf/cvpr/HuLLSHQL24, DBLP:conf/icml/ZuoQLCZHZ0025, butsanets2025radimagenetvqalargescalectmri} and report generation tasks~\cite{demner2015preparing, johnson2019mimic}. This confirms the geometric capabilities acquired effectively transfer to broader medical perception. Our main contributions are as follows:
\begin{itemize}[leftmargin=*]
    \item \textit{Unveiling Geometric Blindness.} We conduct a pilot study and identify a critical gap where MLLMs fail to ground outputs in geometric constraints despite semantic fluency. 
    \item \textit{Geometry-Aware RL Post-Training.} We propose Med-Scout, a novel RL framework that leverages specialized geometric proxy tasks to actively cure geometric blindness in MLLMs using dense geometric rewards.
    \item \textit{The Med-Scout-Bench.} We release Med-Scout-Bench, a novel benchmark constructed from the intrinsic geometric properties of medical images. It serves as a rigorous standard for quantifying geometric perception, addressing a key evaluation gap.
    \item \textit{Substantial Performance Improvements.} Med-Scout improves geometric perception by over \textbf{40\%} on our benchmark and achieves state-of-the-art generalization on radiological and comprehensive medical VQA benchmarks. 
\end{itemize}
\paragraph{Conflict of Interest Disclosure.} The authors declare no conflicts of interest.

\section{Related Work}
\noindent\textbf{MLLMs for the Medical Domain.}
The advancement of medical visual understanding is currently driven by two complementary paradigms. On one front, domain-specific adaptation has achieved remarkable precision through biomedical instruction tuning, as demonstrated by a growing array of specialized models~\cite{DBLP:conf/nips/LiWZULYNPG23, DBLP:journals/corr/abs-2305-17100, DBLP:journals/corr/abs-2406-19280, DBLP:journals/corr/abs-2506-07044, DBLP:conf/miccai/PanLWLZLCOR25}. Simultaneously, general-purpose foundation models~\cite{DBLP:journals/corr/abs-2304-08485, DBLP:journals/corr/abs-2312-14238, DBLP:journals/corr/abs-2508-18265, DBLP:journals/corr/abs-2511-21631} have exhibited surprising zero-shot adaptability to radiological tasks, benefiting from massive-scale pre-training~\cite{zhang2023biomedclip} and robust visual representations. However, despite these distinct strengths, both paradigms share a critical vulnerability: they frequently prioritize linguistic fluency over physical grounding. This results in pervasive ``geometric blindness,'' where models successfully describe pathologies but fail to adhere to the strict spatial constraints and anatomical layouts inherent in medical images.

\noindent\textbf{Enhance MLLM with Proxy Tasks.}
Recent advancements have integrated verifiable proxy tasks into RL to enhance visual grounding. Jigsaw-R1~\cite{DBLP:journals/tmlr/WangZTLXYB25} and Visual Jigsaw~\cite{DBLP:journals/corr/abs-2509-25190} established that grid reconstruction significantly improves fine-grained perception, and formalize such objective tasks into RL pipelines. ViCrit~\cite{DBLP:journals/corr/abs-2506-10128} employs executable programs for robust verification. Approaches like Euclid~\cite{DBLP:journals/corr/abs-2412-08737}, GeoPQA~\cite{DBLP:journals/corr/abs-2509-17437}, and GeoGPT4V~\cite{DBLP:conf/emnlp/CaiBGZSZ24} attempt to augment MLLMs with geometric priors, but their task formulations remain incompatible with the unique requirements of medical perception.
All these methods fail to address comprehensive clinical constraints like multi-scale consistency and anomaly focus.
Moreover, their reliance on sparse binary feedback lacks the granularity essential for complex medical reasoning.

\section{Pilot Study}
\label{pilot_study}

\begin{figure*}[h]
    \centering
    \includegraphics[width=0.9\linewidth]{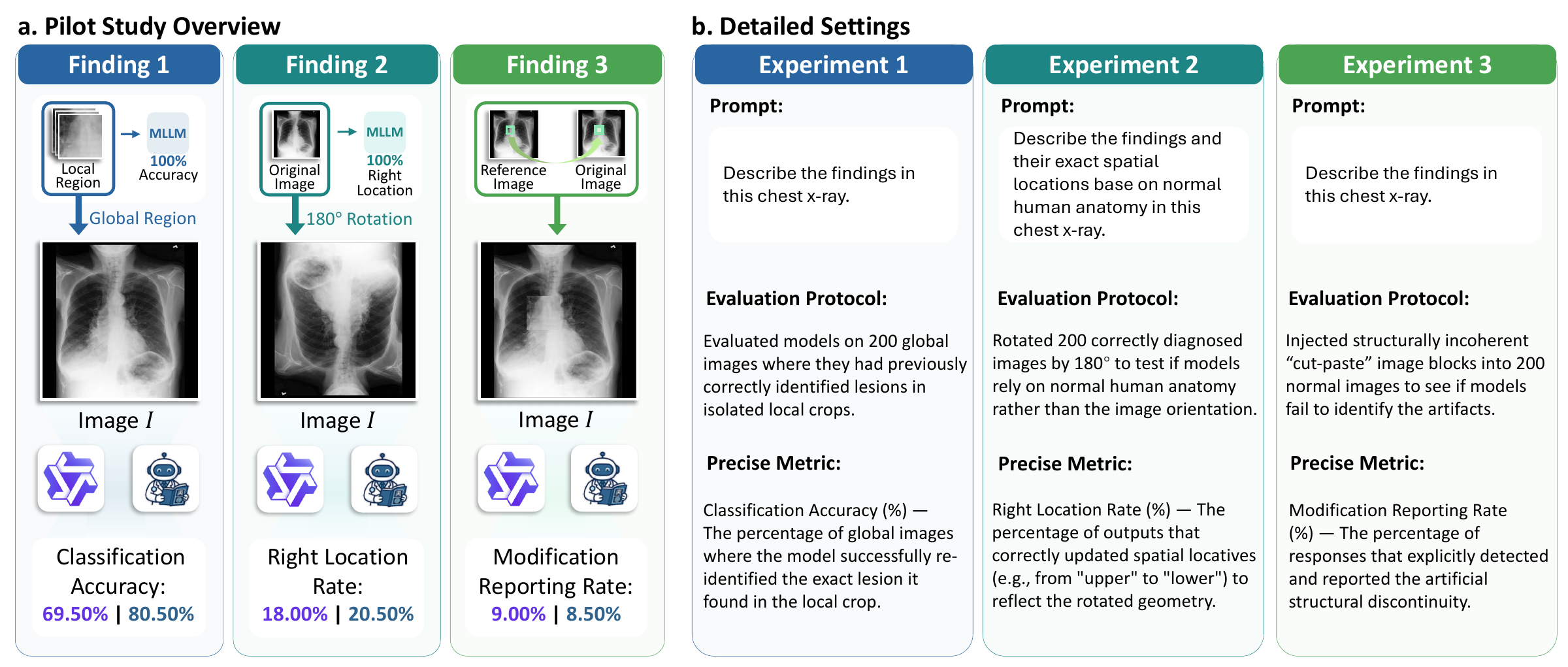}
    \caption{Pilot Study: (1) Scale Blindness: The model correctly describes findings in a local crop but fails in the full global view; (2) Topology Blindness: When the image is rotated, the model fails to update location descriptions; and (3) Anomaly Blindness: The model completely overlooks obvious artificial modifications.}
    \label{fig:pilot_study}
\end{figure*}

We conducted three preliminary experiments on Qwen3-VL-8B-Instruct~\cite{DBLP:journals/corr/abs-2511-21631} and Lingshu-7B~\cite{DBLP:journals/corr/abs-2506-07044} to investigate the geometric fidelity of MLLMs, as shown in Figure~\ref{fig:pilot_study}. Our three findings and extensive experiments are described as follows.

\subsection{Finding I: Inconsistency Between Different Scales}
We investigated whether models can consistently integrate visual features across scales. We specifically filtered for 200 positive samples where the models correctly identified lesions in cropped local views. We then evaluated the models on the corresponding original global images of these successfully recognized crops.

\textbf{Result:} Models failed to detect the same lesions in over 20\% of the global views. This performance drop indicates the models struggle to preserve visual perception when the spatial scale changes.

\subsection{Finding II: Blindness to Relative Spatial Positions}
We tested whether models verify actual relative spatial positions or rely on semantic priors. We selected 200 images with correctly identified pathology locations and rotated them by 180 degrees. In this inverted view, anatomically ``upper'' structures appear at the visual bottom.

\textbf{Result:} Models failed to adapt their spatial descriptions in 80\% of cases. This performance drop indicates the models rely on rigid priors rather than reasoning about the actual topological positions.

\subsection{Finding III: Insensitivity to Structural Anomalies}
We examined sensitivity to pixel-level structural consistency using a ``cut-paste'' protocol. We artificially replaced image patches with structurally incoherent blocks in 200 samples and prompted models to describe findings.

\textbf{Result:} Models failed to identify the artifacts in over 90\% of cases. Rather than identifying the artifact, they produced standard reports that entirely overlooked the artificial perturbations. This indicates the models' blindness to abnormal geometric manifestations.

\subsection{Insufficiency of Chain-of-Thought Reasoning}
To determine whether explicit reasoning could alleviate these perceptual deficits, we repeated the three experiments using Chain-of-Thought (CoT) prompting (Table~\ref{tab:cot_pilot_study}). CoT reasoning yielded only marginal performance fluctuations, failing to fundamentally cure geometric blindness. This indicates that simply enforcing textual reasoning is insufficient without the low-level geometric perception required to ground semantic logic. Crucially, it strongly proves that these failures stem from a fundamental deficit in geometric reasoning capabilities, rather than a mere prompting artifact.

\begin{table}[t]
\centering
\caption{Performance comparison under \textit{No-CoT} and \textit{CoT} modes across the three pilot study experiments.}
\label{tab:cot_pilot_study}
\begin{center}
\begin{small}
\begin{sc}
\resizebox{\columnwidth}{!}{%
\begin{tabular}{llll}
\toprule
Model & Experiment 1 & Experiment 2 & Experiment 3 \\
\midrule
\rowcolor{headergray} \multicolumn{4}{l}{\textit{No-CoT}} \\
Qwen3-VL-8B-Instruct & 69.50\% & 18.00\% & 9.00\% \\
Lingshu-7B & 80.50\% & 20.50\% & 8.50\% \\
\midrule
\rowcolor{headergray} \multicolumn{4}{l}{\textit{CoT}} \\
Qwen3-VL-8B-Instruct & 67.00\% & 20.50\% & 12.00\% \\
Lingshu-7B & 77.50\% & 22.00\% & 7.00\% \\
\bottomrule
\end{tabular}%
}
\end{sc}
\end{small}
\end{center}
\vskip -0.2in
\end{table}

\subsection{Summary}
These experiments reveal critical blindness of current MLLMs: (1) inability to transfer recognition between different scales; (2) blindness to real topological positions; and (3) insensitivity to structural inconsistencies. This semantic-geometric misalignment directly motivates our proposed Med-Scout framework to cure these blind spots, thereby grounding semantic generation in geometric constraints.

\section{Med-Scout}
\begin{figure*}[h]
    \centering
    \includegraphics[width=1\linewidth]{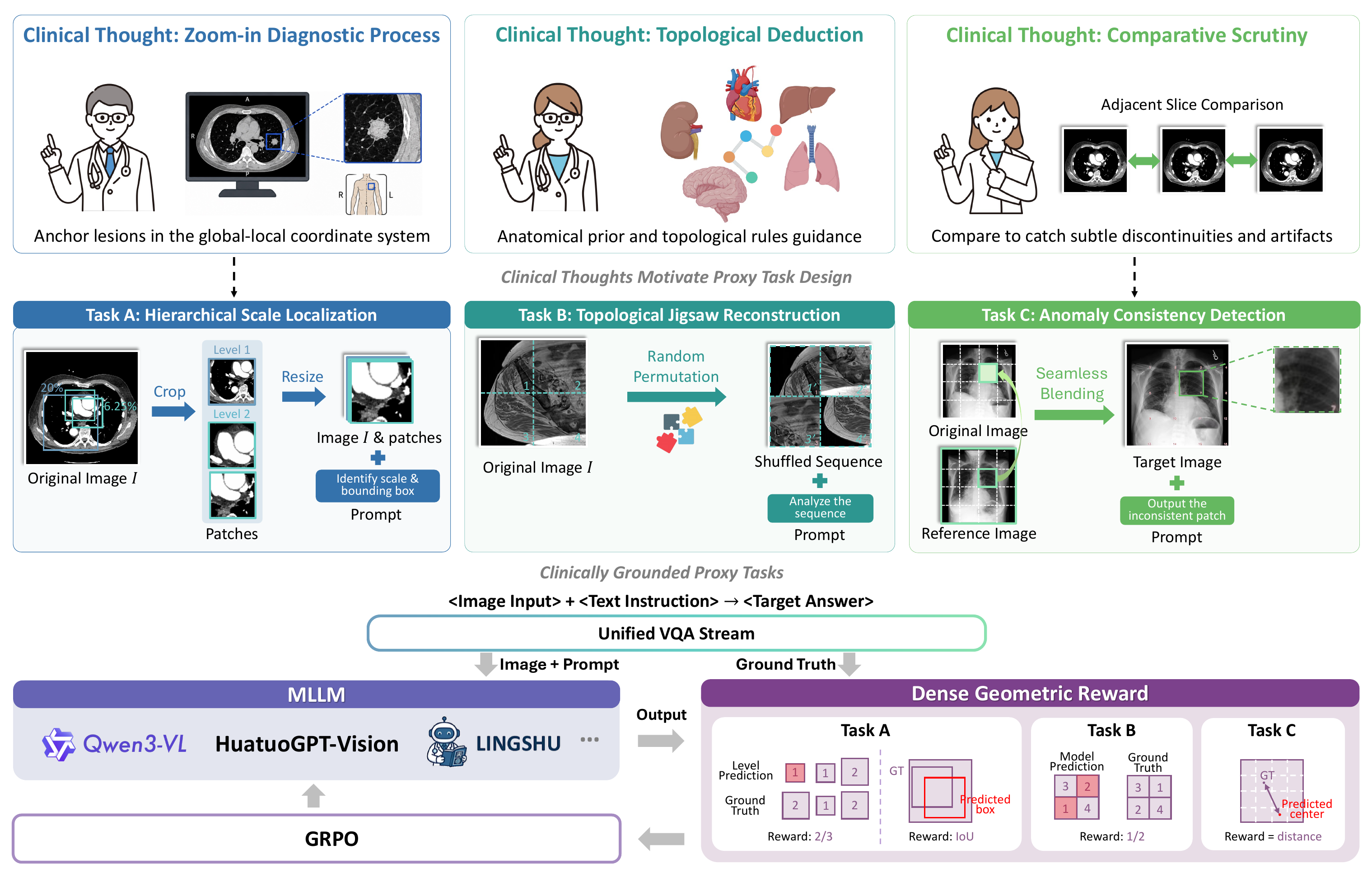}
    \caption{Overview of the Med-Scout Framework. We transform unlabeled medical images into three proxy tasks to cure geometric blindness actively. The framework is optimized using GRPO with a Dense Geometric Reward mechanism that provides stable feedback.}
    \vskip -0.1in
    \label{fig:Med-Scout}
\end{figure*}

\subsection{Task Formulation}
\label{sec:task_formulation}

We formulate Med-Scout as a geometry-aware RL post-training process, as shown in Figure~\ref{fig:Med-Scout}. Our goal is to transform unlabeled medical images into three verifiable geometric proxy tasks $\mathcal{T} = \{\mathcal{T}_\text{scale}, \mathcal{T}_{\text{topo}}, \mathcal{T}_{\text{anom}}\}$ as shown in Appendix~\ref{task_generation}.
Formally, given a raw medical image $I \in \mathbb{R}^{H \times W}$, we generate a VQA case.

\noindent\textbf{Task A: Hierarchical Scale Localization ($\mathcal{T}_\text{{scale}}$).}

\noindent\textbf{Construction:} To enforce multi-scale processing under high cognitive load, we simultaneously crop $N=3$ local patches from the original image $I$. These patches are sampled from two distinct scale levels: Level 1 (20\% of image area) and Level 2 (6.25\% of image area). To filter out non-informative background noise, the center coordinates are strictly restricted to the central normalized region (within the $[0.2, 0.8]$ range).

\textbf{Objective:} The model identifies the scale level and predicts the normalized bounding box $b=(x_1,y_1,x_2,y_2)$ for each of the $N$ patches. This compels the model to maintain multiple coordinate contexts and master absolute spatial grounding.

\noindent\textbf{Task B: Topological Jigsaw Reconstruction ($\mathcal{T}_\text{{topo}}$).}

\noindent\textbf{Construction:} We partition the image $I$ into a $2 \times 2$ grid and apply a random permutation $\sigma$ to generate a shuffled observation $I_\text{{shuffled}}$, retaining sufficient foreground semantic content while requiring both horizontal and vertical spatial reasoning.

\textbf{Objective:} The model must reconstruct the sequence of original indices corresponding to the shuffled positions (reading left-to-right, top-to-bottom). This task forces the model to deduce the canonical global layout of anatomical structures through logical geometric deduction.

\noindent\textbf{Task C: Anomaly Consistency Detection ($\mathcal{T}_\text{{anom}}$).}

\noindent\textbf{Construction:} We employ a fine-grained ``cut-paste'' strategy on a $4 \times 4$ grid. We specifically target the core anatomical region by replacing one patch within the central indices with a reference patch from $I_\text{{ref}}$. To ensure realism, $I_\text{{ref}}$ is selected via modality-specific protocols: adjacent slices for volumetric data (CT/ MRI) or the top-1 most similar image retrieved via BiomedCLIP~\cite{zhang2023biomedclip} for X-ray.

\textbf{Objective:} The model outputs the grid index of the inconsistent patch. By focusing on the high-density central region with a fine granularity, this task necessitates comparative reasoning to detect subtle pixel-level discontinuities that violate anatomical coherence.

\noindent\textbf{Unified VQA Stream \& Optimization.}
We unify these tasks into a standard open-set VQA format as shown in Appendix~\ref{VQA} and require the model to generate precise answers:
\begin{itemize}[leftmargin=*]
    \item \textbf{Direct Mode:} The model directly generates the final answer.
    \item \textbf{Reasoning Mode:} The model generates a CoT~\cite{DBLP:conf/nips/Wei0SBIXCLZ22} reasoning path followed by the answer, allowing it to articulate geometric constraints before concluding.
\end{itemize}
The final training objective is to maximize the expected reward $\mathcal{R}$ across these generative tasks.

\subsection{Reward Formulation}
\label{sec:reward}

To overcome the sparsity of binary feedback in traditional RL, we design a dense geometric reward mechanism. Instead of a simple pass/fail metric, our system calculates continuous reward values based on the degree of geometric deviation, encouraging progressive refinement. The total reward $\mathcal{R}$ is composed of three components:
\begin{equation}
   \mathcal{R} = \mathcal{R}_\text{{acc}} + \mathcal{R}_\text{{fmt}} + \mathbb{I}_\text{{CoT}} \cdot \mathcal{R}_\text{{reason}}
\end{equation}
where $\mathcal{R}_\text{{acc}} \in [0, 1]$ measures task accuracy, $\mathcal{R}_\text{{fmt}} \in [0, 0.5]$ ensures output format compliance, and $\mathcal{R}_\text{{reason}} \in [0, 0.5]$ enforces reasoning structure in CoT mode.

\noindent\textbf{Dense Geometric Rewards ($\mathcal{R}_\text{{acc}}$).}
We tailor specific continuous metrics for each task type to quantify geometric precision, capped at a maximum of 1.0.

\begin{itemize}[leftmargin=*]
    \item \textbf{Scale ($\mathcal{T}_\text{{scale}}$):} 
    This task entails quantifying attributes and localizing targets for $N$ objects simultaneously. We evaluate performance across two dimensions, normalizing by $N$ to ensure invariant optimization magnitude:
    \begin{enumerate}[leftmargin=*]
        \item \textit{Value Estimation:} For discrete scale levels, we compute the average classification correctness. For a target index sequence $Y^*$ of length $N$:
        \begin{equation}
             \mathcal{R}_\text{{val}} = \frac{1}{N} \sum_{i=1}^{N} \mathbb{I}(\hat{y}_i = y^*_i)
        \end{equation}
        \item \textit{Box Localization:} To enforce spatial precision, we calculate the average Intersection over Union (IoU) between the predicted box $\hat{b}_i$ and the ground truth $b^*_i$:
        \begin{equation}
             \mathcal{R}_\text{{box}} = \frac{1}{N} \sum_{i=1}^{N} \text{IoU}(\hat{b}_i, b^*_i)
        \end{equation}
    \end{enumerate}
    
    \item \textbf{Topology ($\mathcal{T}_\text{{topo}}$):} 
    To reward partial logical correctness, we utilize an element-wise alignment metric rather than exact sequence matching. For a target index sequence $S^*$ of length $N$:
    \begin{equation}
        \mathcal{R}_\text{{topo}} = \frac{1}{N} \sum_{i=1}^{N} \mathbb{I}(\hat{s}_i = s^*_i)
    \end{equation}
    This ensures the model receives credit for every correctly positioned patch, even if the global sequence is imperfect.

    \item \textbf{Anomaly ($\mathcal{T}_\text{{anom}}$):} 
    The objective is to identify the single swapped patch index $k \in \{0, \dots, 15\}$ within a $4 \times 4$ grid. To provide dense supervision, we first map the flattened index $k$ to 2D grid coordinates $(u, v)$ via $u = \lfloor k/4 \rfloor$ and $v = k \pmod 4$. We then compute the reward based on the Euclidean distance between the predicted patch coordinates $(\hat{u}, \hat{v})$ and the ground truth $(u^*, v^*)$:
    \begin{equation}
        \mathcal{R}_\text{{anom}} = \exp\left(-\frac{\sqrt{(\hat{u} - u^*)^2 + (\hat{v} - v^*)^2}}{\tau}\right)
    \end{equation}
    where $\tau$ is a temperature hyperparameter. This distance-based mechanism guides the model towards the correct spatial region, rewarding predictions that are geometrically plausible even if the exact index is missed.
\end{itemize}

\noindent\textbf{Universal Format Reward ($\mathcal{R}_\text{{fmt}}$).}
To strictly enforce output protocols, we evaluate format compliance at the item level rather than the sequence level. Given a target answer containing $N$ required elements, the reward is calculated as:
\begin{equation}
    \mathcal{R}_\text{{fmt}} = \frac{0.5}{N} \sum_{i=1}^{N} \mathbb{I}(\hat{a}_i \in \Phi_\text{{regex}})
\end{equation}
where $\Phi_\text{{regex}}$ represents the valid pattern for each sub-component.

\noindent\textbf{Reasoning Structure Reward ($\mathcal{R}_\text{{reason}}$).}
Exclusively in Reasoning Mode, we impose an additional structural constraint to enforce the CoT pattern \texttt{<think>...<answer>...}. This encourages the model to maintain the reasoning buffer:
\begin{equation}
    \mathcal{R}_\text{{reason}} = 
    \begin{cases} 
    0.5 & \text{if } \hat{Y} \in \Phi_{\text{CoT}} \\
    0 & \text{otherwise}
    \end{cases}
\end{equation}
Consequently, a perfectly reasoned and accurate response in CoT mode yields a maximum total reward of $\mathcal{R}=2.0$.

\subsection{Med-Scout-Bench}
\label{sec:benchmark}

To quantitatively evaluate geometric blindness, we introduce Med-Scout-Bench, a novel benchmark encompassing diverse anatomical regions from the mainstream radiological modalities (CT, MRI, and X-ray), pivotal for clinical geometric analysis.

\noindent\textbf{Dataset Construction.} We synthesize an initial pool of 108,000 VQA cases. We utilize \textit{TotalSegmentor} (CT/MRI)~\cite{wasserthal2023totalsegmentator, akinci2025totalsegmentator} to guarantee comprehensive anatomical coverage across the entire body, alongside \textit{MIMIC-CXR}~\cite{johnson2019mimic} for planar radiography. From this pool, we sampled a high-quality subset of 10,800 cases (10\%) as the benchmark. This benchmark is strictly balanced across the three radiological modalities, ensuring unbiased evaluation.

\noindent\textbf{Evaluation Protocol.}
We adopt a unified evaluation setting to ensure consistency: all tasks are formulated as open-set VQA queries rather than multiple-choice options. We employ the LLM-as-a-Judge~\cite{DBLP:journals/corr/abs-2411-15594} paradigm to robustly assess the semantic correctness of these generative responses, overcoming the limitations of rigid string matching.

\section{Experiments}

\subsection{Experimental Settings}

\noindent\textbf{Med-Scout Training.} 
All Med-Scout experiments are conducted for 7,200 steps on a node with $6\times$ NVIDIA RTX PRO 6000 GPUs using GRPO. We optimize the model using AdamW with a peak learning rate of $1\times 10^{-6}$, a cosine decay schedule, and a warm-up ratio of 0.01. For the GRPO configuration, we set the global batch size to 192 and the group size to $G=8$, with a KL divergence coefficient $\beta=0.04$ to ensure stable updates. Training is performed on the 97,200-sample split of our constructed dataset excluding the Med-Scout-Bench.
We adopt the LLM-as-a-Judge paradigm using Gemini-3-Flash to evaluate open-ended responses rigorously.

\noindent\textbf{Baselines.}
To assess the generalizability of our framework across different model scales and domains, we apply Med-Scout to four backbones: the general-purpose Qwen3-VL-4B/8B-Instruct~\cite{DBLP:journals/corr/abs-2511-21631} and the medical specialists Lingshu-7B~\cite{DBLP:journals/corr/abs-2506-07044} and HuatuoGPT-Vision-7B~\cite{DBLP:journals/corr/abs-2406-19280}. We benchmark our approach against a wide range of state-of-the-art (SOTA) MLLMs, including both proprietary and open-source models.
\begin{itemize}[leftmargin=*]
    \item \textbf{Proprietary Models}: We evaluate the latest commercial leaders, GPT-5 and Gemini-3-Flash, to establish upper-bound performance benchmarks.
    \item \textbf{General-purpose MLLMs}: We select representative open-source models including InternVL3-8B~\cite{DBLP:journals/corr/abs-2504-10479}, Qwen2.5-VL-3B-Instruct, Qwen2.5-VL-7B-Instruct~\cite{DBLP:journals/corr/abs-2502-13923}, as well as the newer Qwen3-VL-4B-Instruct and Qwen3-VL-8B-Instruct~\cite{DBLP:journals/corr/abs-2511-21631}.
    \item \textbf{Medical MLLMs}: For domain-specific comparison, we utilize LLaVA-Med-7B~\cite{DBLP:conf/nips/LiWZULYNPG23}, MedGemma-4B-IT~\cite{DBLP:journals/corr/abs-2507-05201}, HuatuoGPT-Vision-7B~\cite{DBLP:journals/corr/abs-2406-19280}, and Lingshu-7B~\cite{DBLP:journals/corr/abs-2506-07044}.
\end{itemize}
To ensure evaluation fairness, all models are assessed within the same standardized evaluation environment.

\noindent\textbf{Benchmarks.} We conduct comprehensive evaluations across three dataset categories. First, to quantify geometric perception, we utilize our Med-Scout-Bench.
Second, we assess standard radiological VQA using RadImageNet-VQA~\cite{butsanets2025radimagenetvqalargescalectmri}, VQA-RAD~\cite{lau2018dataset}, and SLAKE~\cite{DBLP:conf/isbi/LiuZXMYW21}, alongside MIMIC-CXR and IU-Xray for report generation. Finally, to verify broad generalization, we extend our evaluation to PMC-VQA~\cite{DBLP:journals/corr/abs-2305-10415}, OmniMedVQA~\cite{DBLP:conf/cvpr/HuLLSHQL24}, and MedXpertQA~\cite{DBLP:conf/cvpr/HuLLSHQL24} for broader medical domains and diverse modalities.

\noindent\textbf{Evaluation Metrics.}
For Med-Scout-Bench, we report the DGR score computed directly via the reward functions defined in Section~\ref{sec:reward}. For VQA benchmarks, we report the response accuracy. For report generation tasks, we report metrics including ROUGE-L~\cite{lin2004rouge}, CIDEr~\cite{DBLP:conf/cvpr/VedantamZP15}, and SemScore~\cite{DBLP:journals/corr/abs-2401-17072} for both semantic-based and model-based evaluation.

\subsection{Results on Med-Scout-Bench}
\begin{figure}[t]
    \centering
    \includegraphics[width=1\linewidth]{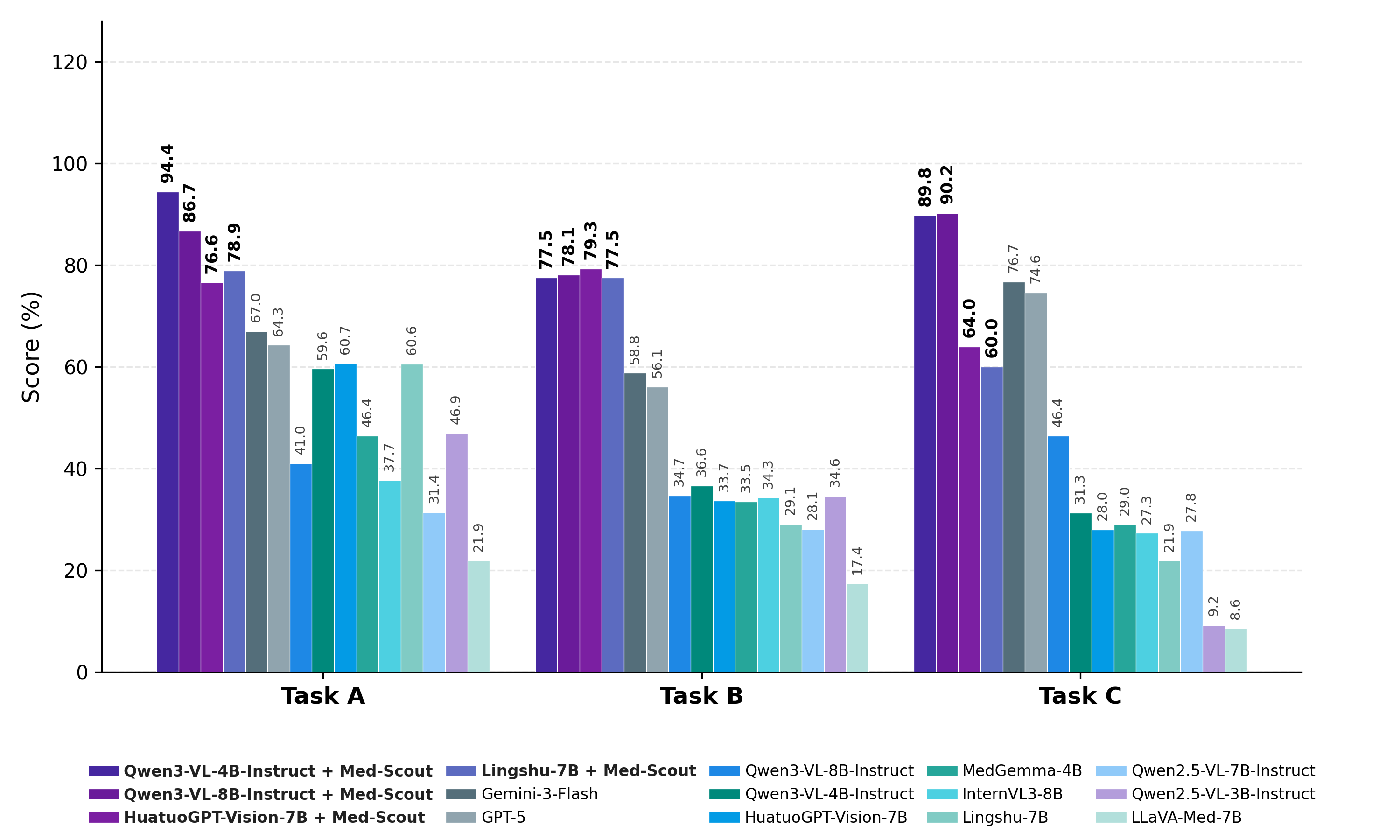}
    \caption{Performance comparison on Med-Scout-Bench.}
    \label{fig:med-scout-bench}
    \vskip -0.2in
\end{figure}
Figure~\ref{fig:med-scout-bench} illustrates the primary models evaluated in this paper, while Table~\ref{tab:med_scout_bench} provides a more detailed performance comparison incorporating additional baseline models on Med-Scout-Bench. As demonstrated in both, Med-Scout yields substantial improvements across all backbones.
For instance, the aligned Qwen3-VL-8B-Instruct improves average accuracy from 39.7\% to 83.6\%, while Lingshu-7B rises from 31.9\% to 71.9\%.
Notably, open-source models outperform leading proprietary models GPT-5 and Gemini-3-Flash. Moreover, general-domain models show greater improvement than medical specialists, suggesting that a strong vision-language foundation enables models to learn geometric features more effectively.

\begin{table*}[t]
\centering
\caption{Performance comparison with SOTA MLLMs across radiological and general medical VQA benchmarks. Rad-VQA represents RadImageNet-VQA. All accuracy metrics are scaled by a factor of 100 to enhance clarity and comprehension. The best results are highlighted.}
\label{tab:main_results}
\begin{center}
\begin{small}
\begin{sc}
\renewcommand{\arraystretch}{0.5}
\begin{tabular*}{\textwidth}{@{\extracolsep{\fill}}lllllll}
\toprule
\multirow{2}{*}{Model} & \multicolumn{3}{c}{Radiological VQA} & \multicolumn{3}{c}{Generalization} \\
\cmidrule(lr){2-4} \cmidrule(lr){5-7}
 & Rad-VQA & VQA-RAD & SLAKE & PMC-VQA & OmniMedVQA & MedXpertQA \\
\midrule
\rowcolor{headergray} \multicolumn{7}{l}{\textit{Proprietary Models}} \\
GPT-5 & 59.1 & 66.4 & 73.9 & 57.7 & 76.9 & 54.8 \\
Gemini-3-Flash & 60.7 & 70.2 & 76.1 & \textbf{58.1} & 75.3 & \textbf{56.0} \\
\midrule
\rowcolor{headergray} \multicolumn{7}{l}{\textit{General-purpose MLLMs}} \\
InternVL3-8B & 58.4 & 65.6 & 72.9 & 52.0 & 78.2 & 22.4 \\
Qwen2.5-VL-3B-Instruct & 54.1 & 60.2 & 63.5 & 50.2 & 61.5 & 24.3 \\
Qwen2.5-VL-7B-Instruct & 55.7 & 65.3 & 67.9 & 51.8 & 63.8 & 21.9 \\
Qwen3-VL-4B-Instruct & 41.5 & 59.9 & 73.4 & 42.8 & 45.5 & 27.0 \\
\rowcolor{color1} \quad + \textbf{Med-Scout} &
45.7{\scriptsize\textcolor{green!40!black}{$\uparrow$\,4.2}} & 
62.9{\scriptsize\textcolor{green!40!black}{$\uparrow$\,3.0}} &
75.6{\scriptsize\textcolor{green!40!black}{$\uparrow$\,2.2}} &
45.1{\scriptsize\textcolor{green!40!black}{$\uparrow$\,2.3}} &
48.8{\scriptsize\textcolor{green!40!black}{$\uparrow$\,3.3}} &
27.7{\scriptsize\textcolor{green!40!black}{$\uparrow$\,0.7}} \\
Qwen3-VL-8B-Instruct & 41.6 & 63.2 & 69.6 & 43.9 & 42.9 & 30.4 \\
\rowcolor{color2} \quad + \textbf{Med-Scout} &
45.3{\scriptsize\textcolor{green!40!black}{$\uparrow$\,3.7}} &
65.8{\scriptsize\textcolor{green!40!black}{$\uparrow$\,2.6}} &
72.0{\scriptsize\textcolor{green!40!black}{$\uparrow$\,2.4}} &
45.5{\scriptsize\textcolor{green!40!black}{$\uparrow$\,1.6}} &
46.0{\scriptsize\textcolor{green!40!black}{$\uparrow$\,3.1}} &
30.8{\scriptsize\textcolor{green!40!black}{$\uparrow$\,0.4}} \\
\midrule
\rowcolor{headergray} \multicolumn{7}{l}{\textit{Medical MLLMs}} \\
LLaVA-Med-7B & 44.3 & 50.6 & 50.1 & 32.4 & 46.8 & 19.9 \\
MedGemma-4B-IT & 49.8 & 70.8 & 77.9 & 48.7 & 70.3 & 22.0 \\
HuatuoGPT-Vision-7B & 48.8 & 67.0 & 67.8 & 53.0 & 75.0 & 22.4 \\
\rowcolor{color3} \quad + \textbf{Med-Scout} &
52.1{\scriptsize\textcolor{green!40!black}{$\uparrow$\,3.3}} &
70.1{\scriptsize\textcolor{green!40!black}{$\uparrow$\,3.1}} &
71.0{\scriptsize\textcolor{green!40!black}{$\uparrow$\,3.2}} &
55.9{\scriptsize\textcolor{green!40!black}{$\uparrow$\,2.9}} &
75.4{\scriptsize\textcolor{green!40!black}{$\uparrow$\,0.4}} &
22.7{\scriptsize\textcolor{green!40!black}{$\uparrow$\,0.3}} \\
Lingshu-7B & 61.2 & 68.9 & 82.8 & 56.3 & 81.4 & 27.4 \\
\rowcolor{color4} \quad + \textbf{Med-Scout} &
\textbf{64.0}{\scriptsize\textcolor{green!40!black}{$\uparrow$\,2.8}} &
\textbf{71.0}{\scriptsize\textcolor{green!40!black}{$\uparrow$\,2.1}} &
\textbf{83.0}{\scriptsize\textcolor{green!40!black}{$\uparrow$\,0.2}} &
57.4{\scriptsize\textcolor{green!40!black}{$\uparrow$\,1.1}} &
\textbf{81.9}{\scriptsize\textcolor{green!40!black}{$\uparrow$\,0.5}} &
28.0{\scriptsize\textcolor{green!40!black}{$\uparrow$\,0.6}} \\
\bottomrule
\end{tabular*}%
\end{sc}
\end{small}
\end{center}
\vskip -0.1in
\end{table*}

\subsection{Results on Radiological VQA Benchmarks}
Tables~\ref{tab:main_results} and~\ref{tab:report_generation} present the evaluation results on radiological VQA and report generation benchmarks, respectively. Med-Scout consistently enhances performance across both discriminative and generative tasks. In VQA, Qwen3-VL-4B achieves a notable 4.2\% gain on Rad-VQA, while our aligned Lingshu-7B surpasses Gemini-3-Flash on VQA-RAD (71.0\% vs. 70.2\%).

The benefits of geometric alignment are even more pronounced in report generation tasks, where geometric accuracy is critical. As shown in Table~\ref{tab:report_generation}, Med-Scout boosts Qwen3-VL-4B's CIDEr score by 4.3\% on MIMIC-CXR. Most remarkably, Lingshu-7B with Med-Scout achieves an SOTA CIDEr, dramatically outperforming proprietary models. This confirms that curing geometric blindness enables models to generate clinically accurate reports with superior geometric consistency.

\subsection{Generalization to Comprehensive Medical VQA Benchmarks}
We further extend our evaluation to broader medical domains using PMC-VQA, OmniMedVQA, and MedXpertQA. As shown in Table~\ref{tab:main_results}, Med-Scout confers consistent improvements, demonstrating that geometric awareness is fundamental to general medical perception. Notably, HuatuoGPT-Vision-7B achieves a substantial 2.9\% gain on PMC-VQA. Even more impressively, Lingshu-7B manages to break through its performance ceiling, further elevating its already SOTA results. This confirms that our framework effectively enhances robustness in complex, open-domain medical perception.

\begin{table*}[t]
\centering
\caption{Performance comparison on medical report generation benchmarks. Results are reported on ROUGE-L, CIDEr, and SemScore metrics. All scores are scaled by a factor of 100 to enhance clarity and comprehension. The best results are highlighted.}
\label{tab:report_generation}
\begin{center}
\begin{small}
\begin{sc}
\resizebox{\textwidth}{!}{%
\renewcommand{\arraystretch}{0.5}
\begin{tabular*}{\textwidth}{@{\extracolsep{\fill}}lllllll}
\toprule
\multirow{2}{*}{Model} & \multicolumn{3}{c}{MIMIC-CXR} & \multicolumn{3}{c}{IU-Xray} \\
\cmidrule(lr){2-4} \cmidrule(lr){5-7}
 & ROUGE-L & CIDEr & SemScore & ROUGE-L & CIDEr & SemScore \\
\midrule
\rowcolor{headergray} \multicolumn{7}{l}{\textit{Proprietary Models}} \\
GPT-5 & 14.7 & 89.4 & 25.2 & 34.6 & 139.9 & 48.3 \\
Gemini-3-Flash & 24.9 & 90.0 & 29.8 & 35.1 & 135.8 & 48.0 \\
\midrule
\rowcolor{headergray} \multicolumn{7}{l}{\textit{General-purpose MLLMs}} \\
InternVL3-8B & 20.7 & 64.3 & 21.1 & 22.4 & 70.2 & 30.6 \\
Qwen2.5-VL-3B-Instruct & 21.4 & 57.7 & 19.4 & 26.7 & 72.9 & 37.2 \\
Qwen2.5-VL-7B-Instruct & 22.9 & 63.9 & 18.6 & 26.6 & 78.8 & 36.3 \\
Qwen3-VL-4B-Instruct & 21.8 & 60.9 & 19.8 & 25.8 & 81.4 & 36.5 \\
\rowcolor{color1} \quad + \textbf{Med-Scout} &
23.4{\scriptsize\textcolor{green!40!black}{$\uparrow$\,1.6}} &
65.2{\scriptsize\textcolor{green!40!black}{$\uparrow$\,4.3}} &
21.5{\scriptsize\textcolor{green!40!black}{$\uparrow$\,1.7}} &
27.1{\scriptsize\textcolor{green!40!black}{$\uparrow$\,1.3}} &
84.2{\scriptsize\textcolor{green!40!black}{$\uparrow$\,2.8}} &
38.6{\scriptsize\textcolor{green!40!black}{$\uparrow$\,2.1}} \\
Qwen3-VL-8B-Instruct & 21.3 & 64.8 & 19.6 & 27.1 & 75.9 & 38.7 \\
\rowcolor{color2} \quad + \textbf{Med-Scout} &
23.8{\scriptsize\textcolor{green!40!black}{$\uparrow$\,2.5}} &
68.1{\scriptsize\textcolor{green!40!black}{$\uparrow$\,3.3}} &
21.0{\scriptsize\textcolor{green!40!black}{$\uparrow$\,1.4}} &
29.5{\scriptsize\textcolor{green!40!black}{$\uparrow$\,2.4}} &
79.6{\scriptsize\textcolor{green!40!black}{$\uparrow$\,3.7}} &
41.6{\scriptsize\textcolor{green!40!black}{$\uparrow$\,2.9}} \\
\midrule
\rowcolor{headergray} \multicolumn{7}{l}{\textit{Medical MLLMs}} \\
LLaVA-Med-7B & 16.4 & 49.2 & 16.7 & 19.5 & 73.7 & 17.4 \\
MedGemma-4B-IT & 27.4 & 83.8 & 30.1 & 30.6 & 107.5 & 46.8 \\
HuatuoGPT-Vision-7B & 23.6 & 75.6 & 24.6 & 30.9 & 109.6 & 40.7 \\
\rowcolor{color3} \quad + \textbf{Med-Scout} &
25.7{\scriptsize\textcolor{green!40!black}{$\uparrow$\,2.1}} &
79.0{\scriptsize\textcolor{green!40!black}{$\uparrow$\,3.4}} &
25.8{\scriptsize\textcolor{green!40!black}{$\uparrow$\,1.2}} &
32.1{\scriptsize\textcolor{green!40!black}{$\uparrow$\,1.2}} &
111.7{\scriptsize\textcolor{green!40!black}{$\uparrow$\,2.1}} &
43.1{\scriptsize\textcolor{green!40!black}{$\uparrow$\,2.4}} \\
Lingshu-7B & 30.9 & 104.9 & 29.7 & 37.7 & 180.8 & 48.4 \\
\rowcolor{color4} \quad + \textbf{Med-Scout} &
\textbf{31.4}{\scriptsize\textcolor{green!40!black}{$\uparrow$\,0.5}} &
\textbf{105.2}{\scriptsize\textcolor{green!40!black}{$\uparrow$\,0.3}} &
\textbf{30.3}{\scriptsize\textcolor{green!40!black}{$\uparrow$\,0.6}} &
\textbf{38.0}{\scriptsize\textcolor{green!40!black}{$\uparrow$\,0.3}} &
\textbf{183.3}{\scriptsize\textcolor{green!40!black}{$\uparrow$\,2.5}} &
\textbf{48.6}{\scriptsize\textcolor{green!40!black}{$\uparrow$\,0.2}} \\
\bottomrule
\end{tabular*}%
}
\end{sc}
\end{small}
\end{center}
\vskip -0.1in
\end{table*}

\subsection{Direct Mode vs. Reasoning Mode}
\begin{figure}[b]
\vskip -0.3in
    \centering
    \includegraphics[width=1\linewidth]{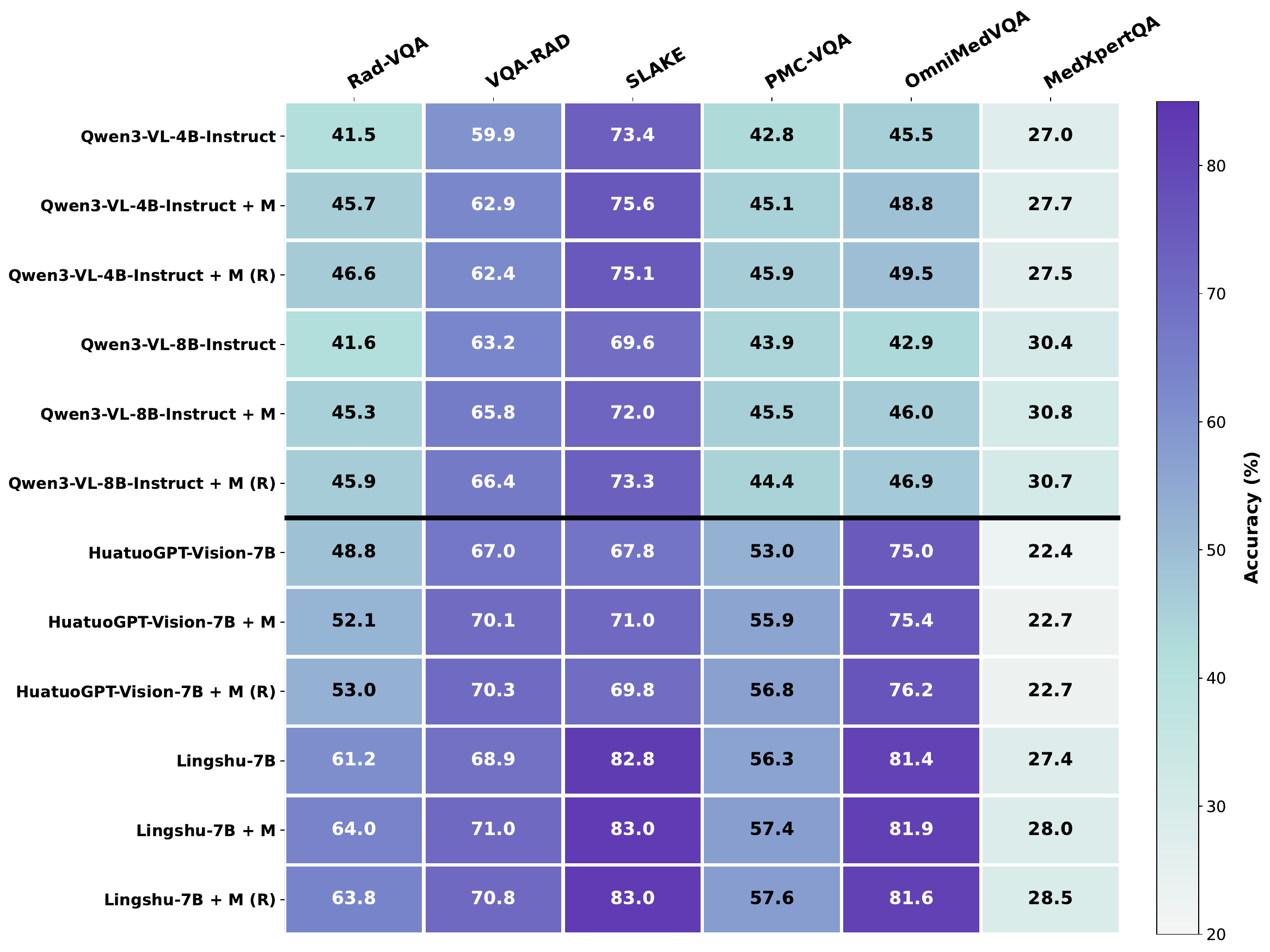}
    \caption{Performance comparison on six public benchmarks. Purple colors correspond to higher accuracy. Direct Mode (+M) and Reasoning Mode (+M (R)) show close performance.}
    \label{fig:heatmap}
\end{figure}

As illustrated in Figure~\ref{fig:heatmap}, the negligible performance gap between Direct Mode (+M) and Reasoning Mode (+M (R)) suggests that explicit CoT is not strictly necessary.
However, this similarity may partially stem from our reasoning structure reward ($\mathcal{R}_\text{{reason}}$) prioritizing structural constraints over logical validity. 

\subsection{Data Scaling Analysis}
To evaluate the scalability and data efficiency of our framework, we conducted a quantitative analysis across varying training set sizes.

\begin{figure}[b]
\vskip -0.2in
    \centering
    \includegraphics[width=1\linewidth]{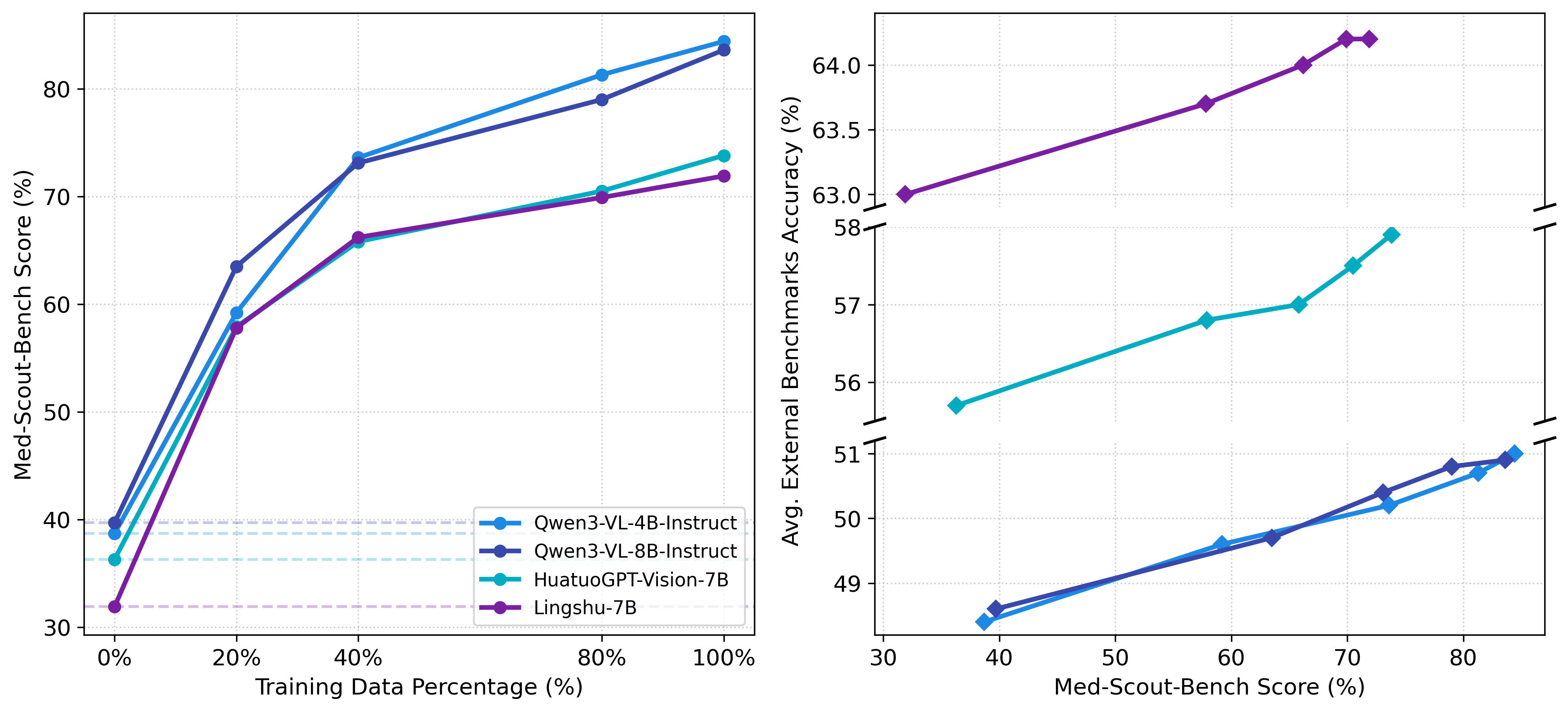}
    \caption{Data Scaling and Generalization Analysis. Left: Continuous performance improvement on Med-Scout-Bench with increasing training data. Right: Strong positive correlation between Med-Scout-Bench scores and average accuracy on external benchmarks.}
    \label{fig:data_scaling_effects}
\end{figure}

\noindent\textbf{Med-Scout-Bench Scalability.} As shown in Figure~\ref{fig:data_scaling_effects} (Left), performance on Med-Scout-Bench improves consistently across all backbone models as the training data volume increases from 20\% to 100\%. The substantial performance gain in the early stages underscores the high efficiency and quality of our automatically generated supervision signals, while the continuous upward trend without saturation suggests the models have not yet reached their capacity limits.

\noindent\textbf{Generalization Correlation.} Figure~\ref{fig:data_scaling_effects} (Right) reveals a strong positive correlation between internal alignment scores and performance on six external benchmarks. This confirms that improving geometric awareness enhances general medical perception. Consequently, Med-Scout-Bench serves as a reliable indicator of broader clinical visual reasoning capabilities.

\subsection{Comparison with Existing Proxy Tasks}
To rigorously evaluate the effectiveness of our proxy task design, we compare Med-Scout against existing visual proxy tasks using both Qwen3-VL-4B-Instruct and Qwen3-VL-8B-Instruct as backbones. While recent methods like Jigsaw-R1~\cite{DBLP:journals/tmlr/WangZTLXYB25} and ViCrit~\cite{DBLP:journals/corr/abs-2506-10128} successfully introduce geometric constraints for general-domain models, they fundamentally lack the medical specificity required to handle complex anatomical structures and subtle clinical anomalies. To ensure a strictly fair comparison, we intentionally restrict Med-Scout to use standard sparse rewards, disabling our advanced dense geometric reward (DGR) mechanism. As demonstrated in Table~\ref{tab:proxy_comparison}, even under this constrained setting, Med-Scout consistently achieves the highest average accuracy across both radiological VQA and broader medical generalization domains. This compellingly confirms that domain-specific geometric alignment, rather than mere general spatial awareness, is essential for robust medical perception.

\begin{table}[h]
\centering
\caption{Comparison with existing proxy tasks using Qwen3-VL-Instruct models. We report the average accuracy for Radiological VQA and Generalization benchmarks. Even with sparse rewards, Med-Scout outperforms general-domain tasks.}
\label{tab:proxy_comparison}
\begin{center}
\begin{small}
\begin{sc}
\resizebox{\columnwidth}{!}{%
\begin{tabular}{lcccc}
\toprule
Method & Med. & Geo. & Rad-VQA (Avg.) & Gen. (Avg.) \\
\midrule
\rowcolor{headergray} \multicolumn{5}{l}{\textbf{Qwen3-VL-4B-Instruct}} \\
Baseline & - & - & 58.3 & 38.4 \\
\quad + Jigsaw-R1 & \ding{55} & \ding{51}  &
57.6{\scriptsize\textcolor{red}{$\downarrow$\,0.7}} &
38.3{\scriptsize\textcolor{red}{$\downarrow$\,0.1}} \\
\quad + ViCrit    & \ding{55} & \ding{55}  &
57.7{\scriptsize\textcolor{red}{$\downarrow$\,0.6}} &
38.4{\scriptsize\textcolor{blue}{$\uparrow$\,0.0}} \\
\rowcolor{color1} \quad + \textbf{Med-Scout} & 
\textbf{\ding{51}} & \textbf{\ding{51}} & 
60.8{\scriptsize\textcolor{green!40!black}{$\uparrow$\,2.5}} & 
40.2{\scriptsize\textcolor{green!40!black}{$\uparrow$\,1.8}} \\
\quad $\Delta$ & 
- & - & 
\color{green!40!black}{+3.2} & 
\color{green!40!black}{+1.9} \\
\midrule
\rowcolor{headergray} \multicolumn{5}{l}{\textbf{Qwen3-VL-8B-Instruct}} \\
Baseline & - & - & 58.1 & 39.1 \\
\quad + Jigsaw-R1 & \ding{55} & \ding{51}  &
57.6{\scriptsize\textcolor{red}{$\downarrow$\,0.7}} &
38.3{\scriptsize\textcolor{red}{$\downarrow$\,0.1}} \\
\quad + ViCrit    & \ding{55} & \ding{55}  &
57.7{\scriptsize\textcolor{red}{$\downarrow$\,0.6}} &
38.4{\scriptsize\textcolor{blue}{$\uparrow$\,0.0}} \\
\rowcolor{color2} \quad + \textbf{Med-Scout} & 
\textbf{\ding{51}} & \textbf{\ding{51}} & 
60.4{\scriptsize\textcolor{green!40!black}{$\uparrow$\,2.3}} & 
40.5{\scriptsize\textcolor{green!40!black}{$\uparrow$\,1.4}} \\
\quad $\Delta$ & 
- & - & 
\color{green!40!black}{+3.0} & 
\color{green!40!black}{+1.5} \\
\bottomrule
\end{tabular}%
}
\end{sc}
\end{small}
\end{center}
\vskip -0.1in
\end{table}

\subsection{Extensive Analysis}
We conducted an extensive analysis to evaluate the impact of Med-Scout on the spatial discrimination capabilities of MLLMs and their visual attention focus on target regions.
\begin{figure*}[t]
    \centering
    \includegraphics[width=1\linewidth]{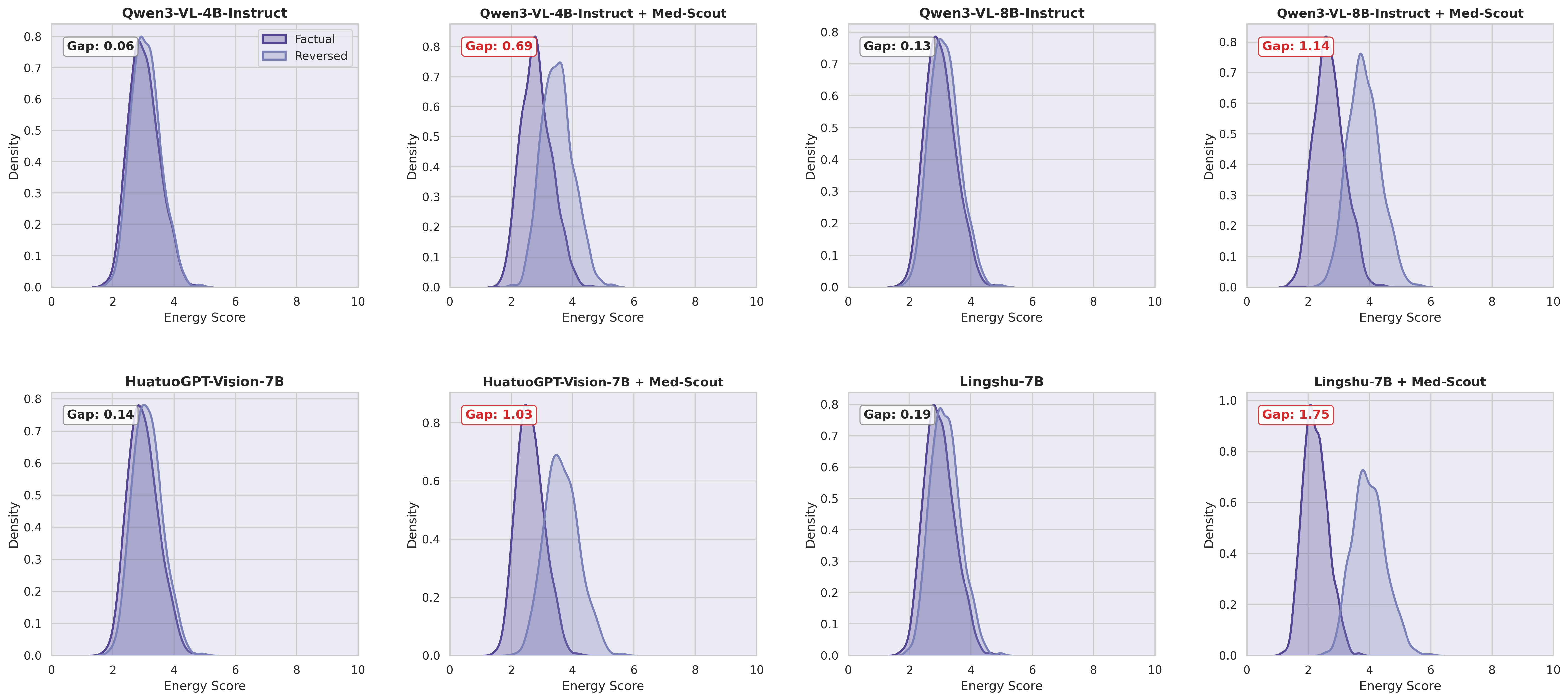}
    \caption{Energy Landscape of Factual Consistency. We visualize energy distributions for 800 factual (purple) versus spatially inverted (blue) report pairs. Med-Scout establishes a distinct energy barrier.}
    \label{fig:energy_landscape_comparison}
\end{figure*}

\begin{figure}[b]
 \vskip -0.2in
    \centering
    \includegraphics[width=1\linewidth]{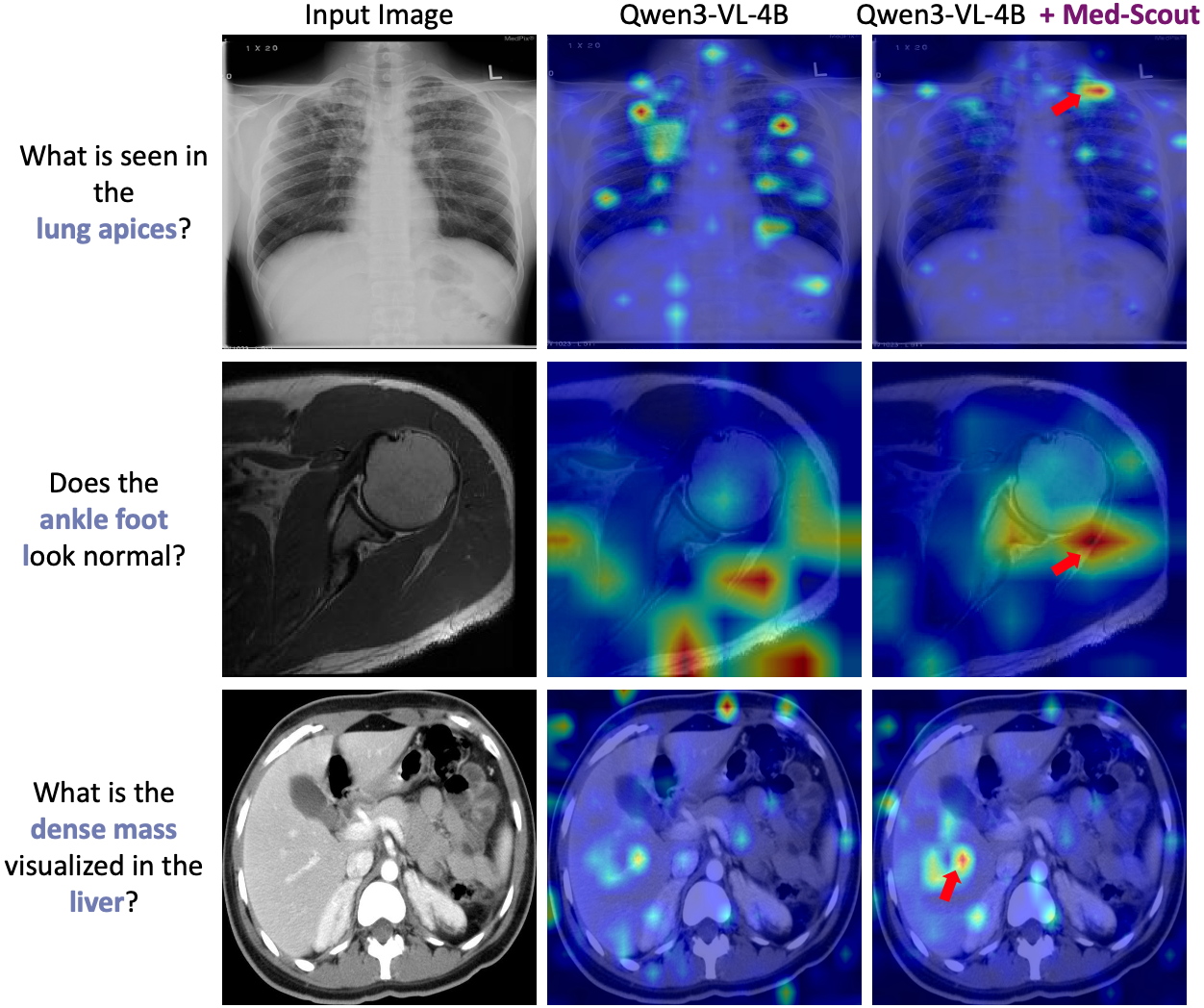}
    \caption{Visualization of attention maps on critical anatomical regions. We compare the visual attention of the baseline Qwen3-VL-4B-Instruct (Middle) versus the model aligned with Med-Scout (Right) given specific anatomical queries. Med-Scout demonstrates a highly concentrated focus on the critical target regions.}
    \label{fig:attention}
\end{figure}

\noindent\textbf{Energy Landscape of Factual Consistency.} 
Following the Energy-Based model~\cite{DBLP:journals/corr/abs-2101-03288}, we quantify the compatibility between visual evidence $\mathbf{x}$ and textual description $\mathbf{y}$ via the energy function $E(\mathbf{x}, \mathbf{y}) = -\log P_\theta(\mathbf{y} | \mathbf{x})$, implemented as the negative log-likelihood of the target response. 
To evaluate this, we constructed a probe dataset of 800 factual-counterfactual pairs from MIMIC-CXR by inverting anomaly-related spatial locatives in reports. 
As visualized in Figure~\ref{fig:energy_landscape_comparison}, the baseline models exhibit a collapsed landscape where factual and perturbed descriptions share overlapping energy distributions, indicating geometric blindness. 
In contrast, Med-Scout establishes a distinct energy barrier, effectively assigning high-energy states to spatial hallucinations while preserving low energy for factual descriptions. This separation suggests that Med-Scout has successfully internalized the spatial constraints of medical imagery, moving beyond mere language priors to achieve rigorous geometric alignment.


\noindent\textbf{Analysis of Attention on Critical Region.}
We followed~\cite{DBLP:conf/iclr/0002KCI25} and visualized the attention maps of the MLLM on specific anatomical queries (Figure~\ref{fig:attention}). The baseline model exhibits scattered attention and frequently drifts to irrelevant background noise. In contrast, Med-Scout demonstrates a highly concentrated focus on the critical regions, such as \textit{lung apices}, \textit{ankle joints}, or \textit{dense mass in liver}. This shift confirms that our challenging geometric proxy tasks motivate the model to transition from superficial scanning to fine-grained visual scrutiny, thus effectively enhancing the model's sensitivity to local visual evidence. This improved grounding not only translates directly to the external benchmarks' performance but also provides the reliability essential for medical understanding.

\section{Conclusion}
Motivated by a pilot study revealing the significant ``geometric blindness'' of MLLMs in medical perception, this paper proposes Med-Scout and Med-Scout-Bench, a geometry-aware RL post-training framework designed to cure this blindness and a novel benchmark aimed to quantify this limitation rigorously. By aligning semantic generation with geometric constraints through three intrinsic proxy tasks and a dense reward mechanism that stabilizes the optimization process, Med-Scout significantly enhances the geometric perception of existing MLLMs and further improves performance on radiological and comprehensive medical VQA benchmarks. Moreover, further analysis reveals the potential of this strategy in maintaining the geometric truth of medical images, thereby substantially enhancing visual capabilities.

\section*{Acknowledgments}
This work was supported by the Guangdong Basic and Applied Basic Research Foundation
(2026A1515011793), and the Youth S\&T Talent Support Programme of Guangdong Provincial Association for Science and Technology (SKXRC2025467), and the Transvascular Implantation Devices Research Institute (KY052025003). 

\section*{Impact Statement}
This work addresses a fundamental divergence in training priorities. While general MLLMs focus on linguistic fluency, medical applications require strict adherence to geometric constraints. We identify this mismatch as the root cause of ``geometric blindness,'' a critical deficit where models fail to ground their outputs in physical reality.

We resolve this bottleneck with Med-Scout, a cost-efficient post-training framework. By extracting intrinsic geometric rules directly from unlabeled data instead of relying on expensive expert annotations, we use RL to align models with visual logic. Crucially, we prove that curing this specific geometric deficit drives substantial improvements in downstream tasks, such as report generation and medical visual question answering. This establishes a scalable and data-efficient pathway toward reliable clinical AI that strictly respects the physical truth of medical images.




\bibliography{example_paper}

@article{DBLP:journals/corr/abs-2504-21051,
  author       = {Jiarui Ye and
                  Hao Tang},
  title        = {Multimodal Large Language Models for Medicine: {A} Comprehensive Survey},
  journal      = {CoRR},
  volume       = {abs/2504.21051},
  year         = {2025},
  url          = {https://doi.org/10.48550/arXiv.2504.21051}
        
        ,
  doi          = {10.48550/ARXIV.2504.21051},
  eprinttype    = {arXiv},
  eprint       = {2504.21051},
  bibsource    = {dblp computer science bibliography, https://dblp.org}
}

@article{DBLP:journals/corr/abs-2304-08485,
  author       = {Haotian Liu and
                  Chunyuan Li and
                  Qingyang Wu and
                  Yong Jae Lee},
  title        = {Visual Instruction Tuning},
  journal      = {CoRR},
  volume       = {abs/2304.08485},
  year         = {2023},
  url          = {https://doi.org/10.48550/arXiv.2304.08485},
  doi          = {10.48550/ARXIV.2304.08485},
  eprinttype    = {arXiv},
  eprint       = {2304.08485},
  bibsource    = {dblp computer science bibliography, https://dblp.org}
}

@article{DBLP:journals/corr/abs-2502-13923,
  author       = {Shuai Bai and
                  Keqin Chen and
                  Xuejing Liu and
                  Jialin Wang and
                  Wenbin Ge and
                  Sibo Song and
                  Kai Dang and
                  Peng Wang and
                  Shijie Wang and
                  Jun Tang and
                  Humen Zhong and
                  Yuanzhi Zhu and
                  Ming{-}Hsuan Yang and
                  Zhaohai Li and
                  Jianqiang Wan and
                  Pengfei Wang and
                  Wei Ding and
                  Zheren Fu and
                  Yiheng Xu and
                  Jiabo Ye and
                  Xi Zhang and
                  Tianbao Xie and
                  Zesen Cheng and
                  Hang Zhang and
                  Zhibo Yang and
                  Haiyang Xu and
                  Junyang Lin},
  title        = {{Q}wen2.5-{VL} Technical Report},
  journal      = {CoRR},
  volume       = {abs/2502.13923},
  year         = {2025},
  url          = {https://doi.org/10.48550/arXiv.2502.13923},
  doi          = {10.48550/ARXIV.2502.13923},
  eprinttype    = {arXiv},
  eprint       = {2502.13923},
  bibsource    = {dblp computer science bibliography, https://dblp.org}
}

@article{DBLP:journals/corr/abs-2511-21631,
  author       = {Shuai Bai and
                  Yuxuan Cai and
                  Ruizhe Chen and
                  Keqin Chen and
                  Xionghui Chen and
                  Zesen Cheng and
                  Lianghao Deng and
                  Wei Ding and
                  Chang Gao and
                  Chunjiang Ge and
                  Wenbin Ge and
                  Zhifang Guo and
                  Qidong Huang and
                  Jie Huang and
                  Fei Huang and
                  Binyuan Hui and
                  Shutong Jiang and
                  Zhaohai Li and
                  Mingsheng Li and
                  Mei Li and
                  Kaixin Li and
                  Zicheng Lin and
                  Junyang Lin and
                  Xuejing Liu and
                  Jiawei Liu and
                  Chenglong Liu and
                  Yang Liu and
                  Dayiheng Liu and
                  Shixuan Liu and
                  Dunjie Lu and
                  Ruilin Luo and
                  Chenxu Lv and
                  Rui Men and
                  Lingchen Meng and
                  Xuancheng Ren and
                  Xingzhang Ren and
                  Sibo Song and
                  Yuchong Sun and
                  Jun Tang and
                  Jianhong Tu and
                  Jianqiang Wan and
                  Peng Wang and
                  Pengfei Wang and
                  Qiuyue Wang and
                  Yuxuan Wang and
                  Tianbao Xie and
                  Yiheng Xu and
                  Haiyang Xu and
                  Jin Xu and
                  Zhibo Yang and
                  Mingkun Yang and
                  Jianxin Yang and
                  An Yang and
                  Bowen Yu and
                  Fei Zhang and
                  Hang Zhang and
                  Xi Zhang and
                  Bo Zheng and
                  Humen Zhong and
                  Jingren Zhou and
                  Fan Zhou and
                  Jing Zhou and
                  Yuanzhi Zhu and
                  Ke Zhu},
  title        = {{Q}wen3-{VL} Technical Report},
  journal      = {CoRR},
  volume       = {abs/2511.21631},
  year         = {2025},
  url          = {https://doi.org/10.48550/arXiv.2511.21631},
  doi          = {10.48550/ARXIV.2511.21631},
  eprinttype    = {arXiv},
  eprint       = {2511.21631},
  bibsource    = {dblp computer science bibliography, https://dblp.org}
}

@article{DBLP:journals/corr/abs-2406-19280,
  author       = {Junying Chen and
                  Ruyi Ouyang and
                  Anningzhe Gao and
                  Shunian Chen and
                  Guiming Hardy Chen and
                  Xidong Wang and
                  Ruifei Zhang and
                  Zhenyang Cai and
                  Ke Ji and
                  Guangjun Yu and
                  Xiang Wan and
                  Benyou Wang},
  title        = {Huatuo{GPT}-{V}ision, Towards Injecting Medical Visual Knowledge into
                  Multimodal {LLM}s at Scale},
  journal      = {CoRR},
  volume       = {abs/2406.19280},
  year         = {2024},
  url          = {https://doi.org/10.48550/arXiv.2406.19280},
  doi          = {10.48550/ARXIV.2406.19280},
  eprinttype    = {arXiv},
  eprint       = {2406.19280},
  bibsource    = {dblp computer science bibliography, https://dblp.org}
}

@article{DBLP:journals/corr/abs-2507-05201,
  author       = {Andrew Sellergren and
                  Sahar Kazemzadeh and
                  Tiam Jaroensri and
                  Atilla P. Kiraly and
                  Madeleine Traverse and
                  Timo Kohlberger and
                  Shawn Xu and
                  Fayaz Jamil and
                  C{\'{\i}}an Hughes and
                  Charles Lau and
                  Justin Chen and
                  Fereshteh Mahvar and
                  Liron Yatziv and
                  Tiffany L. Chen and
                  Bram Sterling and
                  Stefanie Anna Baby and
                  Susanna Maria Baby and
                  Jeremy Lai and
                  Samuel Schmidgall and
                  Lu Yang and
                  Kejia Chen and
                  Per Bjornsson and
                  Shashir Reddy and
                  Ryan Brush and
                  Kenneth Philbrick and
                  Mercy Asiedu and
                  Ines Mezerreg and
                  Howard Hu and
                  Howard Yang and
                  Richa Tiwari and
                  Sunny Jansen and
                  Preeti Singh and
                  Yun Liu and
                  Shekoofeh Azizi and
                  Aishwarya Kamath and
                  Johan Ferret and
                  Shreya Pathak and
                  Nino Vieillard and
                  Ramona Merhej and
                  Sarah Perrin and
                  Tatiana Matejovicova and
                  Alexandre Ram{\'{e}} and
                  Morgane Rivi{\`{e}}re and
                  Louis Rouillard and
                  Thomas Mesnard and
                  Geoffrey Cideron and
                  Jean{-}Bastien Grill and
                  Sabela Ramos and
                  Edouard Yvinec and
                  Michelle Casbon and
                  Elena Buchatskaya and
                  Jean{-}Baptiste Alayrac and
                  Dmitry Lepikhin and
                  Vlad Feinberg and
                  Sebastian Borgeaud and
                  Alek Andreev and
                  Cassidy Hardin and
                  Robert Dadashi and
                  L{\'{e}}onard Hussenot and
                  Armand Joulin and
                  Olivier Bachem and
                  Yossi Matias and
                  Katherine Chou and
                  Avinatan Hassidim and
                  Kavi Goel and
                  Cl{\'{e}}ment Farabet and
                  Joelle K. Barral and
                  Tris Warkentin and
                  Jonathon Shlens and
                  David J. Fleet and
                  Victor Cotruta and
                  Omar Sanseviero and
                  Gus Martins and
                  Phoebe Kirk and
                  Anand Rao and
                  Shravya Shetty and
                  David F. Steiner and
                  Can Kirmizibayrak and
                  Rory Pilgrim and
                  Daniel Golden and
                  Lin Yang},
  title        = {Med{G}emma Technical Report},
  journal      = {CoRR},
  volume       = {abs/2507.05201},
  year         = {2025},
  url          = {https://doi.org/10.48550/arXiv.2507.05201},
  doi          = {10.48550/ARXIV.2507.05201},
  eprinttype    = {arXiv},
  eprint       = {2507.05201},
  bibsource    = {dblp computer science bibliography, https://dblp.org}
}

@article{DBLP:journals/corr/abs-2506-07044,
  author       = {LASA Team and
                  Weiwen Xu and
                  Hou Pong Chan and
                  Long Li and
                  Mahani Aljunied and
                  Ruifeng Yuan and
                  Jianyu Wang and
                  Chenghao Xiao and
                  Guizhen Chen and
                  Chaoqun Liu and
                  Zhaodonghui Li and
                  Yu Sun and
                  Junao Shen and
                  Chaojun Wang and
                  Jie Tan and
                  Deli Zhao and
                  Tingyang Xu and
                  Hao Zhang and
                  Yu Rong},
  title        = {Lingshu: {A} Generalist Foundation Model for Unified Multimodal Medical
                  Understanding and Reasoning},
  journal      = {CoRR},
  volume       = {abs/2506.07044},
  year         = {2025},
  url          = {https://doi.org/10.48550/arXiv.2506.07044},
  doi          = {10.48550/ARXIV.2506.07044},
  eprinttype    = {arXiv},
  eprint       = {2506.07044},
  bibsource    = {dblp computer science bibliography, https://dblp.org}
}

@article{DBLP:journals/corr/abs-2412-05265,
  author       = {Kevin P. Murphy},
  title        = {Reinforcement Learning: An Overview},
  journal      = {CoRR},
  volume       = {abs/2412.05265},
  year         = {2024},
  url          = {https://doi.org/10.48550/arXiv.2412.05265},
  doi          = {10.48550/ARXIV.2412.05265},
  eprinttype    = {arXiv},
  eprint       = {2412.05265},
  bibsource    = {dblp computer science bibliography, https://dblp.org}
}

@article{DBLP:journals/corr/SchulmanWDRK17,
  author       = {John Schulman and
                  Filip Wolski and
                  Prafulla Dhariwal and
                  Alec Radford and
                  Oleg Klimov},
  title        = {Proximal Policy Optimization Algorithms},
  journal      = {CoRR},
  volume       = {abs/1707.06347},
  year         = {2017},
  url          = {http://arxiv.org/abs/1707.06347},
  eprinttype    = {arXiv},
  eprint       = {1707.06347},
  bibsource    = {dblp computer science bibliography, https://dblp.org}
}

@article{DBLP:journals/corr/abs-2402-03300,
  author       = {Zhihong Shao and
                  Peiyi Wang and
                  Qihao Zhu and
                  Runxin Xu and
                  Junxiao Song and
                  Mingchuan Zhang and
                  Y. K. Li and
                  Y. Wu and
                  Daya Guo},
  title        = {DeepSeekMath: Pushing the Limits of Mathematical Reasoning in Open
                  Language Models},
  journal      = {CoRR},
  volume       = {abs/2402.03300},
  year         = {2024},
  url          = {https://doi.org/10.48550/arXiv.2402.03300},
  doi          = {10.48550/ARXIV.2402.03300},
  eprinttype    = {arXiv},
  eprint       = {2402.03300},
  bibsource    = {dblp computer science bibliography, https://dblp.org}
}

@misc{butsanets2025radimagenetvqalargescalectmri,
      title={{RadImageNet}-{VQA}: A Large-Scale {CT} and {MRI} Dataset for Radiologic Visual Question Answering}, 
      author={Léo Butsanets and Charles Corbière and Julien Khlaut and Pierre Manceron and Corentin Dancette},
      year={2025},
      eprint={2512.17396},
      archivePrefix={arXiv},
      primaryClass={cs.CV},
      url={https://arxiv.org/abs/2512.17396}, 
}

@article{lau2018dataset,
  title={A dataset of clinically generated visual questions and answers about radiology images},
  author={Lau, Jason J and Gayen, Soumya and Ben Abacha, Asma and Demner-Fushman, Dina},
  journal={Scientific data},
  volume={5},
  number={1},
  pages={1--10},
  year={2018},
  publisher={Nature Publishing Group}
}

@inproceedings{DBLP:conf/isbi/LiuZXMYW21,
  author       = {Bo Liu and
                  Li{-}Ming Zhan and
                  Li Xu and
                  Lin Ma and
                  Yan Yang and
                  Xiao{-}Ming Wu},
  title        = {Slake: {A} Semantically-Labeled Knowledge-Enhanced Dataset For Medical
                  Visual Question Answering},
  booktitle    = {18th {IEEE} International Symposium on Biomedical Imaging, {ISBI}
                  2021, Nice, France, April 13-16, 2021},
  pages        = {1650--1654},
  publisher    = {{IEEE}},
  year         = {2021},
  url          = {https://doi.org/10.1109/ISBI48211.2021.9434010},
  doi          = {10.1109/ISBI48211.2021.9434010},
  bibsource    = {dblp computer science bibliography, https://dblp.org}
}

@article{DBLP:journals/corr/abs-2305-10415,
  author       = {Xiaoman Zhang and
                  Chaoyi Wu and
                  Ziheng Zhao and
                  Weixiong Lin and
                  Ya Zhang and
                  Yanfeng Wang and
                  Weidi Xie},
  title        = {{PMC-VQA:} Visual Instruction Tuning for Medical Visual Question Answering},
  journal      = {CoRR},
  volume       = {abs/2305.10415},
  year         = {2023},
  url          = {https://doi.org/10.48550/arXiv.2305.10415},
  doi          = {10.48550/ARXIV.2305.10415},
  eprinttype    = {arXiv},
  eprint       = {2305.10415},
  bibsource    = {dblp computer science bibliography, https://dblp.org}
}

@inproceedings{DBLP:conf/cvpr/HuLLSHQL24,
  author       = {Yutao Hu and
                  Tianbin Li and
                  Quanfeng Lu and
                  Wenqi Shao and
                  Junjun He and
                  Yu Qiao and
                  Ping Luo},
  title        = {Omni{M}ed{VQA}: {A} New Large-Scale Comprehensive Evaluation Benchmark
                  for Medical {LVLM}},
  booktitle    = {{IEEE/CVF} Conference on Computer Vision and Pattern Recognition,
                  {CVPR} 2024, Seattle, WA, USA, June 16-22, 2024},
  pages        = {22170--22183},
  publisher    = {{IEEE}},
  year         = {2024},
  url          = {https://doi.org/10.1109/CVPR52733.2024.02093},
  doi          = {10.1109/CVPR52733.2024.02093
        
        
        
        },
  bibsource    = {dblp computer science bibliography, https://dblp.org}
}

@inproceedings{DBLP:conf/icml/ZuoQLCZHZ0025,
  author       = {Yuxin Zuo and
                  Shang Qu and
                  Yifei Li and
                  Zhang{-}Ren Chen and
                  Xuekai Zhu and
                  Ermo Hua and
                  Kaiyan Zhang and
                  Ning Ding and
                  Bowen Zhou},
  title        = {Med{X}pert{QA}: Benchmarking Expert-Level Medical Reasoning and Understanding},
  booktitle    = {Forty-second International Conference on Machine Learning, {ICML}
                  2025, Vancouver, BC, Canada, July 13-19, 2025},
  publisher    = {OpenReview.net},
  year         = {2025},
  url          = {https://openreview.net/forum?id=IyVcxU0RKI},
  bibsource    = {dblp computer science bibliography, https://dblp.org}
}

@article{johnson2019mimic,
  title={{MIMIC-CXR}, a de-identified publicly available database of chest radiographs with free-text reports},
  author={Johnson, Alistair EW and Pollard, Tom J and Berkowitz, Seth J and Greenbaum, Nathaniel R and Lungren, Matthew P and Deng, Chih-ying and Mark, Roger G and Horng, Steven},
  journal={Scientific data},
  volume={6},
  number={1},
  pages={317},
  year={2019},
  publisher={Nature Publishing Group UK London}
}

@article{demner2015preparing,
  title={Preparing a collection of radiology examinations for distribution and retrieval},
  author={Demner-Fushman, Dina and Kohli, Marc D and Rosenman, Marc B and Shooshan, Sonya E and Rodriguez, Laritza and Antani, Sameer and Thoma, George R and McDonald, Clement J},
  journal={Journal of the American Medical Informatics Association},
  volume={23},
  number={2},
  pages={304--310},
  year={2015},
  publisher={Oxford Academic}
}

@inproceedings{DBLP:conf/nips/LiWZULYNPG23,
  author       = {Chunyuan Li and
                  Cliff Wong and
                  Sheng Zhang and
                  Naoto Usuyama and
                  Haotian Liu and
                  Jianwei Yang and
                  Tristan Naumann and
                  Hoifung Poon and
                  Jianfeng Gao},
  editor       = {Alice Oh and
                  Tristan Naumann and
                  Amir Globerson and
                  Kate Saenko and
                  Moritz Hardt and
                  Sergey Levine},
  title        = {L{L}a{VA-M}ed: Training a Large Language-and-Vision Assistant for Biomedicine
                  in One Day},
  booktitle    = {Advances in Neural Information Processing Systems 36: Annual Conference
                  on Neural Information Processing Systems 2023, NeurIPS 2023, New Orleans,
                  LA, USA, December 10 - 16, 2023},
  year         = {2023},
  bibsource    = {dblp computer science bibliography, https://dblp.org}
}

@article{DBLP:journals/corr/abs-2508-18265,
  author       = {Weiyun Wang and
                  Zhangwei Gao and
                  Lixin Gu and
                  Hengjun Pu and
                  Long Cui and
                  Xingguang Wei and
                  Zhaoyang Liu and
                  Linglin Jing and
                  Shenglong Ye and
                  Jie Shao and
                  Zhaokai Wang and
                  Zhe Chen and
                  Hongjie Zhang and
                  Ganlin Yang and
                  Haomin Wang and
                  Qi Wei and
                  Jinhui Yin and
                  Wenhao Li and
                  Erfei Cui and
                  Guanzhou Chen and
                  Zichen Ding and
                  Changyao Tian and
                  Zhenyu Wu and
                  JingJing Xie and
                  Zehao Li and
                  Bowen Yang and
                  Yuchen Duan and
                  Xuehui Wang and
                  Zhi Hou and
                  Haoran Hao and
                  Tianyi Zhang and
                  Songze Li and
                  Xiangyu Zhao and
                  Haodong Duan and
                  Nianchen Deng and
                  Bin Fu and
                  Yinan He and
                  Yi Wang and
                  Conghui He and
                  Botian Shi and
                  Junjun He and
                  Yingtong Xiong and
                  Han Lv and
                  Lijun Wu and
                  Wenqi Shao and
                  Kaipeng Zhang and
                  Huipeng Deng and
                  Biqing Qi and
                  Jiaye Ge and
                  Qipeng Guo and
                  Wenwei Zhang and
                  Songyang Zhang and
                  Maosong Cao and
                  Junyao Lin and
                  Kexian Tang and
                  Jianfei Gao and
                  Haian Huang and
                  Yuzhe Gu and
                  Chengqi Lyu and
                  Huanze Tang and
                  Rui Wang and
                  Haijun Lv and
                  Wanli Ouyang and
                  Limin Wang and
                  Min Dou and
                  Xizhou Zhu and
                  Tong Lu and
                  Dahua Lin and
                  Jifeng Dai and
                  Weijie Su and
                  Bowen Zhou and
                  Kai Chen and
                  Yu Qiao and
                  Wenhai Wang and
                  Gen Luo},
  title        = {Intern{VL}3.5: Advancing Open-Source Multimodal Models in Versatility,
                  Reasoning, and Efficiency},
  journal      = {CoRR},
  volume       = {abs/2508.18265},
  year         = {2025},
  url          = {https://doi.org/10.48550/arXiv.2508.18265},
  doi          = {10.48550/ARXIV.2508.18265},
  eprinttype    = {arXiv},
  eprint       = {2508.18265},
  bibsource    = {dblp computer science bibliography, https://dblp.org}
}

@article{DBLP:journals/corr/abs-2312-14238,
  author       = {Zhe Chen and
                  Jiannan Wu and
                  Wenhai Wang and
                  Weijie Su and
                  Guo Chen and
                  Sen Xing and
                  Muyan Zhong and
                  Qinglong Zhang and
                  Xizhou Zhu and
                  Lewei Lu and
                  Bin Li and
                  Ping Luo and
                  Tong Lu and
                  Yu Qiao and
                  Jifeng Dai},
  title        = {Intern{VL}: Scaling up Vision Foundation Models and Aligning for Generic
                  Visual-Linguistic Tasks},
  journal      = {CoRR},
  volume       = {abs/2312.14238},
  year         = {2023},
  url          = {https://doi.org/10.48550/arXiv.2312.14238},
  doi          = {10.48550/ARXIV.2312.14238},
  eprinttype    = {arXiv},
  eprint       = {2312.14238},
  bibsource    = {dblp computer science bibliography, https://dblp.org}
}

@article{DBLP:journals/corr/abs-2305-17100,
  author       = {Kai Zhang and
                  Jun Yu and
                  Zhiling Yan and
                  Yixin Liu and
                  Eashan Adhikarla and
                  Sunyang Fu and
                  Xun Chen and
                  Chen Chen and
                  Yuyin Zhou and
                  Xiang Li and
                  Lifang He and
                  Brian D. Davison and
                  Quanzheng Li and
                  Yong Chen and
                  Hongfang Liu and
                  Lichao Sun},
  title        = {Biomed{GPT}: {A} Unified and Generalist Biomedical Generative Pre-trained
                  Transformer for Vision, Language, and Multimodal Tasks},
  journal      = {CoRR},
  volume       = {abs/2305.17100},
  year         = {2023},
  url          = {https://doi.org/10.48550/arXiv.2305.17100},
  doi          = {10.48550/ARXIV.2305.17100},
  eprinttype    = {arXiv},
  eprint       = {2305.17100},
  bibsource    = {dblp computer science bibliography, https://dblp.org}
}

@inproceedings{DBLP:conf/miccai/PanLWLZLCOR25,
  author       = {Jiazhen Pan and
                  Che Liu and
                  Junde Wu and
                  Fenglin Liu and
                  Jiayuan Zhu and
                  Hongwei Bran Li and
                  Chen Chen and
                  Cheng Ouyang and
                  Daniel Rueckert},
  editor       = {James C. Gee and
                  Daniel C. Alexander and
                  Jaesung Hong and
                  Juan Eugenio Iglesias and
                  Carole H. Sudre and
                  Archana Venkataraman and
                  Polina Golland and
                  Jong Hyo Kim and
                  Jinah Park},
  title        = {Med{VLM-R1}: Incentivizing Medical Reasoning Capability of Vision-Language
                  Models ({VLM}s) via Reinforcement Learning},
  booktitle    = {Medical Image Computing and Computer Assisted Intervention - {MICCAI}
                  2025 - 28th International Conference, Daejeon, South Korea, September
                  23-27, 2025, Proceedings, Part {VII}},
  series       = {Lecture Notes in Computer Science},
  volume       = {15966},
  pages        = {337--347},
  publisher    = {Springer},
  year         = {2025},
  url          = {https://doi.org/10.1007/978-3-032-04981-0\_32},
  doi          = {10.1007/978-3-032-04981-0\_32},
  bibsource    = {dblp computer science bibliography, https://dblp.org}
}

@article{zhang2023biomedclip,
  title={Biomedclip: a multimodal biomedical foundation model pretrained from fifteen million scientific image-text pairs},
  author={Zhang, Sheng and Xu, Yanbo and Usuyama, Naoto and Xu, Hanwen and Bagga, Jaspreet and Tinn, Robert and Preston, Sam and Rao, Rajesh and Wei, Mu and Valluri, Naveen and others},
  journal={arXiv preprint arXiv:2303.00915
        
        
        
        
        
        
        
        
        
        },
  year={2023}
}

@article{DBLP:journals/tmlr/WangZTLXYB25,
  author       = {Zifu Wang and
                  Junyi Zhu and
                  Bo Tang and
                  Zhiyu Li and
                  Feiyu Xiong and
                  Jiaqian Yu and
                  Matthew B. Blaschko},
  title        = {Jigsaw-{R1}: {A} Study of Rule-based Visual Reinforcement Learning with
                  Jigsaw Puzzles},
  journal      = {Trans. Mach. Learn. Res.},
  volume       = {2025},
  year         = {2025},
  url          = {https://openreview.net/forum?id=XqQCsuyPve},
  bibsource    = {dblp computer science bibliography, https://dblp.org}
}

@article{DBLP:journals/corr/abs-2509-25190,
  author       = {Penghao Wu and
                  Yushan Zhang and
                  Haiwen Diao and
                  Bo Li and
                  Lewei Lu and
                  Ziwei Liu},
  title        = {Visual Jigsaw Post-Training Improves {MLLM}s},
  journal      = {CoRR},
  volume       = {abs/2509.25190},
  year         = {2025},
  url          = {https://doi.org/10.48550/arXiv.2509.25190},
  doi          = {10.48550/ARXIV.2509.25190},
  eprinttype    = {arXiv},
  eprint       = {2509.25190},
  bibsource    = {dblp computer science bibliography, https://dblp.org}
}

@article{DBLP:journals/corr/abs-2506-10128,
  author       = {Xiyao Wang and
                  Zhengyuan Yang and
                  Chao Feng and
                  Yongyuan Liang and
                  Yuhang Zhou and
                  Xiaoyu Liu and
                  Ziyi Zang and
                  Ming Li and
                  Chung{-}Ching Lin and
                  Kevin Lin and
                  Linjie Li and
                  Furong Huang and
                  Lijuan Wang},
  title        = {Vi{C}rit: {A} Verifiable Reinforcement Learning Proxy Task for Visual
                  Perception in {VLM}s},
  journal      = {CoRR},
  volume       = {abs/2506.10128},
  year         = {2025},
  url          = {https://doi.org/10.48550/arXiv.2506.10128},
  doi          = {10.48550/ARXIV.2506.10128},
  eprinttype    = {arXiv},
  eprint       = {2506.10128},
  bibsource    = {dblp computer science bibliography, https://dblp.org}
}

@inproceedings{DBLP:conf/nips/Wei0SBIXCLZ22,
  author       = {Jason Wei and
                  Xuezhi Wang and
                  Dale Schuurmans and
                  Maarten Bosma and
                  Brian Ichter and
                  Fei Xia and
                  Ed H. Chi and
                  Quoc V. Le and
                  Denny Zhou},
  editor       = {Sanmi Koyejo and
                  S. Mohamed and
                  A. Agarwal and
                  Danielle Belgrave and
                  K. Cho and
                  A. Oh},
  title        = {Chain-of-Thought Prompting Elicits Reasoning in Large Language Models},
  booktitle    = {Advances in Neural Information Processing Systems 35: Annual Conference
                  on Neural Information Processing Systems 2022, NeurIPS 2022, New Orleans,
                  LA, USA, November 28 - December 9, 2022},
  year         = {2022},
  bibsource    = {dblp computer science bibliography, https://dblp.org}
}

@article{wasserthal2023totalsegmentator,
  title={TotalSegmentator: robust segmentation of 104 anatomic structures in {CT} images},
  author={Wasserthal, Jakob and Breit, Hanns-Christian and Meyer, Manfred T and Pradella, Maurice and Hinck, Daniel and Sauter, Alexander W and Heye, Tobias and Boll, Daniel T and Cyriac, Joshy and Yang, Shan and others},
  journal={Radiology: Artificial Intelligence},
  volume={5},
  number={5},
  pages={e230024},
  year={2023},
  publisher={Radiological Society of North America}
}

@article{akinci2025totalsegmentator,
  title={Totalsegmentator {mri}: Robust sequence-independent segmentation of multiple anatomic structures in {mri}},
  author={Akinci D’Antonoli, Tugba and Berger, Lucas K and Indrakanti, Ashraya K and Vishwanathan, Nathan and Weiss, Jakob and Jung, Matthias and Berkarda, Zeynep and Rau, Alexander and Reisert, Marco and K{\"u}stner, Thomas and others},
  journal={Radiology},
  volume={314},
  number={2},
  pages={e241613},
  year={2025},
  publisher={Radiological Society of North America}
}

@article{DBLP:journals/corr/abs-2411-15594,
  author       = {Jiawei Gu and
                  Xuhui Jiang and
                  Zhichao Shi and
                  Hexiang Tan and
                  Xuehao Zhai and
                  Chengjin Xu and
                  Wei Li and
                  Yinghan Shen and
                  Shengjie Ma and
                  Honghao Liu and
                  Yuanzhuo Wang and
                  Jian Guo},
  title        = {A Survey on LLM-as-a-Judge},
  journal      = {CoRR},
  volume       = {abs/2411.15594},
  year         = {2024},
  url          = {https://doi.org/10.48550/arXiv.2411.15594},
  doi          = {10.48550/ARXIV.2411.15594},
  eprinttype    = {arXiv},
  eprint       = {2411.15594},
  bibsource    = {dblp computer science bibliography, https://dblp.org}
}

@article{DBLP:journals/corr/abs-2504-10479,
  author       = {Jinguo Zhu and
                  Weiyun Wang and
                  Zhe Chen and
                  Zhaoyang Liu and
                  Shenglong Ye and
                  Lixin Gu and
                  Hao Tian and
                  Yuchen Duan and
                  Weijie Su and
                  Jie Shao and
                  Zhangwei Gao and
                  Erfei Cui and
                  Xuehui Wang and
                  Yue Cao and
                  Yangzhou Liu and
                  Xingguang Wei and
                  Hongjie Zhang and
                  Haomin Wang and
                  Weiye Xu and
                  Hao Li and
                  Jiahao Wang and
                  Nianchen Deng and
                  Songze Li and
                  Yinan He and
                  Tan Jiang and
                  Jiapeng Luo and
                  Yi Wang and
                  Conghui He and
                  Botian Shi and
                  Xingcheng Zhang and
                  Wenqi Shao and
                  Junjun He and
                  Yingtong Xiong and
                  Wenwen Qu and
                  Peng Sun and
                  Penglong Jiao and
                  Han Lv and
                  Lijun Wu and
                  Kaipeng Zhang and
                  Huipeng Deng and
                  Jiaye Ge and
                  Kai Chen and
                  Limin Wang and
                  Min Dou and
                  Lewei Lu and
                  Xizhou Zhu and
                  Tong Lu and
                  Dahua Lin and
                  Yu Qiao and
                  Jifeng Dai and
                  Wenhai Wang},
  title        = {Intern{VL3}: Exploring Advanced Training and Test-Time Recipes for Open-Source
                  Multimodal Models},
  journal      = {CoRR},
  volume       = {abs/2504.10479},
  year         = {2025},
  url          = {https://doi.org/10.48550/arXiv.2504.10479
        
        }
        
        
        
        
        
        ,
  doi          = {10.48550/ARXIV.2504.10479}
        
        ,
  eprinttype    = {arXiv},
  eprint       = {2504.10479},
  bibsource    = {dblp computer science bibliography, https://dblp.org}
}

@inproceedings{lin2004rouge,
  title={Rouge: A package for automatic evaluation of summaries},
  author={Lin, Chin-Yew},
  booktitle={Text summarization branches out},
  pages={74--81},
  year={2004}
}

@inproceedings{DBLP:conf/cvpr/VedantamZP15,
  author       = {Ramakrishna Vedantam and
                  C. Lawrence Zitnick and
                  Devi Parikh},
  title        = {{CIDE}r: Consensus-based image description evaluation},
  booktitle    = {{IEEE} Conference on Computer Vision and Pattern Recognition, {CVPR}
                  2015, Boston, MA, USA, June 7-12, 2015},
  pages        = {4566--4575},
  publisher    = {{IEEE} Computer Society},
  year         = {2015},
  url          = {https://doi.org/10.1109/CVPR.2015.7299087}
        
        
        
        
        
        
        
        ,
  doi          = {10.1109/CVPR.2015.7299087},
  bibsource    = {dblp computer science bibliography, https://dblp.org}
}

@article{DBLP:journals/corr/abs-2401-17072,
  author       = {Ansar Aynetdinov and
                  Alan Akbik},
  title        = {SemScore: Automated Evaluation of Instruction-Tuned {LLM}s based on
                  Semantic Textual Similarity},
  journal      = {CoRR},
  volume       = {abs/2401.17072},
  year         = {2024},
  url          = {https://doi.org/10.48550/arXiv.2401.17072}
        
        
        
        
        
        ,
  doi          = {10.48550/ARXIV.2401.17072}
        
        
        
        ,
  eprinttype    = {arXiv},
  eprint       = {2401.17072},
  bibsource    = {dblp computer science bibliography, https://dblp.org}
}

@article{DBLP:journals/corr/abs-2101-03288,
  author       = {Yang Song and
                  Diederik P. Kingma},
  title        = {How to Train Your Energy-Based Models},
  journal      = {CoRR},
  volume       = {abs/2101.03288},
  year         = {2021},
  url          = {https://arxiv.org/abs/2101.03288},
  eprinttype    = {arXiv},
  eprint       = {2101.03288},
  bibsource    = {dblp computer science bibliography, https://dblp.org}
}

@inproceedings{DBLP:conf/iclr/0002KCI25,
  author       = {Jiarui Zhang and
                  Mahyar Khayatkhoei and
                  Prateek Chhikara and
                  Filip Ilievski},
  title        = {{MLLM}s Know Where to Look: Training-free Perception of Small Visual
                  Details with Multimodal {LLM}s},
  booktitle    = {The Thirteenth International Conference on Learning Representations,
                  {ICLR} 2025, Singapore, April 24-28, 2025},
  publisher    = {OpenReview.net},
  year         = {2025},
  url          = {https://openreview.net/forum?id=DgaY5mDdmT},
  bibsource    = {dblp computer science bibliography, https://dblp.org}
}

@article{DBLP:journals/corr/abs-2412-08737,
  author       = {Jiarui Zhang and
                  Ollie Liu and
                  Tianyu Yu and
                  Jinyi Hu and
                  Willie Neiswanger},
  title        = {Euclid: Supercharging Multimodal LLMs with Synthetic High-Fidelity
                  Visual Descriptions},
  journal      = {CoRR},
  volume       = {abs/2412.08737},
  year         = {2024},
  url          = {https://doi.org/10.48550/arXiv.2412.08737}
        
        
        
        
        
        ,
  doi          = {10.48550/ARXIV.2412.08737}
        
        ,
  eprinttype    = {arXiv},
  eprint       = {2412.08737},
  bibsource    = {dblp computer science bibliography, https://dblp.org}
}

@inproceedings{DBLP:conf/emnlp/CaiBGZSZ24,
  author       = {Shihao Cai and
                  Keqin Bao and
                  Hangyu Guo and
                  Jizhi Zhang and
                  Jun Song and
                  Bo Zheng},
  editor       = {Yaser Al{-}Onaizan and
                  Mohit Bansal and
                  Yun{-}Nung Chen},
  title        = {GeoGPT4V: Towards Geometric Multi-modal Large Language Models with
                  Geometric Image Generation},
  booktitle    = {Proceedings of the 2024 Conference on Empirical Methods in Natural
                  Language Processing, {EMNLP} 2024, Miami, FL, USA, November 12-16,
                  2024},
  pages        = {750--766},
  publisher    = {Association for Computational Linguistics},
  year         = {2024},
  url          = {https://doi.org/10.18653/v1/2024.emnlp-main.44}
        
        
        
        
        
        ,
  doi          = {10.18653/V1/2024.EMNLP-MAIN.44
        
        },
  bibsource    = {dblp computer science bibliography, https://dblp.org}
}

@article{DBLP:journals/corr/abs-2509-17437,
  author       = {Guizhen Chen and
                  Weiwen Xu and
                  Hao Zhang and
                  Hou Pong Chan and
                  Deli Zhao and
                  Anh Tuan Luu and
                  Yu Rong},
  title        = {GeoPQA: Bridging the Visual Perception Gap in {MLLM}s for Geometric
                  Reasoning},
  journal      = {CoRR},
  volume       = {abs/2509.17437},
  year         = {2025},
  url          = {https://doi.org/10.48550/arXiv.2509.17437},
  doi          = {10.48550/ARXIV.2509.17437},
  eprinttype    = {arXiv},
  eprint       = {2509.17437},
  bibsource    = {dblp computer science bibliography, https://dblp.org}
}

@misc{deepseekai2025deepseekv32pushingfrontieropen,
      title={DeepSeek-V3.2: Pushing the Frontier of Open Large Language Models}, 
      author={DeepSeek-AI and Aixin Liu and Aoxue Mei and Bangcai Lin and Bing Xue and Bingxuan Wang and Bingzheng Xu and Bochao Wu and Bowei Zhang and Chaofan Lin and Chen Dong and Chengda Lu and Chenggang Zhao and Chengqi Deng and Chenhao Xu and Chong Ruan and Damai Dai and Daya Guo and Dejian Yang and Deli Chen and Erhang Li and Fangqi Zhou and Fangyun Lin and Fucong Dai and Guangbo Hao and Guanting Chen and Guowei Li and H. Zhang and Hanwei Xu and Hao Li and Haofen Liang and Haoran Wei and Haowei Zhang and Haowen Luo and Haozhe Ji and Honghui Ding and Hongxuan Tang and Huanqi Cao and Huazuo Gao and Hui Qu and Hui Zeng and Jialiang Huang and Jiashi Li and Jiaxin Xu and Jiewen Hu and Jingchang Chen and Jingting Xiang and Jingyang Yuan and Jingyuan Cheng and Jinhua Zhu and Jun Ran and Junguang Jiang and Junjie Qiu and Junlong Li and Junxiao Song and Kai Dong and Kaige Gao and Kang Guan and Kexin Huang and Kexing Zhou and Kezhao Huang and Kuai Yu and Lean Wang and Lecong Zhang and Lei Wang and Liang Zhao and Liangsheng Yin and Lihua Guo and Lingxiao Luo and Linwang Ma and Litong Wang and Liyue Zhang and M. S. Di and M. Y Xu and Mingchuan Zhang and Minghua Zhang and Minghui Tang and Mingxu Zhou and Panpan Huang and Peixin Cong and Peiyi Wang and Qiancheng Wang and Qihao Zhu and Qingyang Li and Qinyu Chen and Qiushi Du and Ruiling Xu and Ruiqi Ge and Ruisong Zhang and Ruizhe Pan and Runji Wang and Runqiu Yin and Runxin Xu and Ruomeng Shen and Ruoyu Zhang and S. H. Liu and Shanghao Lu and Shangyan Zhou and Shanhuang Chen and Shaofei Cai and Shaoyuan Chen and Shengding Hu and Shengyu Liu and Shiqiang Hu and Shirong Ma and Shiyu Wang and Shuiping Yu and Shunfeng Zhou and Shuting Pan and Songyang Zhou and Tao Ni and Tao Yun and Tian Pei and Tian Ye and Tianyuan Yue and Wangding Zeng and Wen Liu and Wenfeng Liang and Wenjie Pang and Wenjing Luo and Wenjun Gao and Wentao Zhang and Xi Gao and Xiangwen Wang and Xiao Bi and Xiaodong Liu and Xiaohan Wang and Xiaokang Chen and Xiaokang Zhang and Xiaotao Nie and Xin Cheng and Xin Liu and Xin Xie and Xingchao Liu and Xingkai Yu and Xingyou Li and Xinyu Yang and Xinyuan Li and Xu Chen and Xuecheng Su and Xuehai Pan and Xuheng Lin and Xuwei Fu and Y. Q. Wang and Yang Zhang and Yanhong Xu and Yanru Ma and Yao Li and Yao Li and Yao Zhao and Yaofeng Sun and Yaohui Wang and Yi Qian and Yi Yu and Yichao Zhang and Yifan Ding and Yifan Shi and Yiliang Xiong and Ying He and Ying Zhou and Yinmin Zhong and Yishi Piao and Yisong Wang and Yixiao Chen and Yixuan Tan and Yixuan Wei and Yiyang Ma and Yiyuan Liu and Yonglun Yang and Yongqiang Guo and Yongtong Wu and Yu Wu and Yuan Cheng and Yuan Ou and Yuanfan Xu and Yuduan Wang and Yue Gong and Yuhan Wu and Yuheng Zou and Yukun Li and Yunfan Xiong and Yuxiang Luo and Yuxiang You and Yuxuan Liu and Yuyang Zhou and Z. F. Wu and Z. Z. Ren and Zehua Zhao and Zehui Ren and Zhangli Sha and Zhe Fu and Zhean Xu and Zhenda Xie and Zhengyan Zhang and Zhewen Hao and Zhibin Gou and Zhicheng Ma and Zhigang Yan and Zhihong Shao and Zhixian Huang and Zhiyu Wu and Zhuoshu Li and Zhuping Zhang and Zian Xu and Zihao Wang and Zihui Gu and Zijia Zhu and Zilin Li and Zipeng Zhang and Ziwei Xie and Ziyi Gao and Zizheng Pan and Zongqing Yao and Bei Feng and Hui Li and J. L. Cai and Jiaqi Ni and Lei Xu and Meng Li and Ning Tian and R. J. Chen and R. L. Jin and S. S. Li and Shuang Zhou and Tianyu Sun and X. Q. Li and Xiangyue Jin and Xiaojin Shen and Xiaosha Chen and Xinnan Song and Xinyi Zhou and Y. X. Zhu and Yanping Huang and Yaohui Li and Yi Zheng and Yuchen Zhu and Yunxian Ma and Zhen Huang and Zhipeng Xu and Zhongyu Zhang and Dongjie Ji and Jian Liang and Jianzhong Guo and Jin Chen and Leyi Xia and Miaojun Wang and Mingming Li and Peng Zhang and Ruyi Chen and Shangmian Sun and Shaoqing Wu and Shengfeng Ye and T. Wang and W. L. Xiao and Wei An and Xianzu Wang and Xiaowen Sun and Xiaoxiang Wang and Ying Tang and Yukun Zha and Zekai Zhang and Zhe Ju and Zhen Zhang and Zihua Qu},
      year={2025},
      eprint={2512.02556},
      archivePrefix={arXiv},
      primaryClass={cs.CL},
      url={https://arxiv.org/abs/2512.02556}, 
}
\bibliographystyle{icml2026}

\newpage
\appendix
\onecolumn

\section{Dataset Construction Details}
In this appendix section, we provide a detailed breakdown of the data generation pipeline used to construct the dataset for Med-Scout training and the Med-Scout-Bench. Driven by the objective to cure geometric blindness without relying on expensive human annotations, our pipeline automatically extracts verifiable geometric supervision signals directly from raw medical images. We first elaborate on the composition and distribution of the 108,000-sample dataset, which spans diverse modalities including CT, MRI, and X-ray sourced from TotalSegmentor~\cite{wasserthal2023totalsegmentator, akinci2025totalsegmentator} and MIMIC-CXR~\cite{johnson2019mimic}. Subsequently, we describe the rigorous algorithmic protocols employed to synthesize the three geometric proxy tasks. Finally, we present the unified VQA instruction templates used to standardize these tasks for effective RL post-training.
\subsection{Data Composition and Distribution Statistics}
We primarily analyze the Med-Scout-Bench dataset. This is a specific test set of 10,800 examples designed to accurately measure geometric blindness. The rest of the data used for training and validation ($N=97,200$) follows the same pattern, ensuring that the way the model is trained matches the way it is tested.

\begin{figure}[b]
    \centering
    \includegraphics[width=0.7\textwidth]{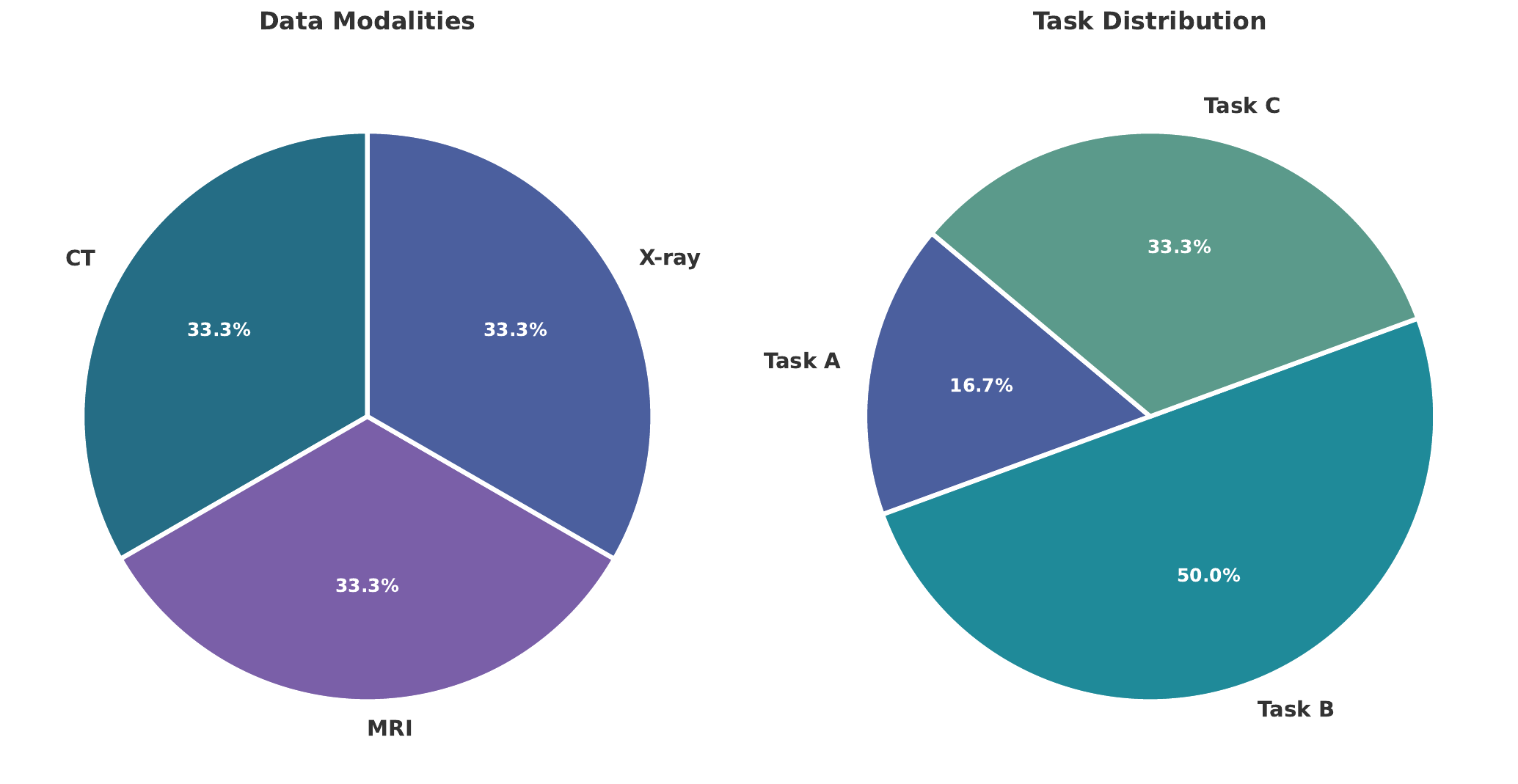}
    \caption{Data Statistics of Med-Scout-Bench. Left: The benchmark maintains a strictly balanced distribution across CT, MRI, and X-ray modalities to ensure unbiased geometric evaluation. Right: The benchmark is under task-specific distribution.}
    \label{fig:data_stats}
\end{figure}

As shown in Figure \ref{fig:data_stats}, the benchmark follows two key distribution patterns:
\begin{itemize}[leftmargin=*]
    \item \textbf{Strict Modality Balance:} To avoid bias toward any specific imaging method, we ensure an equal distribution across the three main modalities. The benchmark is split evenly, with CT, MRI, and X-ray each making up about $33.3\%$ of the data.
    \item \textbf{Task-Specific Distribution:} To ensure balanced training, we allocate sample sizes based on the difficulty of each task. We assign 1,800 samples to Task A (Hierarchical Scale Localization), as this fundamental spatial task requires less data to converge. The largest share (5,400 samples) goes to Task B (Topological Jigsaw Reconstruction); its complex anatomical puzzles require robust reasoning, necessitating more data. Finally, we dedicate 3,600 samples to Task C (Anomaly Consistency Detection), an intermediate amount sufficient for the model to learn to detect fine details and subtle anomalies.
\end{itemize}

\subsection{Task-Specific Generation Protocols}
\label{task_generation}
\subsubsection{Hierarchical Scale Localization}

This task is designed to compel the model to master absolute spatial grounding and multi-scale consistency. By forcing the model to map resized local patches back to their original global coordinates, we simulate a ``zoom-in'' clinical diagnostic process where a radiologist examines local details (e.g., a nodule) while maintaining awareness of its global position (e.g., upper right lung lobe). The generation process is formalized in Algorithm \ref{alg:scale_gen}.

\textbf{Scale Definitions and Sampling.}
We define two distinct scale levels to represent different granularities of anatomical detail:
\begin{itemize}[leftmargin=*]
    \item \textbf{Level 1 (Regional View):} The patch area covers $20\%$ of the original image area ($A_\text{{patch}} \approx 0.2 \times A_\text{{total}}$).
    \item \textbf{Level 2 (Focal View):} The patch area covers $6.25\%$ of the original image area ($A_\text{{patch}} \approx 0.0625 \times A_\text{{total}}$), corresponding to a $4\times$ zoom relative to the original resolution.
\end{itemize}
For each training instance, we randomly extract $N=3$ square patches. To ensure the patches contain meaningful anatomical information rather than non-informative background (e.g., black borders common in raw medical scans), the sampling of crop regions is strictly restricted to the valid central interval $[0.2, 0.8]$ of the image dimensions.

\textbf{Input and Objectives.}
The input to the model consists of the original global image $I_\text{{global}}$ followed by the three resized local patches $\{P_1, P_2, P_3\}$. All patches are resized to the dimensions of $I_\text{{global}}$. The model is prompted to perform two sub-tasks:
\begin{enumerate}[leftmargin=*]
    \item \textbf{Scale Classification:} Identify whether each patch belongs to Scale Level 1 or Level 2.
    \item \textbf{Coordinate Regression:} Explicitly predict the normalized bounding box $(x_1, y_1, x_2, y_2)$ of the patch in the original image frame.
\end{enumerate}

\newcommand{\RETURN}{\STATE \textbf{return} }
 
\begin{algorithm}[htbp]
    \caption{Data Generation for Hierarchical Scale Localization}
    \label{alg:scale_gen}
    \begin{algorithmic}[1]
        \REQUIRE Original Image $I \in \mathbb{R}^{H \times W}$
        \REQUIRE \textbf{Hyperparameters:}
        \STATE \hspace{1em} Num Patches $N \leftarrow 3$
        \STATE \hspace{1em} Scale Ratios $\mathcal{S} \leftarrow \{0.20, 0.0625\}$ \COMMENT{Area ratios for Level 1 and Level 2}
        \STATE \hspace{1em} ROI Bounds $[\alpha_\text{min}, \alpha_\text{max}] \leftarrow [0.2, 0.8]$ \COMMENT{Normalized valid center region}
        \STATE \hspace{1em} Target Size $(H_\text{out}, W_\text{out}) \leftarrow (H, W)$ \COMMENT{Resize targets to original resolution}

        \ENSURE Resized Patches $\mathcal{P}$, Norm. Coordinates $\mathcal{B}$, Scale Labels $\mathcal{L}$
        \STATE Initialize $\mathcal{P} \leftarrow [\text{ }], \mathcal{B} \leftarrow [\text{ }], \mathcal{L} \leftarrow [\text{ }]$
        
        \FOR{$k=1$ to $N$}
            \STATE \textit{// Step 1: Determine Crop Size}
            \STATE Sample ratio $r \in \mathcal{S}$ uniformly
            \STATE Calculate square side length: $l \leftarrow \sqrt{r \cdot H \cdot W}$
            
            \STATE \textit{// Step 2: Calculate Sampling Boundaries}
            \STATE \textit{// Ensure crop is within ROI: start $\ge \alpha_\text{min}$, end $\le \alpha_\text{max}$}
            \STATE $x_\text{min} \leftarrow \alpha_\text{min} \cdot W$
            \STATE $x_\text{max} \leftarrow \alpha_\text{max} \cdot W - l$
            \STATE $y_\text{min} \leftarrow \alpha_\text{min} \cdot H$
            \STATE $y_\text{max} \leftarrow \alpha_\text{max} \cdot H - l$
            
            \STATE \textit{// Step 3: Sample and Extract}
            \STATE Sample top-left $x \sim \text{Uniform}(x_\text{min}, x_\text{max})$
            \STATE Sample top-left $y \sim \text{Uniform}(y_\text{min}, y_\text{max})$
            \STATE $Patch_\text{raw} \leftarrow I[y : y+l, \,\, x : x+l]$
            
            \STATE \textit{// Step 4: Resize and Normalize}
            \STATE $Patch_\text{resize} \leftarrow \text{Resize}(Patch_\text{raw}, (H_\text{out}, W_\text{out}))$
            \STATE $Box_\text{norm} \leftarrow [x/W, y/H, (x+l)/W, (y+l)/H]$
            
            \STATE Append $Patch_\text{resize}$ to $\mathcal{P}$, $Box_\text{norm}$ to $\mathcal{B}$, $r$ to $\mathcal{L}$
        \ENDFOR
        \RETURN $\mathcal{P}, \mathcal{B}, \mathcal{L}$
    \end{algorithmic}
\end{algorithm}

\subsubsection{Topological Jigsaw Reconstruction}
This task tests the model's ability to infer the overall anatomical structure from specific details. Unlike other jigsaw tasks that often use complex, many-piece grids, we use a simple $2 \times 2$ grid. This choice is essential for medical imaging:
\begin{itemize}[leftmargin=*]
    \item \textbf{Semantic Integrity:} A $2 \times 2$ grid ensures that each section contains recognizable anatomical features, such as a complete left lung or the clear shape of the pelvis.
    \item \textbf{Logical Deduction:} It shifts the reasoning burden from low-level pattern matching to high-level topological deduction (e.g., reasoning that the ``heart'' patch must be spatially adjacent to and above the ``stomach'' patch).
\end{itemize}

\textbf{Generation Protocol.}
The image $I$ is partitioned into four quadrants. We generate a random permutation $\sigma$ of the index set $\{0, 1, 2, 3\}$, where indices correspond to the canonical reading order (top-left, top-right, bottom-left, bottom-right). The quadrants are rearranged according to $\sigma$ to form the shuffled observation $I_\text{shuffled}$. The model is tasked with reconstructing the sequence of original indices. The detailed generation process is described in Algorithm \ref{alg:jigsaw_gen}.

\begin{algorithm}[h]
    \caption{Data Generation for Topological Jigsaw Reconstruction}
    \label{alg:jigsaw_gen}
    \begin{algorithmic}[1]
        \REQUIRE Original Image $I \in \mathbb{R}^{H \times W}$
        \REQUIRE \textbf{Hyperparameters:}
        \STATE \hspace{1em} Grid Dimension $G \leftarrow 2$ \COMMENT{Partitions image into $2 \times 2$ quadrants}
        \STATE \hspace{1em} Patch Height $H_\text{p} \leftarrow H // G$
        \STATE \hspace{1em} Patch Width $W_\text{p} \leftarrow W // G$
        \STATE \hspace{1em} Canonical Indices $\mathcal{K} \leftarrow \{0, 1, 2, 3\}$ \COMMENT{Reading order: TL, TR, BL, BR}

        \ENSURE Shuffled Image $I_\text{shuffled}$, Target Sequence $Y^*$
        
        \STATE \textit{// Step 1: Extract Canonical Patches}
        \STATE Define grid coordinates $C \leftarrow \{(0,0), (0, W_\text{p}), (H_\text{p}, 0), (H_\text{p}, W_\text{p})\}$
        \STATE Initialize patch library $\mathcal{P}_\text{lib} \leftarrow [\text{ }]$
        
        \FOR{$k \in \mathcal{K}$}
            \STATE $(y, x) \leftarrow C[k]$
            \STATE Extract $P \leftarrow I[y : y+H_\text{p}, \,\, x : x+W_\text{p}]$
            \STATE Append $P$ to $\mathcal{P}_\text{lib}$
        \ENDFOR
        
        \STATE \textit{// Step 2: Generate Permutation}
        \STATE Generate random permutation $\sigma$ of $\mathcal{K}$
        \STATE Initialize canvas $I_\text{shuffled}$ of size $H \times W$
        
        \STATE \textit{// Step 3: Reconstruct with Shuffled Order}
        \FOR{$i \in \mathcal{K}$}
            \STATE $idx_\text{source} \leftarrow \sigma[i]$ \COMMENT{Select which original patch goes to position $i$}
            \STATE $(y_\text{target}, x_\text{target}) \leftarrow C[i]$ \COMMENT{Get coordinates for position $i$}
            \STATE Place $\mathcal{P}_\text{lib}[idx_\text{source}]$ into $I_\text{shuffled}$ at $(y_\text{target}, x_\text{target})$
        \ENDFOR
        
        \STATE \textit{// Target is the permutation sequence defining the layout}
        \STATE $Y^* \leftarrow \sigma$
        
        \RETURN $I_\text{shuffled}, Y^*$
    \end{algorithmic}
\end{algorithm}

\subsubsection{Anomaly Consistency Detection}
This task trains the model to detect subtle anomalies and structural errors. We generate these examples using a ``cut-paste'' method: a section of the anatomy is replaced by a similar-looking, but incorrect, patch from a reference image ($I_\text{ref}$). To prevent the model from cheating by detecting sharp edges, we use Gaussian noise to blend the boundaries where the images meet.

\textbf{Reference Image Selection ($I_\text{ref}$).}
To ensure the anomaly is non-trivial, we select $I_\text{ref}$ based on modality-specific hardness:
\begin{itemize}[leftmargin=*]
    \item \textbf{Volumetric (CT/MRI):} $I_\text{ref}$ is selected from a nearby slice ($z \pm 5$). This ensures the overall organ shape remains consistent, while introducing slight natural variations.
    \item \textbf{Planar (X-ray):} $I_\text{ref}$ is the top-1 retrieval from the dataset via BiomedCLIP embedding similarity, ensuring semantic density consistency.
\end{itemize}

\textbf{Generation Protocol.}
We partition the image into a $4 \times 4$ grid. The anomaly is injected into the central region defined by $\mathcal{C}_\text{center}$. Crucially, to mitigate the ``sharp edge'' artifact common in cut-paste augmentations, we add Gaussian noise to the boundary pixels of the pasted region. The detailed procedure and hyperparameters are formalized in Algorithm \ref{alg:anomaly_gen}.

\begin{algorithm}[h]
    \caption{Data Generation for Anomaly Consistency Detection}
    \label{alg:anomaly_gen}
    \begin{algorithmic}[1]
        \REQUIRE Target Image $I \in \mathbb{R}^{H \times W}$, Modality $M$, Database $\mathcal{D}$
        \REQUIRE \textbf{Hyperparameters:}
        \STATE \hspace{1em} Grid Dimension $G \leftarrow 4$
        \STATE \hspace{1em} Patch Size $S \leftarrow (H/G, W/G)$
        \STATE \hspace{1em} Center Indices $\mathcal{C}_\text{center} \leftarrow \{5, 6, 9, 10\}$ \COMMENT{Central $2 \times 2$ block in flattened index}
        \STATE \hspace{1em} Noise Level $\sigma_\text{noise} \leftarrow 0.05$ \COMMENT{Std dev for boundary blending}
        \STATE \hspace{1em} Boundary Width $\delta \leftarrow 2$ \COMMENT{Pixel width for noise injection}

        \ENSURE Anomalous Image $I_\text{anom}$, Anomaly Index $k^*$
        
        \STATE \textit{// Step 1: Select Hard Negative Reference}
        \IF{$M \in \{\text{CT}, \text{MRI}\}$}
            \STATE $I_\text{ref} \leftarrow \text{GetSlice}(I.volume, I.z \pm 1)$
        \ELSE
            \STATE $v_\text{I} \leftarrow \text{BioMedCLIP}(I)$
            \STATE $I_\text{ref} \leftarrow \operatorname{argmax}_{\text{J} \in \mathcal{D}, \text{J} \neq \text{I}} (\text{CosSim}(v_\text{I}, v_\text{J}))$
        \ENDIF
        
        \STATE \textit{// Step 2: Inject Anomaly with Edge Blending}
        \STATE Partition $I_\text{ref}$ into grid $\mathcal{P}_\text{ref}$
        \STATE Sample target index $k^* \sim \text{Uniform}(\mathcal{C}_\text{center})$
        \STATE Extract foreign patch $P_\text{foreign} \leftarrow \mathcal{P}_\text{ref}[k^*]$
        
        \STATE $I_\text{anom} \leftarrow I$
        \STATE Paste $P_\text{foreign}$ into $I_\text{anom}$ at position $k^*$
        
        \STATE \textit{// Step 3: Apply Boundary Noise}
        \STATE Get boundary region $\Omega$ of width $\delta$ around position $k^*$
        \STATE Generate noise $\epsilon \sim \mathcal{N}(0, \sigma_\text{noise})$
        \STATE $I_\text{anom}[\Omega] \leftarrow I_\text{anom}[\Omega] + \epsilon$
        
        \RETURN $I_\text{anom}, k^*$
    \end{algorithmic}
\end{algorithm}

\subsection{Unified VQA Instruction Formatting}
\label{VQA}

To facilitate end-to-end training using a unified objective, we standardize all geometric proxy tasks into a consistent open-set VQA format. Instead of using task-specific heads, we formulate these tasks as natural language conversations.

\begin{figure}[H]
    \centering
    \includegraphics[width=\textwidth]{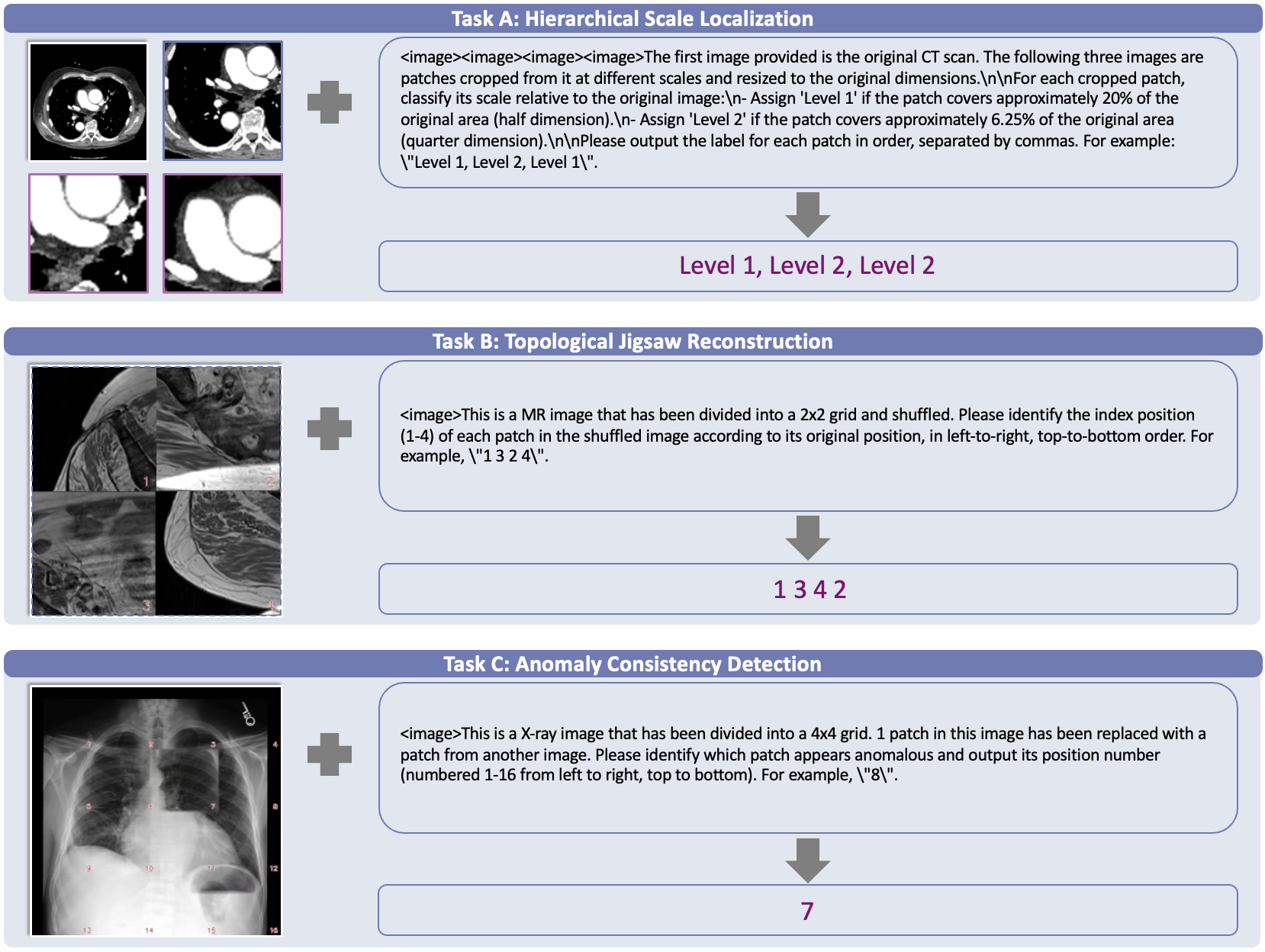}
    \caption{Unified VQA Instruction Examples of Direct Mode.}
    \label{fig:vqa_formatting}
\end{figure}

\begin{figure}[h]
    \centering
    \includegraphics[width=\textwidth]{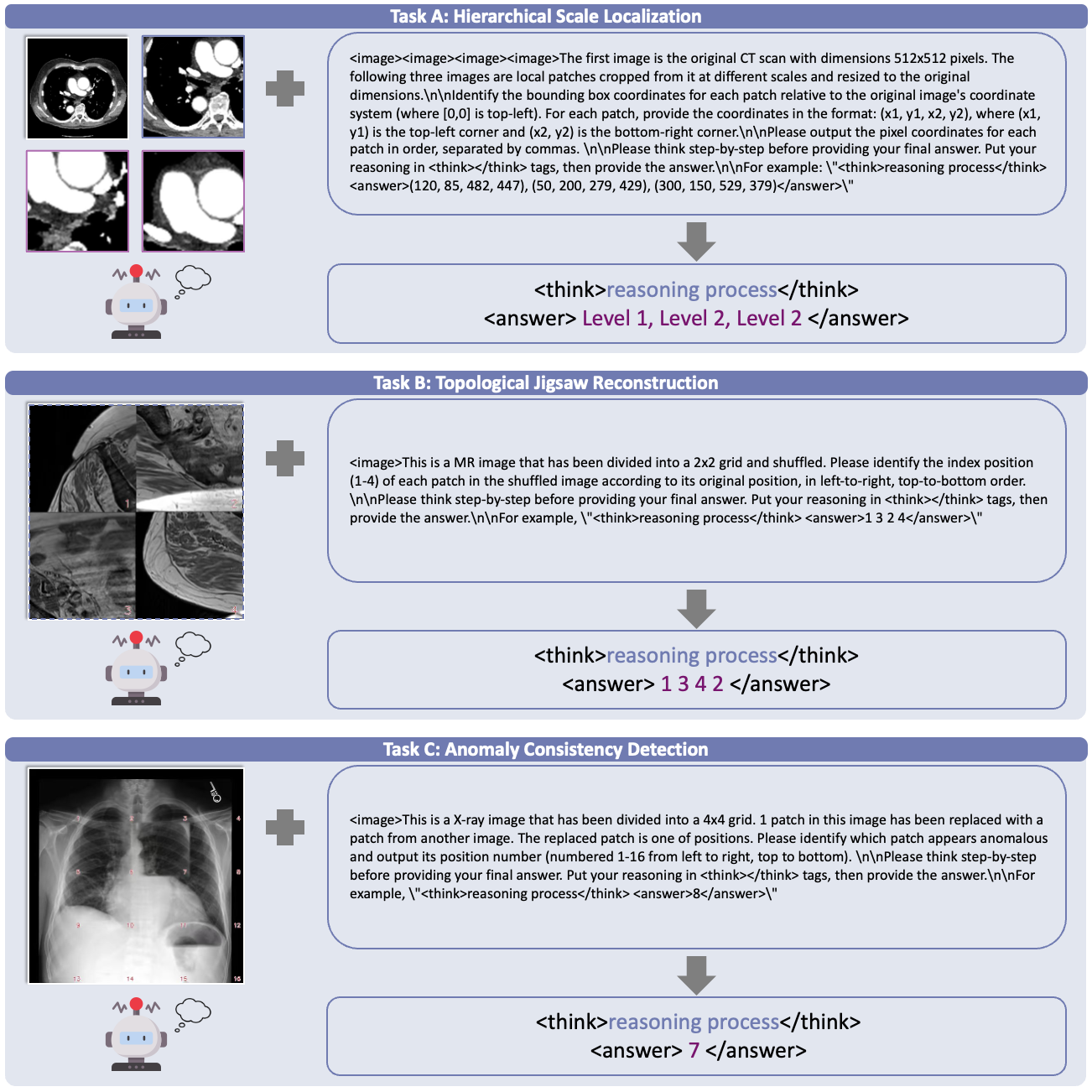}
    \caption{Unified VQA Instruction Examples of Reasoning Mode.}
    \label{fig:vqa_formatting_2}
\end{figure}

As illustrated in Figure \ref{fig:vqa_formatting}, each training instance is composed of three standardized components:
\begin{enumerate}[leftmargin=*]
    \item \textbf{Visual Input:} We use special tokens (e.g., \texttt{<image>}) to represent the medical scans. Note that for the \textit{Hierarchical Scale Localization} task (Figure \ref{fig:vqa_formatting}a), the input specifically supports multi-image sequences (Global View + Local Crops).
    \item \textbf{User Prompt:} A structured instruction that clearly defines the geometric objective and constrains the output format.
    \item \textbf{Target Response:} To support diverse inference strategies, we define two distinct output states:
    \begin{itemize}
        \item \textbf{Direct Mode:} The model directly outputs the concise final answer (e.g., sequence indices or grid coordinates), focusing on strict format.
        \item \textbf{Reasoning Mode:} The model first generates a CoT reasoning path enclosed in \texttt{<think>...</think>} tags to articulate geometric constraints before deriving the final answer enclosed in \texttt{<answer>...</answer>} tags.
    \end{itemize}
    Visual examples of the direct mode and the reasoning mode are provided in Figure~\ref{fig:vqa_formatting} and Figure~\ref{fig:vqa_formatting_2}.
\end{enumerate}

\subsection{Med-Scout-Bench Evaluation Pipeline}
\begin{figure}[H]
    \centering
    \includegraphics[width=\textwidth]{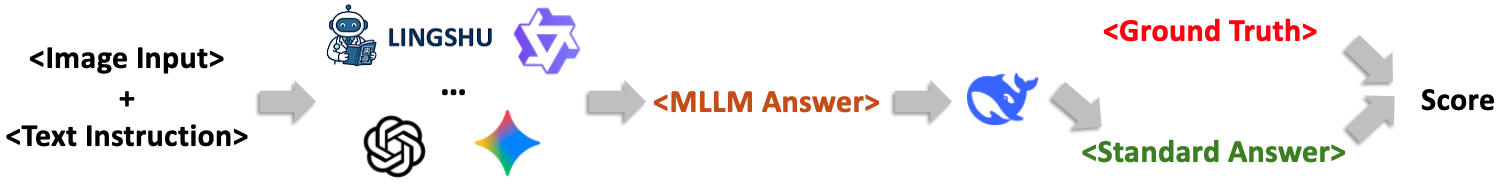}
    \caption{Evaluation pipeline on Med-Scout-Bench.}
    \label{fig:evaluation_pipeline}
\end{figure}

To ensure consistent scoring across models with different styles, we use a standardized evaluation process as shown in Figure~\ref{fig:evaluation_pipeline}. Since MLLMs often provide detailed reasoning that makes simple text matching difficult, we use DeepSeek-V3.2~\cite{deepseekai2025deepseekv32pushingfrontieropen} to extract the conclusion from the raw output. These extracted answers are then automatically compared against the correct labels using a strict scoring script.

\section{Training Implementation Details}
\subsection{Hyperparameters}

We provide a comprehensive list of hyperparameters used during the Med-Scout RL post-training phase in Table~\ref{tab:hyperparameters}. The training configuration is standardized across all backbone models to ensure fair comparison.

\begin{table*}[t]
\centering
\caption{Comprehensive list of hyperparameters for Med-Scout training. The configuration covers optimization, GRPO strategy, reward engineering, and system settings.}
\label{tab:hyperparameters}
\begin{center}
\begin{small}
\begin{sc}
\resizebox{\textwidth}{!}{%
\begin{tabular*}{\textwidth}{@{\extracolsep{\fill}}lll}
\toprule
Hyperparameter & Value & Description \\
\midrule
\rowcolor{headergray} \multicolumn{3}{l}{\textit{Optimization Configuration}} \\
Optimizer & AdamW & $\beta_1=0.9, \beta_2=0.95$ \\
Peak Learning Rate & $1 \times 10^{-6}$ & Lower than SFT to prevent collapse \\
LR Scheduler & Cosine Decay & Minimum LR set to $1 \times 10^{-7}$ \\
Warm-up Ratio & 0.01 & Linear warm-up strategy \\
Weight Decay & 0.1 & Standard regularization \\
Training Steps & 7200 & Total optimization updates \\
\midrule
\rowcolor{headergray} \multicolumn{3}{l}{\textit{GRPO Strategy}} \\
Global Batch Size & 192 & 192 for all models \\
Group Size ($G$) & 8 & Number of outputs sampled per prompt \\
KL Coefficient ($\beta$) & 0.04 & Penalty weight for policy drift \\
Clip Ratio ($\epsilon$) & 0.2 & Standard PPO clipping range \\
\midrule
\rowcolor{headergray} \multicolumn{3}{l}{\textit{Reward Engineering}} \\
Total Max Reward & 2.0 & Sum of accuracy, format, and reasoning rewards \\
Accuracy Cap ($\mathcal{R}_\text{acc}$) & 1.0 & Task-specific geometric precision \\
Format Cap ($\mathcal{R}_\text{fmt}$) & 0.5 & Syntax compliance reward \\
Reasoning Cap ($\mathcal{R}_\text{reason}$) & 0.5 & Active only in Reasoning Mode (CoT) \\
Anomaly Temp ($\tau$) & 0.1 & Temperature for distance-based reward \\
\midrule
\rowcolor{headergray} \multicolumn{3}{l}{\textit{System \& Generation}} \\
Max New Tokens & 1024 & Buffer for CoT reasoning traces \\
Precision & BF16 & Mixed precision training \\
Hardware & $6 \times$ RTX PRO 6000 & NVIDIA RTX PRO 6000 GPUs \\
\bottomrule
\end{tabular*}%
}
\end{sc}
\end{small}
\end{center}
\vskip -0.1in
\end{table*}

\subsection{Reward Curves}
\label{appendix:reward_curves}

\begin{figure*}[h]
    \centering
    \includegraphics[width=\textwidth]{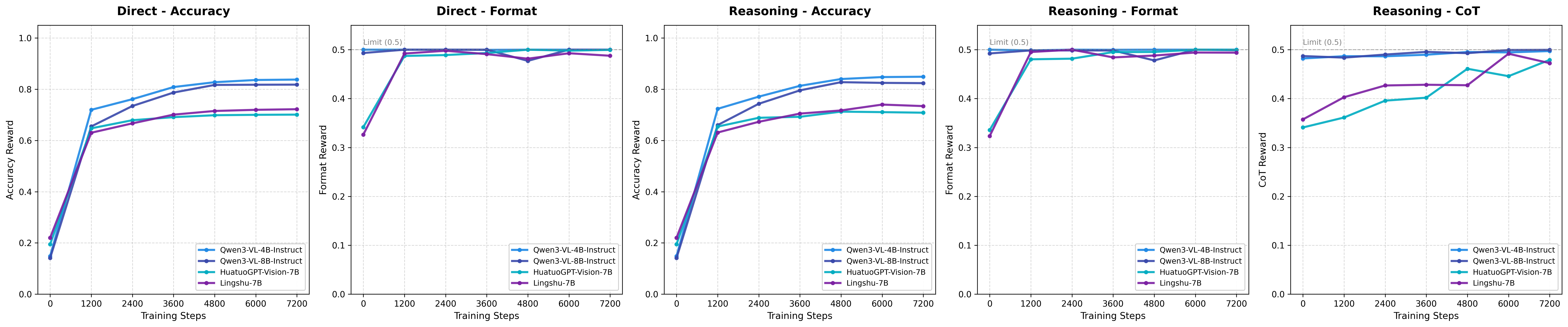}
    \caption{Training Reward Dynamics. The plots illustrate the optimization trajectories for Direct Mode (left two panels) and Reasoning Mode (right three panels). We report the Dense Geometric Rewards ($\mathcal{R}_\text{acc}$), Format Rewards ($\mathcal{R}_\text{fmt}$), and Reasoning Structure Rewards ($\mathcal{R}_\text{reason}$).}
    \label{fig:reward_curves}
\end{figure*}

To verify the stability and convergence of our geometry-aware post-training, we visualize the reward trajectories during the GRPO training phase. Figure \ref{fig:reward_curves} presents the comprehensive learning curves for both the Direct Mode and the Reasoning Mode across the four backbone models.

\subsubsection{Direct Mode Dynamics}
The first two panels of Figure \ref{fig:reward_curves} illustrate the training dynamics under the Direct Mode setting.
\begin{itemize}[leftmargin=*]
    \item \textbf{Dense Geometric Reward ($\mathcal{R}_\text{acc}$):} The \textit{Direct - Accuracy} subplot depicts the steady improvement in geometric perception. We observe that general-domain models (Qwen3-VL series) achieve slightly higher accuracy ($\sim 0.84$) compared to medical specialists ($\sim 0.72$). This suggests that strong foundational vision-language capabilities are beneficial for solving complex spatial reasoning tasks.
    \item \textbf{Format Reward ($\mathcal{R}_\text{fmt}$):} As shown in the \textit{Direct - Format} subplot, all models rapidly master the requisite output format constraints (e.g., coordinate normalization, comma-separated lists). The format reward converges to the maximum value of 0.5 within the first 1,200 steps, indicating that the syntax of the geometric proxy tasks is easily learnable.
\end{itemize}

\subsubsection{Reasoning Mode Dynamics}
In the Reasoning Mode, models are required to generate a CoT trace before the final answer. The last three panels of Figure \ref{fig:reward_curves} display the corresponding reward curves.

\begin{itemize}[leftmargin=*]
    \item \textbf{Dense Geometric Reward ($\mathcal{R}_\text{acc}$):} Compared to Direct Mode, the Reasoning Mode typically exhibits a slightly slower convergence rate initially due to the increased generation length and complexity. However, the final convergence values remain consistent with the Direct Mode, validating the robustness of the alignment strategy.
    \item \textbf{Format Reward ($\mathcal{R}_\text{fmt}$):} Similar to the Direct Mode, the format compliance in Reasoning Mode exhibits extremely rapid convergence. As shown in the \textit{Reasoning - Format} subplot, all models quickly saturate at the maximum reward of 0.5, confirming that the increased sequence length from the reasoning process does not degrade the model's ability to adhere to syntactic instructions.
    \item \textbf{Reasoning Structure Reward ($\mathcal{R}_\text{reason}$):} As illustrated in the \textit{Reasoning - CoT} subplot, the models effectively adapt to the \texttt{<think>...</think>} and \texttt{<answer>...</answer>} structure. Notably, while the general-domain Qwen3 models maintain high structural adherence from the start, the medical specialists (HuatuoGPT, Lingshu) show a distinct learning curve, requiring approximately 3,600 steps to converge to the maximum structural reward.
\end{itemize}

\section{Extensive Experimental Results}

\subsection{Detailed Evaluation on Med-Scout-Bench}
\label{appendix:med_scout_bench_eval}

Due to space constraints in the main text, we focused on the aggregated geometric alignment scores. In this section, we provide a comprehensive performance breakdown for each backbone model across the three specific proxy tasks: Hierarchical Scale Localization ($\mathcal{T}_\text{scale}$), Topological Jigsaw Reconstruction ($\mathcal{T}_\text{topo}$), and Anomaly Consistency Detection ($\mathcal{T}_\text{anom}$).

The detailed numerical results are presented in Table \ref{tab:med_scout_bench}. Beyond the performance gains, three notable patterns emerge from our findings:

\begin{itemize}[leftmargin=*]
    \item \textbf{General-Domain Models Learn Faster and Better:} 
    General-purpose models (e.g., Qwen3-VL series) consistently outperform medical specialist models (e.g., Lingshu, HuatuoGPT) in learning geometric concepts. As shown in the reward curves and final scores, general models converge more rapidly and achieve significantly higher final accuracy. This suggests that a strong fundamental vision-language capability is more critical for grasping abstract spatial logic than domain-specific medical training.
    \item \textbf{Smaller Models Can Be More Efficient (4B vs. 8B):} 
    Contrary to the scaling law expectation that ``larger is better,'' we observe that the Qwen3-VL-4B model learns faster and achieves higher geometric accuracy (Avg. 84.4) compared to its larger 8B counterpart (Avg. 83.6). This indicates that for targeted geometric alignment, parameter efficiency and training dynamics may matter more than sheer model size.
    \item \textbf{Med-Scout Bridges the Gap with Proprietary Models:} 
    In their base states, proprietary models like GPT-5 and Gemini-3-Flash significantly outperform open-source models (scoring 60\% vs. 30\%), likely due to their superior pre-training. However, after Med-Scout post-training, open-source models not only close this gap but easily surpass these closed-source leaders, with scores jumping to the 70\%-90\% range. This demonstrates that specific geometric supervision is highly effective at unlocking capabilities that even the strongest proprietary models lack.
\end{itemize}

\begin{table*}[h]
\centering
\caption{Performance evaluation on the Med-Scout-Bench. The table reports scores on three distinct subtasks and their average. All scores are scaled by a factor of 100 for better readability.}
\label{tab:med_scout_bench}
\begin{center}
\begin{small}
\begin{sc}
\resizebox{\textwidth}{!}{%
\begin{tabular*}{\textwidth}{@{\extracolsep{\fill}}lllll}
\toprule
\multirow{2}{*}{Model} & \multicolumn{4}{c}{Med-Scout-Bench} \\
\cmidrule(lr){2-5}
 & Task A & Task B & Task C & Avg. \\
\midrule
\rowcolor{headergray} \multicolumn{5}{l}{\textit{Proprietary Models}} \\
GPT-5 & 64.3 & 56.1 & 74.6 & 63.6 \\
Gemini-3-Flash & 67.0 & 58.8 & 76.7 & 66.1 \\
\midrule
\rowcolor{headergray} \multicolumn{5}{l}{\textit{General-purpose MLLMs}} \\
InternVL3-8B & 37.7 & 34.3 & 27.3 & 32.5 \\
Qwen2.5-VL-3B-Instruct & 46.9 & 34.6 & 9.2 & 28.2 \\
Qwen2.5-VL-7B-Instruct & 31.4 & 28.1 & 27.8 & 28.5 \\
Qwen2.5-VL-32B-Instruct & 28.9 & 37.3 & 19.6 & 30.0 \\
Qwen2.5-VL-72B-Instruct & 32.7 & 29.9 & 30.4 & 30.5 \\
Qwen3-VL-4B-Instruct & 59.6 & 36.6 & 31.3 & 38.7 \\
\rowcolor{color1} \quad + \textbf{Med-Scout} &
94.4{\scriptsize\textcolor{green!40!black}{$\uparrow$\,34.8}} &
77.5{\scriptsize\textcolor{green!40!black}{$\uparrow$\,40.9}} &
89.8{\scriptsize\textcolor{green!40!black}{$\uparrow$\,58.5}} &
84.4{\scriptsize\textcolor{green!40!black}{$\uparrow$\,45.7}} \\
Qwen3-VL-8B-Instruct & 41.0 & 34.7 & 46.4 & 39.7 \\
\rowcolor{color2} \quad + \textbf{Med-Scout} &
86.7{\scriptsize\textcolor{green!40!black}{$\uparrow$\,45.7}} &
78.1{\scriptsize\textcolor{green!40!black}{$\uparrow$\,43.4}} &
90.2{\scriptsize\textcolor{green!40!black}{$\uparrow$\,43.8}} &
83.6{\scriptsize\textcolor{green!40!black}{$\uparrow$\,43.9}} \\
Qwen3-VL-32B-Instruct & 43.7 & 36.2 & 48.6 & 41.6 \\
\midrule
\rowcolor{headergray} \multicolumn{5}{l}{\textit{Medical MLLMs}} \\
LLaVA-Med-7B & 21.9 & 17.4 & 8.6 & 15.2 \\
MedGemma-4B-IT & 46.4 & 33.5 & 29.0 & 34.1 \\
MedGemma-27B-IT & 47.3 & 30.2 & 33.4 & 34.1 \\
HuatuoGPT-Vision-7B & 60.7 & 33.7 & 28.0 & 36.3 \\
\rowcolor{color3} \quad + \textbf{Med-Scout} &
76.6{\scriptsize\textcolor{green!40!black}{$\uparrow$\,15.9}} &
79.3{\scriptsize\textcolor{green!40!black}{$\uparrow$\,45.6}} &
64.0{\scriptsize\textcolor{green!40!black}{$\uparrow$\,36.0}} &
73.8{\scriptsize\textcolor{green!40!black}{$\uparrow$\,37.5}} \\
HuatuoGPT-Vision-34B & 62.9 & 33.9 & 30.8 & 37.7 \\
Lingshu-7B & 60.6 & 29.1 & 21.9 & 31.9 \\
\rowcolor{color4} \quad + \textbf{Med-Scout} &
78.9{\scriptsize\textcolor{green!40!black}{$\uparrow$\,18.3}} &
77.5{\scriptsize\textcolor{green!40!black}{$\uparrow$\,48.4}} &
60.0{\scriptsize\textcolor{green!40!black}{$\uparrow$\,38.1}} &
71.9{\scriptsize\textcolor{green!40!black}{$\uparrow$\,40.0}} \\
Lingshu-32B & 63.6 & 31.8 & 20.4 & 33.3 \\
\bottomrule
\end{tabular*}%
}
\end{sc}
\end{small}
\end{center}
\vskip -0.1in
\end{table*}

\subsection{Evaluation on Report Generation Tasks}
\label{appendix:report_gen_eval}
The comprehensive results on the MIMIC-CXR and IU-Xray benchmarks are listed in Table \ref{tab:report_generation}. Beyond improving general-purpose models, Med-Scout proves effective at elevating already powerful domain specialists. Lingshu-7B, previously established as a SOTA medical MLLM, breaks through its performance ceiling after our alignment. It achieves a new SOTA CIDEr score on MIMIC-CXR, significantly outperforming proprietary commercial models. This indicates that our method provides a complementary geometric capability that is missing even in extensively trained medical models.

\subsection{Task Difficulty Analysis}

To investigate whether the complexity of the geometric proxy tasks contributes to the final post-training performance, we designed a controlled ablation study with three difficulty levels using Direct Mode. We constructed ``Easy'' and ``Medium'' variants of the training dataset by adjusting the complexity of the proxy tasks:

\begin{itemize}[leftmargin=*]
    \item \textbf{Easy Variant:}
    \begin{itemize}
        \item \textit{Scale:} Single crop input ($N=1$), removing the need for multi-scale comparative reasoning.
        \item \textit{Topology:} Simple $1 \times 2$ grid shuffling, requiring only binary relative positioning.
        \item \textit{Anomaly:} Coarse $2 \times 2$ grid (4 patches), making the foreign patch visually prominent.
    \end{itemize}
    
    \item \textbf{Medium Variant:}
    \begin{itemize}
        \item \textit{Scale:} Two crop inputs ($N=2$), introducing limited multi-view context.
        \item \textit{Topology:} $1 \times 4$ linear strip shuffling, increasing sequence length but lacking vertical spatial logic.
        \item \textit{Anomaly:} Intermediate $4 \times 2$ grid (8 patches), requiring moderate attention granularity.
    \end{itemize}
    
    \item \textbf{Hard Variant (Med-Scout Standard):}
    \begin{itemize}
        \item \textit{Scale:} Three hierarchical crops ($N=3$), forcing robust global-local mapping.
        \item \textit{Topology:} $2 \times 2$ grid shuffling, necessitating 2D spatial reasoning (both horizontal and vertical).
        \item \textit{Anomaly:} Fine-grained $4 \times 4$ grid (16 patches) with hard-negative mining, requiring pixel-level scrutiny.
    \end{itemize}
\end{itemize}

\textbf{Results.} As shown in Table \ref{tab:ablation_difficulty_generalization}, while the difficulty level increases from ``Easy'' variants to the ``Hard'' (standard Med-Scout) setting, the model consistently achieves superior accuracy across external benchmarks. The ``Hard'' configuration, which enforces rigorous constraints, proved essential for achieving the best results. This confirms that high-complexity geometric objectives are necessary to prevent models from relying on superficial pattern matching, instead compelling them to master deep, pixel-level visual reasoning.

\begin{table*}[t]
\centering
\caption{Impact of Training Task Difficulty. We evaluate the Qwen3-VL-8B-Instruct model trained with datasets of varying geometric complexity (Easy, Medium, Hard) across radiological and general medical VQA benchmarks. Higher difficulty in proxy tasks consistently leads to superior generalization performance. The best results are highlighted.}
\label{tab:ablation_difficulty_generalization}
\begin{center}
\begin{small}
\begin{sc}
\resizebox{\textwidth}{!}{%
\begin{tabular*}{\textwidth}{@{\extracolsep{\fill}}lllllll}
\toprule
\multirow{2}{*}{Training Difficulty} & \multicolumn{3}{c}{Radiological VQA} & \multicolumn{3}{c}{Generalization} \\
\cmidrule(lr){2-4} \cmidrule(lr){5-7}
 & Rad-VQA & VQA-RAD & SLAKE & PMC-VQA & OmniMedVQA & MedXpertQA \\
\midrule
\rowcolor{headergray} \multicolumn{7}{l}{\textit{Baseline}} \\
Qwen3-VL-8B-Instruct & 41.6 & 63.2 & 69.6 & 43.9 & 42.9 & 30.4 \\
\midrule
\rowcolor{headergray} \multicolumn{7}{l}{\textit{Med-Scout Variants}} \\
Easy Difficulty&
43.9{\scriptsize\textcolor{blue}{$\uparrow$\,2.3}} &
65.1{\scriptsize\textcolor{blue}{$\uparrow$\,1.9}} &
71.4{\scriptsize\textcolor{blue}{$\uparrow$\,1.8}} &
44.2{\scriptsize\textcolor{blue}{$\uparrow$\,0.3}} &
45.1{\scriptsize\textcolor{blue}{$\uparrow$\,2.2}} &
30.5{\scriptsize\textcolor{blue}{$\uparrow$\,0.1}} \\
Medium Difficulty&
44.8{\scriptsize\textcolor{blue}{$\uparrow$\,3.2}} &
64.7{\scriptsize\textcolor{blue}{$\uparrow$\,1.5}} &
71.8{\scriptsize\textcolor{blue}{$\uparrow$\,2.2}} &
45.0{\scriptsize\textcolor{blue}{$\uparrow$\,1.1}} &
45.7{\scriptsize\textcolor{blue}{$\uparrow$\,2.8}} &
30.5{\scriptsize\textcolor{blue}{$\uparrow$\,0.1}} \\
\rowcolor{color4} \textbf{Hard Difficulty (Ours)}&
\textbf{45.3}{\scriptsize\textcolor{green!40!black}{$\uparrow$\,3.7}} &
\textbf{65.8}{\scriptsize\textcolor{green!40!black}{$\uparrow$\,2.6}} &
\textbf{72.0}{\scriptsize\textcolor{green!40!black}{$\uparrow$\,2.4}} &
\textbf{45.5}{\scriptsize\textcolor{green!40!black}{$\uparrow$\,1.6}} &
\textbf{46.0}{\scriptsize\textcolor{green!40!black}{$\uparrow$\,3.1}} &
\textbf{30.8}{\scriptsize\textcolor{green!40!black}{$\uparrow$\,0.4}} \\
\quad $\Delta$ & 
\color{green!40!black}{+0.5} & 
\color{green!40!black}{+0.7} & 
\color{green!40!black}{+0.2} & 
\color{green!40!black}{+0.5} & 
\color{green!40!black}{+0.3} & 
\color{green!40!black}{+0.3} \\
\bottomrule
\end{tabular*}%
}
\end{sc}
\end{small}
\end{center}
\vskip -0.1in
\end{table*}

\subsection{Impact of Proxy Task Types}
\label{appendix:ablation_task_types}
To investigate the distinct contribution of each geometric proxy task to the overall performance, we conducted a comprehensive ablation study. Using Qwen3-VL-8B-Instruct as the backbone, we compared the baseline performance against variants trained with \textit{Single Task Specialists} (using only one task type), \textit{Leave-One-Out} Configurations (removing exactly one task type), and the full Med-Scout framework across six benchmarks.

The results are reported in Table~\ref{tab:ablation_proxy}. We observe that:
\begin{itemize}[leftmargin=*]
    \item \textbf{Every Proxy Task is Indispensable.} 
    Comparing the full Med-Scout framework against the ``Leave-One-Out'' configurations reveals that removing any single geometric task leads to a consistent performance degradation across all benchmarks. This confirms that these three proxy tasks collectively establishing a holistic geometric perception that is superior to the sum of its parts.
    \item \textbf{Topology and Anomaly Tasks Drive Generalization.} 
    The latter two tasks Topological Jigsaw Reconstruction ($\mathcal{T}_\text{topo}$) and Anomaly Consistency Detection ($\mathcal{T}_\text{anom}$) demonstrate a more critical impact on model performance and generalization. This suggests that the high-level logical deduction required for topology and the fine-grained scrutiny needed for anomaly detection are fundamental capabilities that generalize effectively to diverse medical imaging modalities.
\end{itemize}

\begin{table*}[t]
\centering
\caption{Ablation Study of Proxy Task. We use Qwen3-VL-8B-Instruct as the backbone model to evaluate how different geometric tasks contribute to generalization performance across radiological and broad medical domains.}
\label{tab:ablation_proxy}
\begin{center}
\begin{small}
\begin{sc}
\begin{tabular*}{\textwidth}{@{\extracolsep{\fill}}lllllll}
\toprule
\multirow{2}{*}{Training Configuration} & \multicolumn{3}{c}{Radiological VQA} & \multicolumn{3}{c}{Generalization} \\
\cmidrule(lr){2-4} \cmidrule(lr){5-7}
 & Rad-VQA & VQA-RAD & SLAKE & PMC-VQA & OmniMedVQA & MedXpertQA \\
\midrule
\rowcolor{headergray} \multicolumn{7}{l}{\textit{Baseline}} \\
Qwen3-VL-8B-Instruct & 41.6 & 63.2 & 69.6 & 43.9 & 42.9 & 30.4 \\
\midrule
\rowcolor{headergray} \multicolumn{7}{l}{\textit{Single Task Specialists}} \\
\quad + Scale Only &
42.9{\scriptsize\textcolor{blue}{$\uparrow$\,1.3}} &
64.1{\scriptsize\textcolor{blue}{$\uparrow$\,0.9}} &
70.3{\scriptsize\textcolor{blue}{$\uparrow$\,0.7}} &
44.1{\scriptsize\textcolor{blue}{$\uparrow$\,0.2}} &
43.9{\scriptsize\textcolor{blue}{$\uparrow$\,1.0}} &
30.5{\scriptsize\textcolor{blue}{$\uparrow$\,0.1}} \\
\quad + Topology Only &
44.1{\scriptsize\textcolor{blue}{$\uparrow$\,2.5}} &
64.9{\scriptsize\textcolor{blue}{$\uparrow$\,1.7}} &
70.6{\scriptsize\textcolor{blue}{$\uparrow$\,1.0}} &
44.8{\scriptsize\textcolor{blue}{$\uparrow$\,0.9}} &
44.5{\scriptsize\textcolor{blue}{$\uparrow$\,1.6}} &
30.7{\scriptsize\textcolor{blue}{$\uparrow$\,0.3}} \\
\quad + Anomaly Only &
44.6{\scriptsize\textcolor{blue}{$\uparrow$\,3.0}} &
64.9{\scriptsize\textcolor{blue}{$\uparrow$\,1.7}} &
71.0{\scriptsize\textcolor{blue}{$\uparrow$\,1.4}} &
44.4{\scriptsize\textcolor{blue}{$\uparrow$\,0.5}} &
44.7{\scriptsize\textcolor{blue}{$\uparrow$\,1.8}} &
30.6{\scriptsize\textcolor{blue}{$\uparrow$\,0.2}} \\
\midrule
\rowcolor{headergray} \multicolumn{7}{l}{\textit{Leave-One-Out}} \\
\quad + Med-Scout (w/o Scale) &
44.9{\scriptsize\textcolor{blue}{$\uparrow$\,3.3}} &
65.1{\scriptsize\textcolor{blue}{$\uparrow$\,1.9}} &
71.4{\scriptsize\textcolor{blue}{$\uparrow$\,1.8}} &
44.8{\scriptsize\textcolor{blue}{$\uparrow$\,0.9}} &
45.3{\scriptsize\textcolor{blue}{$\uparrow$\,2.4}} &
30.6{\scriptsize\textcolor{blue}{$\uparrow$\,0.2}} \\
\quad + Med-Scout (w/o Topology) &
44.7{\scriptsize\textcolor{blue}{$\uparrow$\,3.1}} &
65.0{\scriptsize\textcolor{blue}{$\uparrow$\,1.8}} &
71.1{\scriptsize\textcolor{blue}{$\uparrow$\,1.5}} &
45.0{\scriptsize\textcolor{blue}{$\uparrow$\,1.1}} &
45.5{\scriptsize\textcolor{blue}{$\uparrow$\,2.6}} &
30.5{\scriptsize\textcolor{blue}{$\uparrow$\,0.1}} \\
\quad + Med-Scout (w/o Anomaly) &
44.8{\scriptsize\textcolor{blue}{$\uparrow$\,3.2}} &
64.8{\scriptsize\textcolor{blue}{$\uparrow$\,1.6}} &
70.8{\scriptsize\textcolor{blue}{$\uparrow$\,1.2}} &
45.2{\scriptsize\textcolor{blue}{$\uparrow$\,1.3}} &
45.4{\scriptsize\textcolor{blue}{$\uparrow$\,2.5}} &
30.7{\scriptsize\textcolor{blue}{$\uparrow$\,0.3}} \\
\midrule
\rowcolor{color4} \textbf{+ Med-Scout (Full)} &
\textbf{45.3}{\scriptsize\textcolor{green!40!black}{$\uparrow$\,3.7}} &
\textbf{65.8}{\scriptsize\textcolor{green!40!black}{$\uparrow$\,2.6}} &
\textbf{72.0}{\scriptsize\textcolor{green!40!black}{$\uparrow$\,2.4}} &
\textbf{45.5}{\scriptsize\textcolor{green!40!black}{$\uparrow$\,1.6}} &
\textbf{46.0}{\scriptsize\textcolor{green!40!black}{$\uparrow$\,3.1}} &
\textbf{30.8}{\scriptsize\textcolor{green!40!black}{$\uparrow$\,0.4}} \\
\quad $\Delta$ & 
\color{green!40!black}{+0.4} & 
\color{green!40!black}{+0.7} & 
\color{green!40!black}{+0.6} & 
\color{green!40!black}{+0.3} & 
\color{green!40!black}{+0.5} & 
\color{green!40!black}{+0.1} \\
\bottomrule
\end{tabular*}
\end{sc}
\end{small}
\end{center}
\vskip -0.1in
\end{table*}






\subsection{Impact of Explicitly Rewarding Logical Validity in CoT}
\label{sec:logical_validity_cot}
To investigate if explicitly rewarding the logical validity of the CoT reasoning process could further enhance the model's geometric perception capabilities, we conducted an ablation study comparing three distinct post-training configurations across multiple benchmarks:
\begin{itemize}[leftmargin=*]
    \item \textbf{(M)}: Med-Scout Direct Mode.
    \item \textbf{(MR)}: Med-Scout Reasoning Mode (enforcing the \texttt{<think>...</think>} format).
    \item \textbf{(MRL)}: Med-Scout Reasoning Mode with an additional explicit reward tailored to evaluate the logical validity of the reasoning steps.
\end{itemize}

As shown in Table~\ref{tab:logical_validity_cot}, incorporating an explicit logic reward fails to yield overall performance growth and occasionally underperforms compared to the direct and standard reasoning modes. This underperformance suggests that enforcing explicit textual reasoning offers very limited benefits for strong visual tasks that fundamentally rely on rigorous, low-level geometric perception rather than semantic logic.

\begin{table*}[h]
\centering
\caption{Performance comparison of explicitly rewarding the logical validity of the CoT process. (M) denotes Direct Mode, (MR) denotes Reasoning Mode with structural reward, and (MRL) denotes Reasoning Mode with an additional explicit logical validity reward. All scores are scaled by a factor of 100.}
\label{tab:logical_validity_cot}
\begin{center}
\begin{small}
\begin{sc}
\renewcommand{\arraystretch}{0.6}
\resizebox{\textwidth}{!}{%
\begin{tabular*}{\textwidth}{@{\extracolsep{\fill}}lllllll}
\toprule
Model & Rad-VQA & VQA-RAD & SLAKE & PMC-VQA & OmniMedVQA & MedXpertQA \\
\midrule
\textbf{Qwen3-VL-8B-Instruct} & 41.6 & 63.2 & 69.6 & 43.9 & 42.9 & 30.4 \\
\quad (M) & 45.3{\scriptsize\textcolor{green!40!black}{$\uparrow$\,3.7}} & 65.8{\scriptsize\textcolor{green!40!black}{$\uparrow$\,2.6}} & 72.0{\scriptsize\textcolor{green!40!black}{$\uparrow$\,2.4}} & 45.5{\scriptsize\textcolor{green!40!black}{$\uparrow$\,1.6}} & 46.0{\scriptsize\textcolor{green!40!black}{$\uparrow$\,3.1}} & 30.8{\scriptsize\textcolor{green!40!black}{$\uparrow$\,0.4}} \\
\quad (MR) & 45.9{\scriptsize\textcolor{green!40!black}{$\uparrow$\,4.3}} & 66.4{\scriptsize\textcolor{green!40!black}{$\uparrow$\,3.2}} & 73.3{\scriptsize\textcolor{green!40!black}{$\uparrow$\,3.7}} & 44.4{\scriptsize\textcolor{green!40!black}{$\uparrow$\,0.5}} & 46.9{\scriptsize\textcolor{green!40!black}{$\uparrow$\,4.0}} & 30.7{\scriptsize\textcolor{green!40!black}{$\uparrow$\,0.3}} \\
\quad (MRL) & 45.2{\scriptsize\textcolor{green!40!black}{$\uparrow$\,3.6}} & 66.1{\scriptsize\textcolor{green!40!black}{$\uparrow$\,2.9}} & 72.7{\scriptsize\textcolor{green!40!black}{$\uparrow$\,3.1}} & 45.0{\scriptsize\textcolor{green!40!black}{$\uparrow$\,1.1}} & 46.6{\scriptsize\textcolor{green!40!black}{$\uparrow$\,3.7}} & 30.8{\scriptsize\textcolor{green!40!black}{$\uparrow$\,0.4}} \\
\midrule
\textbf{HuatuoGPT-Vision-7B} & 48.8 & 67.0 & 67.8 & 53.0 & 75.0 & 22.4 \\
\quad (M) & 52.1{\scriptsize\textcolor{green!40!black}{$\uparrow$\,3.3}} & 70.1{\scriptsize\textcolor{green!40!black}{$\uparrow$\,3.1}} & 71.0{\scriptsize\textcolor{green!40!black}{$\uparrow$\,3.2}} & 55.9{\scriptsize\textcolor{green!40!black}{$\uparrow$\,2.9}} & 75.4{\scriptsize\textcolor{green!40!black}{$\uparrow$\,0.4}} & 22.7{\scriptsize\textcolor{green!40!black}{$\uparrow$\,0.3}} \\
\quad (MR) & 53.0{\scriptsize\textcolor{green!40!black}{$\uparrow$\,4.2}} & 70.3{\scriptsize\textcolor{green!40!black}{$\uparrow$\,3.3}} & 69.8{\scriptsize\textcolor{green!40!black}{$\uparrow$\,2.0}} & 56.8{\scriptsize\textcolor{green!40!black}{$\uparrow$\,3.8}} & 76.2{\scriptsize\textcolor{green!40!black}{$\uparrow$\,1.2}} & 22.7{\scriptsize\textcolor{green!40!black}{$\uparrow$\,0.3}} \\
\quad (MRL) & 53.4{\scriptsize\textcolor{green!40!black}{$\uparrow$\,4.6}} & 70.1{\scriptsize\textcolor{green!40!black}{$\uparrow$\,3.1}} & 71.0{\scriptsize\textcolor{green!40!black}{$\uparrow$\,3.2}} & 57.3{\scriptsize\textcolor{green!40!black}{$\uparrow$\,4.3}} & 75.8{\scriptsize\textcolor{green!40!black}{$\uparrow$\,0.8}} & 22.4{\scriptsize\textcolor{green!40!black}{$\uparrow$\,0.0}} \\
\midrule
\textbf{Lingshu-7B} & 61.2 & 68.9 & 82.8 & 56.3 & 81.4 & 27.4 \\
\quad (M) & 64.0{\scriptsize\textcolor{green!40!black}{$\uparrow$\,2.8}} & 71.0{\scriptsize\textcolor{green!40!black}{$\uparrow$\,2.1}} & 83.0{\scriptsize\textcolor{green!40!black}{$\uparrow$\,0.2}} & 57.4{\scriptsize\textcolor{green!40!black}{$\uparrow$\,1.1}} & 81.9{\scriptsize\textcolor{green!40!black}{$\uparrow$\,0.5}} & 28.0{\scriptsize\textcolor{green!40!black}{$\uparrow$\,0.6}} \\
\quad (MR) & 63.8{\scriptsize\textcolor{green!40!black}{$\uparrow$\,2.6}} & 70.8{\scriptsize\textcolor{green!40!black}{$\uparrow$\,1.9}} & 83.0{\scriptsize\textcolor{green!40!black}{$\uparrow$\,0.2}} & 57.6{\scriptsize\textcolor{green!40!black}{$\uparrow$\,1.3}} & 81.6{\scriptsize\textcolor{green!40!black}{$\uparrow$\,0.2}} & 28.5{\scriptsize\textcolor{green!40!black}{$\uparrow$\,1.1}} \\
\quad (MRL) & 64.0{\scriptsize\textcolor{green!40!black}{$\uparrow$\,2.8}} & 70.3{\scriptsize\textcolor{green!40!black}{$\uparrow$\,1.4}} & 83.6{\scriptsize\textcolor{green!40!black}{$\uparrow$\,0.8}} & 57.8{\scriptsize\textcolor{green!40!black}{$\uparrow$\,1.5}} & 81.9{\scriptsize\textcolor{green!40!black}{$\uparrow$\,0.5}} & 28.7{\scriptsize\textcolor{green!40!black}{$\uparrow$\,1.3}} \\
\bottomrule
\end{tabular*}%
}
\end{sc}
\end{small}
\end{center}
\vskip -0.1in
\end{table*}

\subsection{Reward Mechanism Comparison}

A critical component of the Med-Scout framework is the DGR mechanism, designed to overcome the sparsity of binary feedback in complex reasoning tasks. To quantify its impact, we compared our approach against a standard sparse reward baseline.

\begin{itemize}[leftmargin=*]
    \item \textbf{Sparse Reward Setting:} The model receives a reward of $\mathcal{R}=1$ only if the generated answer perfectly matches the ground truth (e.g., exact index sequence or coordinates within a strict threshold); otherwise, $\mathcal{R}=0$.
    \item \textbf{Dense Reward Setting (Ours):} As detailed in Section~\ref{sec:reward}, we utilize continuous metrics including IoU for bounding boxes, Euclidean distance decay for anomaly detection, and element-wise alignment for topological sequences.
\end{itemize}

Table~\ref{tab:ablation_reward} presents the comparison using the Qwen3-VL-4B-Instruct and Qwen3-VL-8B-Instruct backbones. The results demonstrate that dense geometric reward provides a distinct optimization advantage over binary feedback. While the sparse reward setting already yields notable improvements over the baseline, the DGR mechanism consistently outperforms the sparse variant across all six benchmarks.

This confirms that the granular feedback provided by DGR is crucial for efficient RL post-training. Specifically, DGR awards partial credit for outputs such as ``near-miss'' scale predictions or approximate anomaly locations. In contrast, rigid pass/fail signals often fail to provide useful gradients for partially correct reasoning. Our DGR effectively guides the model to progressively refine its geometric understanding, which leads to superior generalization across both radiological and broad medical domains.

\begin{table*}[t]
\centering
\caption{Impact of Reward Mechanism. Performance comparison between standard sparse reward and our dense geometric reward on six benchmarks. The dense mechanism provides granular feedback, leading to significantly better generalization.}
\label{tab:ablation_reward}
\begin{center}
\begin{small}
\begin{sc}
\resizebox{\textwidth}{!}{%
\begin{tabular*}{\textwidth}{@{\extracolsep{\fill}}lllllll}
\toprule
\multirow{2}{*}{Model \& Reward Strategy} & \multicolumn{3}{c}{Radiological VQA} & \multicolumn{3}{c}{Generalization} \\
\cmidrule(lr){2-4} \cmidrule(lr){5-7}
 & Rad-VQA & VQA-RAD & SLAKE & PMC-VQA & OmniMedVQA & MedXpertQA \\
\midrule
\rowcolor{headergray} \multicolumn{7}{l}{\textbf{Qwen3-VL-4B-Instruct}} \\
Baseline & 41.5 & 59.9 & 73.4 & 42.8 & 45.5 & 27.0 \\
\quad w/ Sparse Reward &
45.1{\scriptsize\textcolor{blue}{$\uparrow$\,3.6}} &
62.0{\scriptsize\textcolor{blue}{$\uparrow$\,2.1}} &
75.3{\scriptsize\textcolor{blue}{$\uparrow$\,1.9}} &
44.7{\scriptsize\textcolor{blue}{$\uparrow$\,1.9}} &
48.6{\scriptsize\textcolor{blue}{$\uparrow$\,3.1}} &
27.3{\scriptsize\textcolor{blue}{$\uparrow$\,0.3}} \\
\rowcolor{color4} \quad \textbf{w/ DGR (Ours)} & 
\textbf{45.7}{\scriptsize\textcolor{green!40!black}{$\uparrow$\,4.2}} &
\textbf{62.9}{\scriptsize\textcolor{green!40!black}{$\uparrow$\,3.0}} &
\textbf{75.6}{\scriptsize\textcolor{green!40!black}{$\uparrow$\,2.2}} &
\textbf{45.1}{\scriptsize\textcolor{green!40!black}{$\uparrow$\,2.3}} &
\textbf{48.8}{\scriptsize\textcolor{green!40!black}{$\uparrow$\,3.3}} &
\textbf{27.7}{\scriptsize\textcolor{green!40!black}{$\uparrow$\,0.7}} \\
\quad $\Delta$ (DGR vs Sparse) & 
\color{green!40!black}{+0.6} & 
\color{green!40!black}{+0.9} & 
\color{green!40!black}{+0.3} & 
\color{green!40!black}{+0.4} & 
\color{green!40!black}{+0.2} & 
\color{green!40!black}{+0.4} \\
\midrule
\rowcolor{headergray} \multicolumn{7}{l}{\textbf{Qwen3-VL-8B-Instruct}} \\
Baseline & 41.6 & 63.2 & 69.6 & 43.9 & 42.9 & 30.4 \\
\quad w/ Sparse Reward &
44.7{\scriptsize\textcolor{blue}{$\uparrow$\,3.1}} &
65.2{\scriptsize\textcolor{blue}{$\uparrow$\,2.0}} &
71.3{\scriptsize\textcolor{blue}{$\uparrow$\,1.7}} &
45.0{\scriptsize\textcolor{blue}{$\uparrow$\,1.1}} &
45.8{\scriptsize\textcolor{blue}{$\uparrow$\,2.9}} &
30.8{\scriptsize\textcolor{blue}{$\uparrow$\,0.4}} \\
\rowcolor{color4} \quad\textbf{w/ DGR (Ours)} & 
\textbf{45.3}{\scriptsize\textcolor{green!40!black}{$\uparrow$\,3.7}} &
\textbf{65.8}{\scriptsize\textcolor{green!40!black}{$\uparrow$\,2.6}} &
\textbf{72.0}{\scriptsize\textcolor{green!40!black}{$\uparrow$\,2.4}} &
\textbf{45.5}{\scriptsize\textcolor{green!40!black}{$\uparrow$\,1.6}} &
\textbf{46.0}{\scriptsize\textcolor{green!40!black}{$\uparrow$\,3.1}} &
\textbf{30.8}{\scriptsize\textcolor{green!40!black}{$\uparrow$\,0.4}} \\
\quad $\Delta$ (DGR vs Sparse) & 
\color{green!40!black}{+0.6} & 
\color{green!40!black}{+0.6} & 
\color{green!40!black}{+0.7} & 
\color{green!40!black}{+0.5} & 
\color{green!40!black}{+0.2} & 
\color{green!40!black}{+0.0} \\
\bottomrule
\end{tabular*}%
}
\end{sc}
\end{small}
\end{center}
\vskip -0.1in
\end{table*}

\subsection{SFT vs. RL}
\label{appendix:sft_vs_rl}
We observe a distinct contrast between internal alignment scores and external generalization capabilities:
\begin{itemize}[leftmargin=*]
    \item \textbf{SFT Achieves Strong Performance on Internal Validation.} 
    On the internal Med-Scout-Bench (Table~\ref{tab:sft_vs_rl_medscoutbench}), SFT demonstrates remarkable efficacy. It achieves performance levels comparable to or even surpassing the RL-tuned models. For example, HuatuoGPT-Vision-7B achieves an average score of 74.6\% with SFT, compared to 73.8\% with RL. This indicates that models can easily master the output syntax and specific data patterns of the proxy tasks through imitation.
    \item \textbf{RL Enables True Generalization.} 
    However, the apparent competence of SFT collapses on external benchmarks (Table~\ref{tab:sft_vs_rl_allbench}). SFT variants exhibit negligible or even negative performance shifts. For instance, Qwen3-VL-8B-Instruct drops by 0.7\% on Rad-VQA and 0.3\% on MedXpertQA. This reveals that SFT merely overfits to the proxy task patterns without internalizing the underlying geometric reasoning. In contrast, RL achieves consistent gains across all external benchmarks. This confirms that exploration-driven optimization is essential for cultivating a generalized geometric perception that transfers beyond the training data.
\end{itemize}

\begin{table*}[t]
\centering
\caption{Comparison of SFT vs. RL on Med-Scout-Bench.}
\label{tab:sft_vs_rl_medscoutbench}
\begin{center}
\begin{small}
\begin{sc}
\resizebox{\textwidth}{!}{%
\begin{tabular*}{\textwidth}{@{\extracolsep{\fill}}lllll}
\toprule
\multirow{2}{*}{Model / Method} & \multicolumn{4}{c}{Med-Scout-Bench} \\
\cmidrule(lr){2-5}
 & Task A & Task B & Task C & \textbf{Avg.} \\
\midrule
\rowcolor{headergray} \multicolumn{5}{l}{\textit{Qwen3-VL-4B-Instruct}} \\
\quad Baseline & 59.6 & 36.6 & 31.3 & 38.7 \\
\quad + Med-Scout (SFT) & 
89.9{\scriptsize\textcolor{blue}{$\uparrow$\,30.3}} & 
\textbf{78.1}{\scriptsize\textcolor{blue}{$\uparrow$\,41.5}} & 
84.6{\scriptsize\textcolor{blue}{$\uparrow$\,53.3}} & 
82.2{\scriptsize\textcolor{blue}{$\uparrow$\,43.5}} \\
\rowcolor{color1} \quad \textbf{+ Med-Scout (RL)} & 
\textbf{94.4}{\scriptsize\textcolor{green!40!black}{$\uparrow$\,34.8}} & 
77.5{\scriptsize\textcolor{green!40!black}{$\uparrow$\,40.9}} & 
\textbf{89.8}{\scriptsize\textcolor{green!40!black}{$\uparrow$\,58.5}} & 
\textbf{84.4}{\scriptsize\textcolor{green!40!black}{$\uparrow$\,45.7}} \\
\quad $\Delta$ & 
\color{green!40!black}{+4.5} &
\color{red}{-0.6} &
\color{green!40!black}{+5.2} & 
\color{green!40!black}{+2.2} \\
\midrule
\rowcolor{headergray} \multicolumn{5}{l}{\textit{Qwen3-VL-8B-Instruct}} \\
\quad Baseline & 41.0 & 34.7 & 46.4 & 39.7 \\
\quad + Med-Scout (SFT) & 
85.1{\scriptsize\textcolor{blue}{$\uparrow$\,44.1}} & 
76.9{\scriptsize\textcolor{blue}{$\uparrow$\,42.2}} & 
88.7{\scriptsize\textcolor{blue}{$\uparrow$\,42.3}} & 
82.2{\scriptsize\textcolor{blue}{$\uparrow$\,42.5}} \\
\rowcolor{color2} \quad \textbf{+ Med-Scout (RL)} & 
\textbf{86.7}{\scriptsize\textcolor{green!40!black}{$\uparrow$\,45.7}} & 
\textbf{78.1}{\scriptsize\textcolor{green!40!black}{$\uparrow$\,43.4}} & 
\textbf{90.2}{\scriptsize\textcolor{green!40!black}{$\uparrow$\,43.8}} & 
\textbf{83.6}{\scriptsize\textcolor{green!40!black}{$\uparrow$\,43.9}} \\
\quad $\Delta$ & 
\color{green!40!black}{+1.6} &
\color{green!40!black}{+1.2} &
\color{green!40!black}{+1.5} & 
\color{green!40!black}{+1.4} \\
\midrule
\rowcolor{headergray} \multicolumn{5}{l}{\textit{HuatuoGPT-Vision-7B}} \\
\quad Baseline & 60.7 & 33.7 & 28.0 & 36.3 \\
\quad + Med-Scout (SFT) & 
\textbf{79.2}{\scriptsize\textcolor{blue}{$\uparrow$\,18.5}} & 
79.3{\scriptsize\textcolor{blue}{$\uparrow$\,45.6}} & 
\textbf{65.1}{\scriptsize\textcolor{blue}{$\uparrow$\,37.1}} & 
\textbf{74.6}{\scriptsize\textcolor{blue}{$\uparrow$\,38.3}} \\
\rowcolor{color3} \quad \textbf{+ Med-Scout (RL)} & 
76.6{\scriptsize\textcolor{green!40!black}{$\uparrow$\,15.9}} & 
\textbf{79.3}{\scriptsize\textcolor{green!40!black}{$\uparrow$\,45.6}} & 
64.0{\scriptsize\textcolor{green!40!black}{$\uparrow$\,36.0}} & 
73.8{\scriptsize\textcolor{green!40!black}{$\uparrow$\,37.5}} \\
\quad $\Delta$ & 
\color{red}{-2.6} &
\color{green!40!black}{+0.0} &
\color{red}{-1.1} & 
\color{red}{-0.8} \\
\midrule
\rowcolor{headergray} \multicolumn{5}{l}{\textit{Lingshu-7B}} \\
\quad Baseline & 60.6 & 29.1 & 21.9 & 31.9 \\
\quad + Med-Scout (SFT) & 
72.8{\scriptsize\textcolor{blue}{$\uparrow$\,12.2}} & 
\textbf{78.5}{\scriptsize\textcolor{blue}{$\uparrow$\,49.4}} & 
58.0{\scriptsize\textcolor{blue}{$\uparrow$\,36.1}} & 
70.7{\scriptsize\textcolor{blue}{$\uparrow$\,38.8}} \\
\rowcolor{color4} \quad \textbf{+ Med-Scout (RL)} & 
\textbf{78.9}{\scriptsize\textcolor{green!40!black}{$\uparrow$\,18.3}} & 
77.5{\scriptsize\textcolor{green!40!black}{$\uparrow$\,48.4}} & 
\textbf{60.0}{\scriptsize\textcolor{green!40!black}{$\uparrow$\,38.1}} & 
\textbf{71.9}{\scriptsize\textcolor{green!40!black}{$\uparrow$\,40.0}} \\
\quad $\Delta$ & 
\color{green!40!black}{+6.1} &
\color{red}{-1.0} &
\color{green!40!black}{+2.0} & 
\color{green!40!black}{+1.2} \\
\bottomrule
\end{tabular*}%
}
\end{sc}
\end{small}
\end{center}
\end{table*}

\begin{table*}[t]
\centering
\caption{Comparison of SFT vs. RL on six external benchmarks.}
\label{tab:sft_vs_rl_allbench}
\begin{center}
\begin{small}
\begin{sc}
\resizebox{\textwidth}{!}{%
\begin{tabular*}{\textwidth}{@{\extracolsep{\fill}}lllllll}
\toprule
\multirow{2}{*}{Model / Method} & \multicolumn{3}{c}{Radiological VQA} & \multicolumn{3}{c}{Generalization} \\
\cmidrule(lr){2-4} \cmidrule(lr){5-7}
 & Rad-VQA & VQA-RAD & SLAKE & PMC-VQA & OmniMedVQA & MedXpertQA \\
\midrule
\rowcolor{headergray} \multicolumn{7}{l}{\textit{Qwen3-VL-4B-Instruct}} \\
\quad Baseline & 41.5 & 59.9 & 73.4 & 42.8 & 45.5 & 27.0 \\
\quad + Med-Scout (SFT) & 
41.6{\scriptsize\textcolor{blue}{$\uparrow$\,0.1}} & 
59.3{\scriptsize\textcolor{red}{$\downarrow$\,0.6}} & 
73.3{\scriptsize\textcolor{red}{$\downarrow$\,0.1}} & 
42.8{\scriptsize\textcolor{blue}{$\uparrow$\,0.0}} & 
45.7{\scriptsize\textcolor{blue}{$\uparrow$\,0.2}} & 
26.8{\scriptsize\textcolor{red}{$\downarrow$\,0.2}} \\
\rowcolor{color1} \quad \textbf{+ Med-Scout (RL)} & 
\textbf{45.7}{\scriptsize\textcolor{green!40!black}{$\uparrow$\,4.2}} & 
\textbf{62.9}{\scriptsize\textcolor{green!40!black}{$\uparrow$\,3.0}} & 
\textbf{75.6}{\scriptsize\textcolor{green!40!black}{$\uparrow$\,2.2}} & 
\textbf{45.1}{\scriptsize\textcolor{green!40!black}{$\uparrow$\,2.3}} & 
\textbf{48.8}{\scriptsize\textcolor{green!40!black}{$\uparrow$\,3.3}} & 
\textbf{27.7}{\scriptsize\textcolor{green!40!black}{$\uparrow$\,0.7}} \\
\quad $\Delta$ & 
\color{green!40!black}{+4.1} &
\color{green!40!black}{+3.6} &
\color{green!40!black}{+2.3} & 
\color{green!40!black}{+2.3} &
\color{green!40!black}{+3.1} &
\color{green!40!black}{+0.9} \\
\midrule
\rowcolor{headergray} \multicolumn{7}{l}{\textit{Qwen3-VL-8B-Instruct}} \\
\quad Baseline & 41.6 & 63.2 & 69.6 & 43.9 & 42.9 & 30.4 \\
\quad + Med-Scout (SFT) & 
40.9{\scriptsize\textcolor{red}{$\downarrow$\,0.7}} & 
63.1{\scriptsize\textcolor{red}{$\downarrow$\,0.1}} & 
69.8{\scriptsize\textcolor{blue}{$\uparrow$\,0.2}} & 
44.5{\scriptsize\textcolor{blue}{$\uparrow$\,0.6}} & 
42.7{\scriptsize\textcolor{red}{$\downarrow$\,0.2}} & 
30.1{\scriptsize\textcolor{red}{$\downarrow$\,0.3}} \\
\rowcolor{color2} \quad \textbf{+ Med-Scout (RL)} & 
\textbf{45.3}{\scriptsize\textcolor{green!40!black}{$\uparrow$\,3.7}} & 
\textbf{65.8}{\scriptsize\textcolor{green!40!black}{$\uparrow$\,2.6}} & 
\textbf{72.0}{\scriptsize\textcolor{green!40!black}{$\uparrow$\,2.4}} & 
\textbf{45.5}{\scriptsize\textcolor{green!40!black}{$\uparrow$\,1.6}} & 
\textbf{46.0}{\scriptsize\textcolor{green!40!black}{$\uparrow$\,3.1}} & 
\textbf{30.8}{\scriptsize\textcolor{green!40!black}{$\uparrow$\,0.4}} \\
\quad $\Delta$ & 
\color{green!40!black}{+4.4} &
\color{green!40!black}{+2.7} &
\color{green!40!black}{+2.2} & 
\color{green!40!black}{+1.0} &
\color{green!40!black}{+3.3} &
\color{green!40!black}{+0.7} \\
\midrule
\rowcolor{headergray} \multicolumn{7}{l}{\textit{HuatuoGPT-Vision-7B}} \\
\quad Baseline & 48.8 & 67.0 & 67.8 & 53.0 & 75.0 & 22.4 \\
\quad + Med-Scout (SFT) & 
48.5{\scriptsize\textcolor{red}{$\downarrow$\,0.3}} & 
67.1{\scriptsize\textcolor{blue}{$\uparrow$\,0.1}} & 
68.0{\scriptsize\textcolor{blue}{$\uparrow$\,0.2}} & 
52.6{\scriptsize\textcolor{red}{$\downarrow$\,0.4}} & 
75.1{\scriptsize\textcolor{blue}{$\uparrow$\,0.1}} & 
22.1{\scriptsize\textcolor{red}{$\downarrow$\,0.3}} \\
\rowcolor{color3} \quad \textbf{+ Med-Scout (RL)} & 
\textbf{52.1}{\scriptsize\textcolor{green!40!black}{$\uparrow$\,3.3}} & 
\textbf{70.1}{\scriptsize\textcolor{green!40!black}{$\uparrow$\,3.1}} & 
\textbf{71.0}{\scriptsize\textcolor{green!40!black}{$\uparrow$\,3.2}} & 
\textbf{55.9}{\scriptsize\textcolor{green!40!black}{$\uparrow$\,2.9}} & 
\textbf{75.4}{\scriptsize\textcolor{green!40!black}{$\uparrow$\,0.4}} & 
\textbf{22.7}{\scriptsize\textcolor{green!40!black}{$\uparrow$\,0.3}} \\
\quad $\Delta$ & 
\color{green!40!black}{+3.6} &
\color{green!40!black}{+3.0} &
\color{green!40!black}{+3.0} & 
\color{green!40!black}{+3.3} &
\color{green!40!black}{+0.3} &
\color{green!40!black}{+0.6} \\
\midrule
\rowcolor{headergray} \multicolumn{7}{l}{\textit{Lingshu-7B}} \\
\quad Baseline & 61.2 & 68.9 & 82.8 & 56.3 & 81.4 & 27.4 \\
\quad + Med-Scout (SFT) & 
61.3{\scriptsize\textcolor{blue}{$\uparrow$\,0.1}} & 
69.0{\scriptsize\textcolor{blue}{$\uparrow$\,0.1}} & 
82.9{\scriptsize\textcolor{blue}{$\uparrow$\,0.1}} & 
55.9{\scriptsize\textcolor{red}{$\downarrow$\,0.4}} & 
81.1{\scriptsize\textcolor{red}{$\downarrow$\,0.3}} & 
27.0{\scriptsize\textcolor{red}{$\downarrow$\,0.4}} \\
\rowcolor{color4} \quad \textbf{+ Med-Scout (RL)} & 
\textbf{64.0}{\scriptsize\textcolor{green!40!black}{$\uparrow$\,2.8}} & 
\textbf{71.0}{\scriptsize\textcolor{green!40!black}{$\uparrow$\,2.1}} & 
\textbf{83.0}{\scriptsize\textcolor{green!40!black}{$\uparrow$\,0.2}} & 
\textbf{57.4}{\scriptsize\textcolor{green!40!black}{$\uparrow$\,1.1}} & 
\textbf{81.9}{\scriptsize\textcolor{green!40!black}{$\uparrow$\,0.5}} & 
\textbf{28.0}{\scriptsize\textcolor{green!40!black}{$\uparrow$\,0.6}} \\
\quad $\Delta$ & 
\color{green!40!black}{+2.7} &
\color{green!40!black}{+2.0} &
\color{green!40!black}{+0.1} & 
\color{green!40!black}{+1.5} &
\color{green!40!black}{+0.8} &
\color{green!40!black}{+1.0} \\
\bottomrule
\end{tabular*}%
}
\end{sc}
\end{small}
\end{center}
\vskip -0.1in
\end{table*}

\subsection{Case Study}
\label{sec:case_study}

\begin{figure*}[h]
    \centering
    \includegraphics[width=0.7\linewidth]{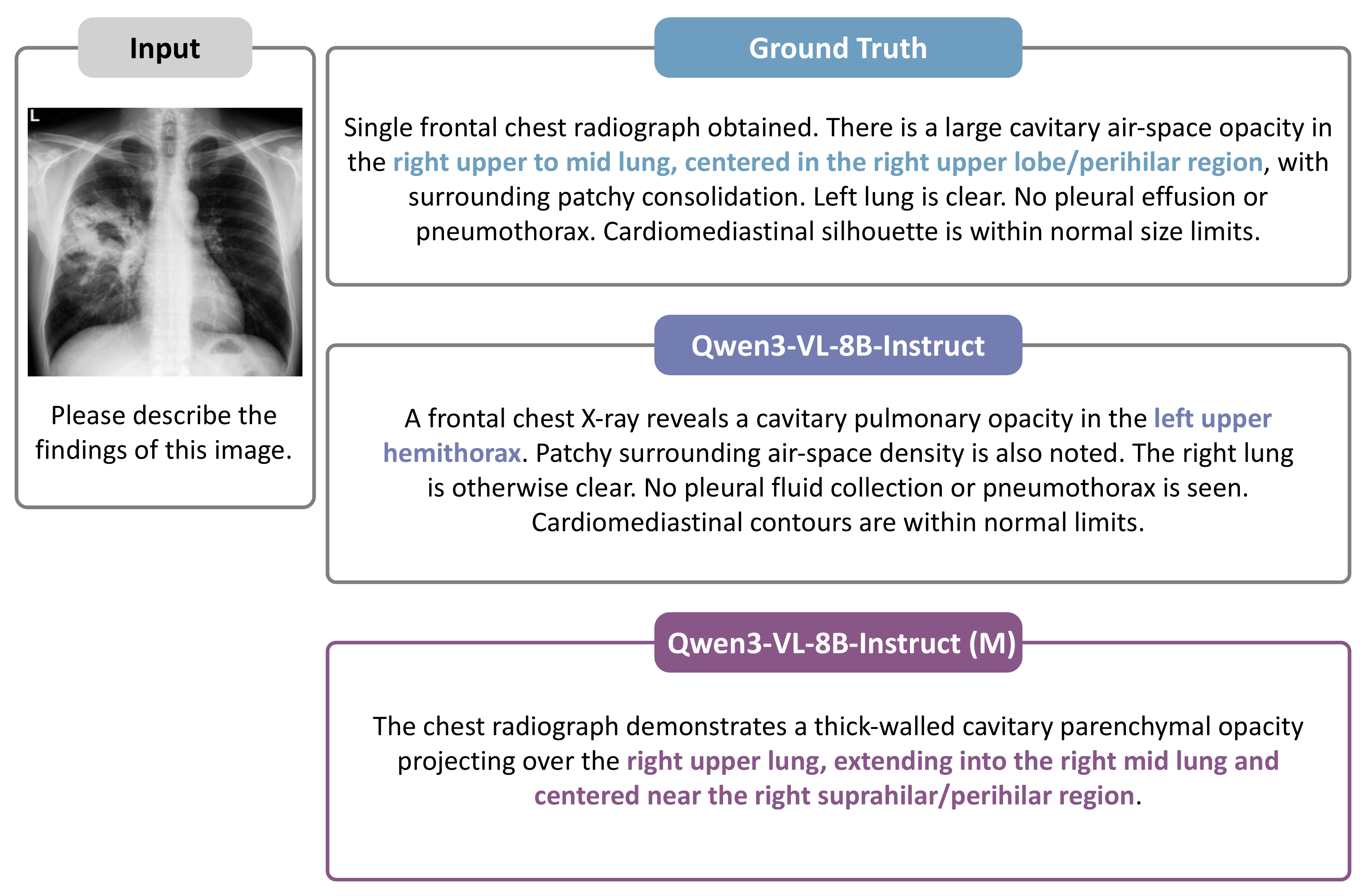}
    \caption{Case Study 1. While the baseline model correctly identifies the pathology semantics, it suffers from severe geometric blindness, mislocalizing the findings to the opposite side of the body. Med-Scout corrects these spatial hallucinations, proving its capability to ground reasoning in intrinsic physical logic.}
    \label{fig:case_study1}
    \vskip -0.1in
\end{figure*}

\begin{figure*}[h]
    \centering
    \includegraphics[width=0.7\linewidth]{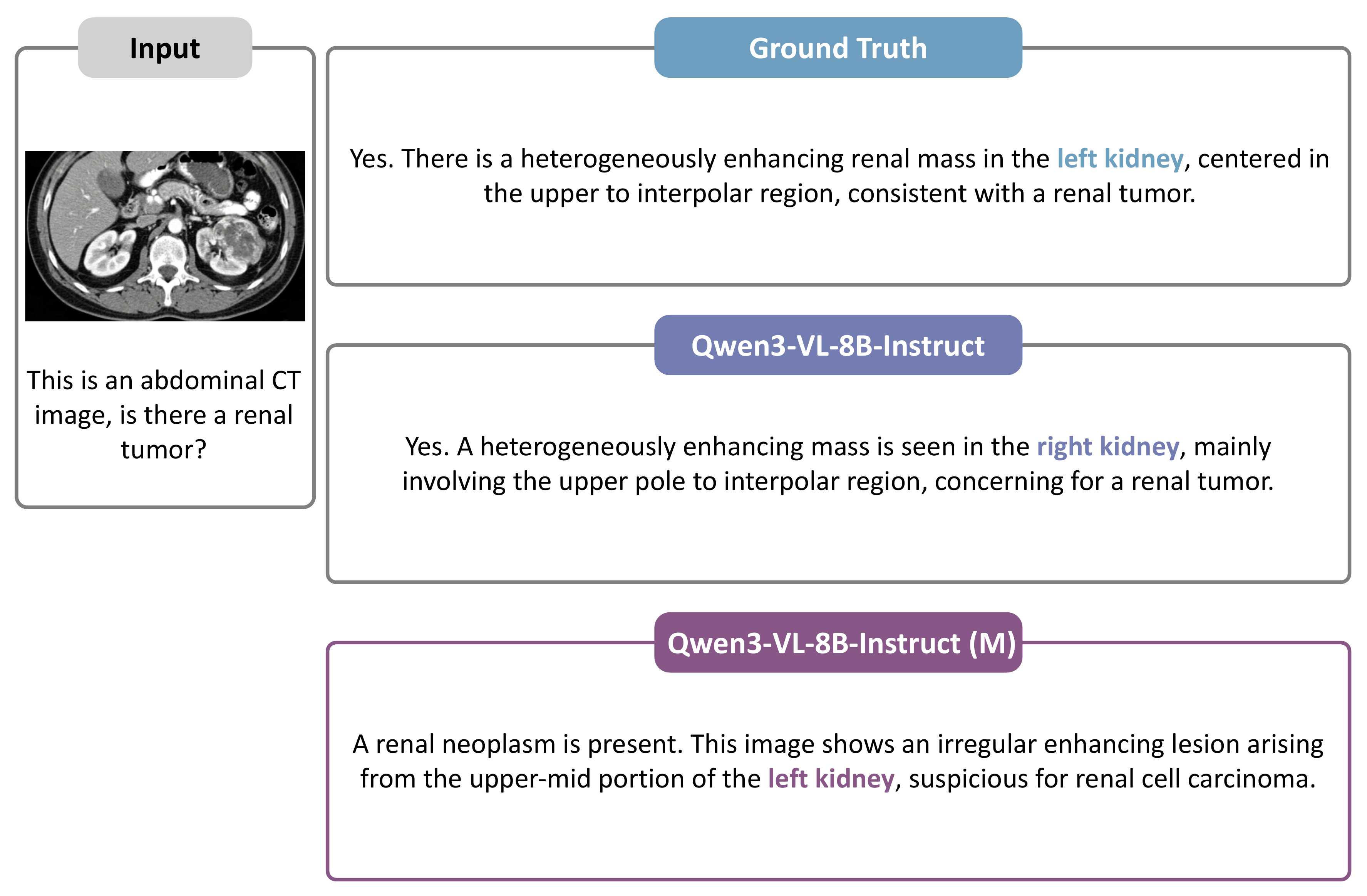}
    \caption{Case Study 2. While the baseline model correctly identifies the pathology semantics, it suffers from severe geometric blindness, mislocalizing the findings to the opposite side of the body. Med-Scout corrects these spatial hallucinations, proving its capability to ground reasoning in intrinsic physical logic.}
    \label{fig:case_study2}
    \vskip -0.1in
\end{figure*}

To qualitatively demonstrate that our framework effectively compels models to ground their reasoning in intrinsic physical logic, we present two representative case studies in Figure~\ref{fig:case_study1} and Figure~\ref{fig:case_study2}. 

As illustrated in the examples, the baseline Qwen3-VL-8B-Instruct model demonstrates a classic symptom of geometric blindness. In both the chest radiograph (Figure~\ref{fig:case_study1}, top) and the abdominal CT (Figure~\ref{fig:case_study2}, bottom), the baseline model successfully identifies the high-level semantic features of the pathologies, generating accurate clinical terms such as ``heterogeneously enhancing mass'' and ``cavitary pulmonary opacity.'' However, it completely fails in spatial grounding, mislocalizing the lesions to the contralateral side (hallucinating ``right kidney'' instead of the left, and ``left upper hemithorax'' instead of the right). This spatial inversion indicates that the baseline model relies heavily on statistical language priors rather than true visual-spatial perception. Conversely, the model aligned with Med-Scout completely rectifies these topological errors. It provides precise, geometrically faithful localizations. This stark contrast compellingly confirms that Med-Scout effectively cures spatial hallucinations, successfully aligning the model's semantic text generation with the objective, intrinsic physical logic of the medical images.

\section{Theoretical Analysis: Geometric Manifold Alignment}
\label{sec:appendix_theory}

To rigorously justify that Med-Scout cures geometric blindness rather than merely overfitting to the proxy task templates, we analyze the learning process through the lens of Manifold Learning and Energy-Based Models (EBMs).

\noindent\textbf{Formulation: Geometric Blindness as Manifold Deviation.}
Let the space of valid medical visual-text pairs lie on a high-dimensional manifold $\mathcal{M}_{geo} \subset \mathcal{X} \times \mathcal{Y}$. A pair $(x, y)$ is geometrically valid if and only if it satisfies a set of intrinsic physical constraints $\mathcal{C}$ (e.g., anatomical topology, scale consistency):
\begin{equation}
    \mathcal{M}_\text{geo} = \{(x, y) \mid \mathcal{C}_\text{scale}(x,y) \land \mathcal{C}_\text{topo}(x,y) \land \mathcal{C}_\text{anom}(x,y) = \text{True} \}
\end{equation}
\textit{Geometric Blindness} occurs when an MLLM learns a strictly semantic distribution $P_{\theta}(y|x)$ that covers a broader, ``hallucinated'' manifold $\mathcal{M}_\text{halluc} \supset \mathcal{M}_\text{geo}$. Within $\mathcal{M}_\text{halluc}$, plausible but geometrically impossible descriptions (e.g., ``liver on the left'') are assigned high probability (low energy), indistinguishable from factual descriptions.

\noindent\textbf{Proxy Tasks as Manifold Constraints.}
Our three proxy tasks function not as simple Q\&A pairs, but as constraint operators that explicitly penalize deviations from $\mathcal{M}_\text{geo}$. By optimizing the dense geometric reward, we are essentially minimizing the energy of the model distribution specifically on the manifold $\mathcal{M}_\text{geo}$. The objective of Med-Scout is to reshape the energy landscape $E(x,y) = -\log P_{\theta}(y|x)$ such that:
\begin{equation}
    E(x, y_\text{halluc}) \gg E(x, y_\text{truth}), \quad \forall y_\text{halluc} \notin \mathcal{M}_\text{geo}
\end{equation}
Specifically, the tasks enforce multi-scale correspondence ($\mathcal{T}_\text{scale}$), topological integrity ($\mathcal{T}_\text{topo}$), and fine-grained structural consistency ($\mathcal{T}_\text{anom}$).

\noindent\textbf{Proof of True Grounding via Energy Landscapes.}
If the model were merely overfitting to the templates of the proxy tasks (e.g., memorizing specific grid indices), the energy landscape reshifting would be confined strictly to the subspace of those templates. However, our empirical analysis on natural language reports (Figure~\ref{fig:energy_landscape_comparison}) proves this is not the case.

In Figure~\ref{fig:energy_landscape_comparison}, we utilized a probe dataset of factual vs. counterfactual reports derived from MIMIC-CXR.
\begin{itemize}
    \item \textbf{Baseline (Blindness):} The overlapping energy distributions indicate that the baseline model treats the true manifold and the hallucinated space $\mathcal{M}_\text{halluc}$ as equiprobable.
    \item \textbf{Med-Scout (Aligned):} The emergence of a distinct energy barrier on this natural language task serves as a theoretical certificate. It demonstrates that the constraints learned from the proxy tasks have successfully propagated to the general probability density function of the model.
\end{itemize}
This confirms that Med-Scout has successfully internalized the intrinsic boundaries of the medical geometric manifold $\mathcal{M}_\text{geo}$, rather than merely minimizing loss on a specific set of training artifacts.

\section{Perspectives}
\label{sec:limitations}

While Med-Scout demonstrates significant improvements in curing geometric blindness, this study also opens several promising directions for future research. Beyond simply scaling models or expanding datasets, we believe the broader value of Med-Scout lies in establishing a clinically inspired paradigm for teaching MLLMs to perceive medical images through verifiable visual logic.

\noindent\textbf{Validation on Larger Model Scales.}
Our experiments were primarily conducted on MLLMs with parameters ranging from 3B to 8B, including Qwen3-VL-4B/8B, Lingshu-7B, and HuatuoGPT-Vision-7B. This choice was mainly driven by computational resource constraints rather than any inherent limitation of the proposed framework. Med-Scout is designed to be model-size agnostic, since its core supervision comes from geometric constraints derived from the image itself rather than from model-specific architectures. Therefore, the proposed geometric alignment principles can be naturally applied to larger-scale foundation models, such as 70B-level or even stronger MLLMs.

Scaling Med-Scout to larger models may bring two complementary benefits. On the one hand, larger models usually possess stronger semantic priors and broader medical knowledge, which may help them integrate geometric evidence with clinical concepts more effectively. On the other hand, Med-Scout can regularize such models by forcing their language generation to respect objective visual facts, thereby reducing the risk that stronger linguistic ability leads to more fluent but geometrically incorrect hallucinations. This suggests an important future direction: combining the rich knowledge of large foundation models with explicit geometry-aware post-training to build medical MLLMs that are not only knowledgeable, but also visually faithful.

\noindent\textbf{Learning from Clinical Reading Logic.}
A central insight of Med-Scout is that geometric perception should not be treated as an isolated low-level visual skill. In clinical practice, physicians rarely read medical images by passively recognizing isolated visual patterns. Instead, they follow a systematic reasoning process: they first build a global anatomical impression, then zoom into suspicious regions, compare local structures with surrounding tissues or adjacent slices, and finally verify whether the observed findings are consistent with anatomical topology and clinical priors. This structured reading behavior is an important source of inspiration for Med-Scout.

The three proxy tasks in Med-Scout can be viewed as computational abstractions of these clinical habits. Hierarchical scale localization mimics the clinician's global-to-local reading process, where a finding must be anchored in both the whole image and a local region. Topological jigsaw reconstruction reflects the use of anatomical layout and spatial continuity to infer whether structures are placed correctly. Anomaly consistency detection resembles comparative scrutiny, where physicians identify subtle discontinuities by comparing a suspicious region with nearby tissues, adjacent slices, or visually similar references. In this sense, Med-Scout does not merely construct artificial pretext tasks; it distills clinical visual reasoning into objective and verifiable learning signals.

This perspective points to a broader research direction: future medical MLLMs should learn not only from diagnostic labels or textual reports, but also from the cognitive procedures used by clinicians during image interpretation. Many clinical reasoning patterns can potentially be transformed into proxy tasks, such as tracing anatomical continuity across slices, checking left-right symmetry, comparing temporal changes across follow-up scans, verifying lesion-organ relationships, or distinguishing true abnormalities from imaging artifacts. By converting these human reading strategies into scalable training objectives, future work can move beyond imitation of final clinical conclusions and instead teach MLLMs how to observe, compare, verify, and reason like medical experts.

\noindent\textbf{From Clinical Insight to MLLM Optimization.}
Another important future direction is to systematically bridge clinical insight and MLLM optimization. Clinical reasoning is often implicit, experience-driven, and difficult to annotate at scale. However, many of its underlying principles are grounded in visual regularities that already exist within unlabeled medical images. Med-Scout shows that these regularities can be mined through proxy tasks and optimized with dense rewards. This provides a practical pathway for transforming expert reading logic into machine-learnable supervision without requiring exhaustive manual annotation.

Future work can further expand this idea by designing a richer library of clinically grounded proxy tasks. For example, temporal comparison tasks can be derived from longitudinal scans to teach progression reasoning; cross-slice consistency tasks can be constructed from volumetric CT and MRI to enhance 3D anatomical understanding; symmetry-based tasks can encourage models to detect unilateral abnormalities; and report-image consistency tasks can help align textual descriptions with precise visual evidence. These tasks would allow MLLMs to internalize the intermediate steps of clinical perception, rather than merely optimizing for final answer accuracy. Such a direction may help close the gap between language-based medical knowledge and the perceptual discipline required for reliable clinical decision support.

\noindent\textbf{Scope of Medical Modalities.}
Our current training data and Med-Scout-Bench focus on CT, MRI, and X-ray images. We selected these three modalities because they are representative forms of structural medical imaging and contain strong geometric constraints, including anatomical scale, spatial topology, and structural continuity. However, the core philosophy of Med-Scout, namely mining intrinsic visual logic from unlabeled data through clinically grounded proxy tasks, is not limited to these modalities.

The same principle can be extended to a broader spectrum of medical imaging. In pathology, models may learn to reason across different magnifications in whole-slide images, reflecting how pathologists switch between low-power tissue organization and high-power cellular details. In ultrasound, proxy tasks may focus on structural continuity, boundary consistency, and view-dependent anatomical changes. In dermoscopy, models may benefit from tasks that emphasize lesion symmetry, border irregularity, and local texture consistency. These extensions require modality-specific task designs, but the underlying goal remains the same: to extract reliable visual reasoning signals from the internal structure of medical images.

\noindent\textbf{Toward Clinically Faithful Medical MLLMs.}
Ultimately, Med-Scout suggests that the next stage of medical MLLM development should move from answer imitation toward perception alignment. A clinically reliable model should not only generate plausible medical language, but also ground each conclusion in the physical and geometric truth of the image. This requires models to develop habits similar to clinical image reading: observing at multiple scales, respecting anatomical topology, comparing subtle differences, and verifying visual consistency before producing a conclusion.

By demonstrating that such habits can be approximated through unlabeled data, proxy tasks, and geometry-aware reinforcement learning, Med-Scout provides a scalable step toward this goal. We envision future medical MLLMs as systems that combine three forms of intelligence: the semantic knowledge of foundation models, the perceptual discipline of geometry-aware optimization, and the structured reasoning logic inspired by clinicians. This integration may lead to medical AI systems that are more robust, more interpretable, and more faithful to clinical visual evidence, thereby providing stronger support for trustworthy medical understanding and improving the reliability of downstream clinical applications.




\end{document}